\documentclass[runningheads]{llncs}

\usepackage{eccv}
\usepackage{eccvabbrv}
\usepackage{graphicx}
\usepackage{booktabs}
\usepackage{algorithm}
\usepackage{algpseudocode}
\usepackage{mathtools}
\usepackage{dblfloatfix}
\usepackage{siunitx}
\usepackage{balance}
\usepackage{rotating}
\usepackage{makecell}
\usepackage{caption}
\usepackage{multicol}
\usepackage{wrapfig}
\usepackage[dvipsnames]{xcolor}
\usepackage{amsmath}
\usepackage{amssymb}
\usepackage{bm}
\usepackage{pifont}
\usepackage{enumitem}
\usepackage{soul}
\usepackage{colortbl}
\usepackage{listings}
\usepackage{arydshln}
\usepackage{multirow}
\usepackage{etoolbox}
\usepackage[accsupp]{axessibility} 
\usepackage{hyperref}
\usepackage{orcidlink}

\definecolor{OurOrange}{HTML}{ff5400} %

\def\eg{\emph{e.g}\bmvaOneDot}
\def\Eg{\emph{E.g}\bmvaOneDot}
\def\etal{\emph{et al}\bmvaOneDot}
\def \ie {\emph{i.e.},}
\def \eg {\emph{e.g.},}
\def \etal {\emph{et al.}}
\def \wrt {\emph{w.r.t.}}
\def \etc {\emph{etc.}}

\newcommand{\tit}[1]{\smallbreak\noindent\textbf{#1.}}
\newcommand{\medtit}[1]{\medbreak\noindent\textbf{#1.}}
\newcommand{\bigtit}[1]{\bigbreak\noindent\textbf{#1.}}
\newcommand{\tinytit}[1]{\noindent\textbf{#1.}}
\newcommand{\tinytextit}[1]{\noindent\textit{#1:}}
\newcommand{\mbb}[1]{\mathbf{#1}}

\begin{document}

\title{Merging and Splitting Diffusion Paths \\ for Semantically Coherent Panoramas} 
\titlerunning{Merging and Splitting Diffusion Paths for Coherent Panoramas}

\author{Fabio Quattrini\orcidlink{0009-0004-3244-6186} \and Vittorio Pippi\orcidlink{0009-0001-7365-6348} \and \\ Silvia Cascianelli\orcidlink{0000-0001-7885-6050} \and Rita Cucchiara\orcidlink{0000-0002-2239-283X}}

\authorrunning{F.~Quattrini et al.}

\institute{University of Modena and Reggio Emilia, Modena, Italy\\
\email{\{name.surname\}@unimore.it}}

\maketitle

\begin{abstract}
\vspace{-8pt}
Diffusion models have become the State-of-the-Art for text-to-image generation, and increasing research effort has been dedicated to adapting the inference process of pretrained diffusion models to achieve zero-shot capabilities. 
An example is the generation of panorama images, which has been tackled in recent works by combining independent diffusion paths over overlapping latent features, which is referred to as \textit{joint diffusion}, obtaining perceptually aligned panoramas. 
However, these methods often yield semantically incoherent outputs and trade-off diversity for uniformity. 
To overcome this limitation, we propose the \textbf{M}erge-\textbf{A}ttend-\textbf{D}iffuse operator, which can be plugged into different types of pretrained diffusion models used in a joint diffusion setting to improve the perceptual and semantical coherence of the generated panorama images. 
Specifically, we merge the diffusion paths, reprogramming self- and cross-attention to operate on the aggregated latent space. Extensive quantitative and qualitative experimental analysis, together with a user study, demonstrate that our method maintains compatibility with the input prompt and visual quality of the generated images while increasing their semantic coherence. We release the code at \url{https://github.com/aimagelab/MAD}.
\vspace{-8pt}
\vspace{-5pt}
\end{abstract}

\section{Introduction}
\label{sec:intro}
Diffusion models~\cite{sohl2015deep, song2019generative, song2020score, ho2020denoising} have achieved impressive performance in various domains, including image generation~\cite{dhariwal2021diffusion, ramesh2022hierarchical, saharia2022image, rombach2022high, balaji2022ediffi}, video generation~\cite{wu2023tune, ho2022video, esser2023structure}, and audio synthesis~\cite{kong2020diffwave, chen2020wavegrad, popov2021grad}, especially when guided by a natural language prompt. 
Once trained on large-scale datasets, these models can be exploited for zero-shot adaptation in downstream tasks, such as inpainting~\cite{lugmayr2022repaint}, image editing~\cite{hertz2022prompt, avrahami2022blended, avrahami2023blended}, 3D modeling~\cite{lin2023magic3d, xu2023dream3d}, and video manipulation~\cite{khachatryan2023text2video, wang2023zero}. 

The most popular text-to-image diffusion models are trained to generate fixed-size images. Although some methods are trained with different image resolutions and aspect ratios~\cite{podell2023sdxl, chen2023pixartalpha}, not all image dimensions can be provided during training. This is especially true for oddly-shaped images, such as panoramas. 
Note that obtaining large or oddly-shaped images by simply applying rescaling or superresolution techniques to a generated square image can lead to unsatisfactory results due to the artifacts that would be introduced, the lack of detail, and the uneven enlargement that long images require. 
In light of these considerations, and as the size of the models (and the associated training cost) continues to increase, there is a growing research effort toward inference-time strategies to exploit pretrained models for generating images with unseen shapes and sizes.

\begin{figure}[t]
    \centering
    \includegraphics[width=.85\linewidth]{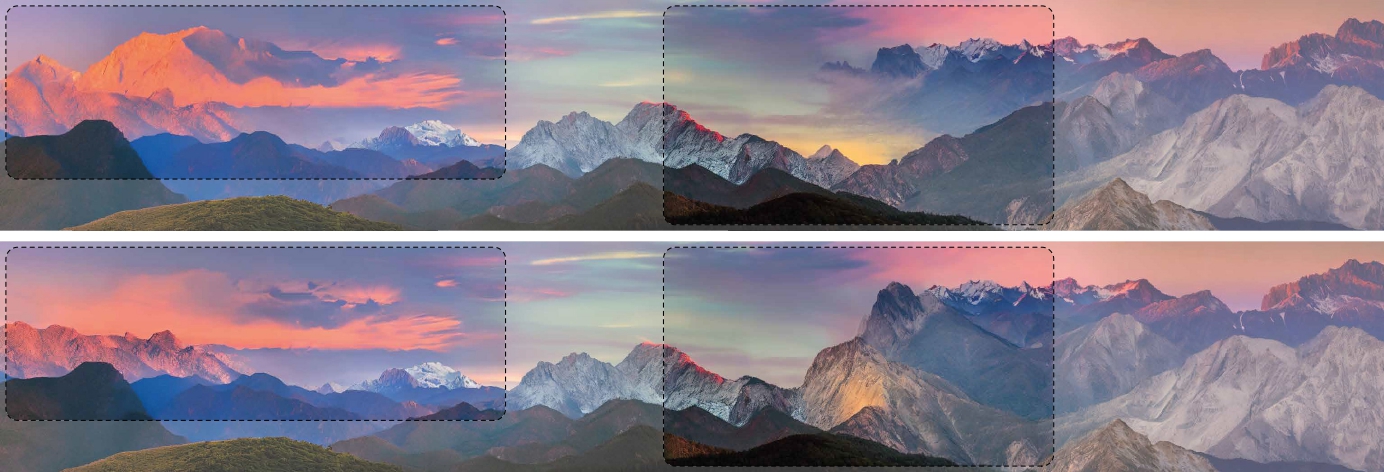}\vspace{-.2em}
    \caption{MultiDiffusion~\cite{bar2023multidiffusion} blends the mountains with the clouds, lacking semantic coherence (top). Applying the proposed MAD leads to semantic coherence (bottom).}
    \label{fig:mountain_comparison_method}\vspace{-1.8em}
\end{figure}

Among those, \textit{iterative diffusion} methods~\cite{avrahami2023blended,zhang2023diffcollage} involve generating an initial image and then, with a series of outpainting steps, generating the whole panorama image. However, this strategy struggles to maintain global coherence and tends to repeat similar patterns~\cite{lee2023syncdiffusion}. 
Another family of approaches, referred to as \textit{joint diffusion} methods~\cite{bar2023multidiffusion, lee2023syncdiffusion}, entails considering the panorama image canvas, splitting it into overlapping views, performing the diffusion denoising process on each view independently, and combining the resulting scores or latent features at each timestep.
This strategy achieves better-blended transitions between adjacent views. However, these are still processed independently, and thus, the resulting image still lacks global coherence and sometimes presents evident artifacts (\eg~a mountain smoothly turning into a cloud as in~\cref{fig:mountain_comparison_method}-top). 
Note that, by coherence, one can mean \textit{perceptual coherence}, \ie~exhibiting similar colors, illumination conditions, and texture throughout the image, or \textit{semantic coherence}, \ie~containing the right amount of elements (objects or parts) with a consistent layout and arranged in a meaningful position in the image. 
Achieving perceptual coherence has been approached in~\cite{lee2023syncdiffusion} by performing gradient descent during the denoising process with a perceptual loss. Instead, obtaining semantic coherence is still an open challenge, which entails modeling long-range relations among the elements (for example, the images in~\cref{fig:mountain_comparison_method} are both perceptually coherent, but the first one, which is obtained with the State-of-the-Art MultiDiffusion~\cite{bar2023multidiffusion} strategy, contains semantic inconsistencies).

In this work, we take a step towards the generation of long images that are both perceptually and semantically coherent while maintaining variability and realism. 
To this end, we devise a joint diffusion pipeline to exploit pretrained diffusion models by manipulating their attention operators. Similar to previous approaches~\cite{bar2023multidiffusion, lee2023syncdiffusion}, we combine the predictions over multiple, strided views. Unlike those approaches, which process the views independently and combine them \textit{at the end} of each denoising step, we devise a strategy to provide additional interaction points \textit{within} the denoising step. This way, our approach can both model the long image as a whole and focus on the details in each view. 
In particular, our strategy is based on our proposed Merge-Attend-Diffuse (MAD) operator, which can be selectively applied to the attention layers inside the noise prediction network of a pretrained diffusion model. MAD takes as input the latent features corresponding to the different panorama views, merges them into a single tensor by averaging the overlapped regions, and feeds it to the self- or cross-attention layer it is applied to. Finally, the obtained latent tensor is split back into views for the next layers, which can follow the standard joint diffusion paradigm. 
Our proposed strategy based on the MAD operator can be applied to different types of pretrained diffusion models to generate perceptually and semantically coherent montages. In this work, we utilize as backbone a pretrained Latent Diffusion Model (LDM), namely, the popular Stable Diffusion~\cite{rombach2022high}, and a Latent Consistency Model (LCM)~\cite{luo2023latent} and demonstrate its effectiveness via qualitative and quantitative evaluations, as well as a user study. The code for our approach is available at \url{https://github.com/aimagelab/MAD}.
\section{Related Work}
\label{sec:related}

Diffusion models~\cite{sohl2015deep, song2019generative, song2020score, song2020improved, ho2020denoising, nichol2021improved, dhariwal2021diffusion} are powerful generative probabilistic models that have achieved drastic performance improvements compared to previous methods, such as Generative Adversarial Networks~\cite{goodfellow2014generative, brock2018large, karras2020analyzing}, especially in the case of text-to-image generation, where the generation process is conditioned on a natural language prompt~\cite{ramesh2022hierarchical, rombach2022high, saharia2022image, podell2023sdxl, chen2023pixartalpha}.
The introduction of LDMs~\cite{rombach2022high} has even increased the popularity of diffusion models. LDMs perform the diffusion process in the latent space and thus require fewer computational resources for both training and inference.
Moreover, recent works have introduced strategies for fine-tuning diffusion models over additional guiding signals to gain better controllability~\cite{gafni2022make, brooks2023instructpix2pix, avrahami2023spatext, zhang2023adding} and for manipulating the diffusion process at inference-time to obtain zero-shot capabilities~\cite{meng2021sdedit, lugmayr2022repaint, hertz2022prompt, avrahami2022blended, avrahami2023blended, mokady2023null, couairon2022diffedit, tumanyan2023plug}. 
Some of the proposed strategies for inference-time adaptation exploit the attention layers of pretrained models to perform tasks such as image editing~\cite{hertz2022prompt}, varied-size image generation~\cite{jin2024training}, or zero-shot video generation~\cite{khachatryan2023text2video}.
Inspired by these works, we propose to modify the attention operations of diffusion models pretrained on square images for generating coherent panorama images in a zero-shot fashion.

\tit{Inference-Time Panorama Generation}
With most of the available diffusion models trained to generate images with limited aspect ratio ranges, being able to exploit such models to generate larger images, possibly at different, odd aspect ratios, without retraining has received increasing interest~\cite{avrahami2023blended, zhang2023diffcollage, bar2023multidiffusion, lee2023syncdiffusion, he2023scalecrafter, jin2024training}.
One of the most commonly considered pretrained models in this context is Stable Diffusion~\cite{rombach2022high}, which is adapted at inference-time to generate long images by following either an iterative diffusion paradigm~\cite{avrahami2023blended, zhang2023diffcollage} or a joint diffusion one~\cite{bar2023multidiffusion, lee2023syncdiffusion}, as we do in this work. 
It is worth noting that large-scale pretrained models can generate long images to some extent, but these images lack variability, which hinders their realism. 
To overcome this limitation, the authors of~\cite{bar2023multidiffusion} introduced the joint diffusion paradigm with MultiDiffusion, an efficient inference-time adaptation strategy to generate long images with variability. This entails generating overlapped squared views and then recombining them in the final panorama image by averaging the noise at the overlap. However, MultiDiffusion can introduce artifacts and lead to perceptually incoherent outputs.
To tackle this issue, the recently-proposed SyncDiffusion~\cite{lee2023syncdiffusion} entails providing guidance in the form of the gradient from a perceptual similarity loss~\cite{zhang2018unreasonable} between the central view of the panorama and the other views. However, SyncDiffusion does not provide guidance for semantic coherence. Moreover, its perceptual coherence guidance is computationally expensive to obtain and requires schedulers that compute the foreseen denoised observation, like~\cite{song2020improved}. 
In this work, we aim to generate varying and both perceptually and semantically coherent long images. 
To this end, different from the previous methods tackling this task, we modify the attention operations inside the backbone used to predict the noise at each sampling step. 

\section{Preliminaries}
\setlength{\belowdisplayskip}{4pt} \setlength{\belowdisplayshortskip}{4pt}
\setlength{\abovedisplayskip}{4pt} \setlength{\abovedisplayshortskip}{4pt}

In this section, we briefly review diffusion and consistency models to provide relevant preliminaries to our pipeline. 

\tit{Diffusion Models}
Diffusion models~\cite{sohl2015deep, ho2020denoising, song2020score, song2020denoising} are a family of probabilistic models that learn to gradually transform a Gaussian noise input $\mathbf{x}_T {\sim} \mathcal{N}(0, \mathbf{I})$ into a data sample $\mathbf{x}_0$ by approximating the data distribution $q$, $\mathbf{x}_0 {\sim} q$ in $T$ steps. 
In particular, the diffusion models framework defines a forward process that gradually injects noise into the data, transforming the data distribution $q(\mathbf{x}_0)$ into the marginal distribution via the transition kernel
\begin{equation*}
    q(\mathbf{x}_t|\mathbf{x}_0) = \mathcal{N} (\mathbf{x}_t; \alpha_t\mathbf{x}_0, \sigma_t^2 \mathbf{I}),
\end{equation*}
where $(\alpha_t, \sigma_t)$ specifies a differentiable noise schedule such that $q(\mathbf{x}_t){\approx} \mathcal{N}(0, \mathbf{I})$. 
The model is trained to reverse this process by learning to denoise $\mathbf{x}_T{\sim}q(\mathbf{x}_t|\mathbf{x}_0)$. 
Predicting $\mathbf{x}_0$ is done iteratively in the reverse diffusion process by estimating $\mathbf{x}_{t-1}$ starting from $\mathbf{x}_t$. 
A possible strategy is to use the $\mathbf{\epsilon}$-prediction parameterization for sampling, as done in Denoising Diffusion Probabilistic Models (DDPMs)~\cite{ho2020denoising}. This entails parametrizing both $\mathbf{x}_t$ and $\mathbf{x}_{t-1}$ as a combination of $\mathbf{x}_0$ and the noise $\mathbf{\epsilon}$, scheduled according to the noise schedule.
In this way, the model can be trained to estimate $\mathbf{\epsilon}$ directly, which is obtained by optimizing
\begin{equation*}
    \mathbb{E}_{q(\mathbf{x}_0)}[\|\mathbf{\epsilon}_{\theta}(\mathbf{x}_t, t) - \mathbf{\epsilon}\|_2^2].
\end{equation*}
In our pipeline for long image generation, we exploit a diffusion model that has been trained with this strategy. However, in our inference-time adaptation, we use the sampler presented in Denoising Diffusion Implicit Models~\cite{song2020denoising}. 
This is also adopted by other works tackling the long image generation task since it is more efficient than DDPM for requiring fewer sampling steps (usually 25 or 50).

\tit{Latent diffusion models} 
Training diffusion models in the high-resolution pixel-space can be computationally prohibitive. 
Latent diffusion models (LDMs) \cite{rombach2022high, podell2023sdxl} tackle this issue by operating in the latent space of an autoencoder, typically a VQ-GAN~\cite{esser2021taming} or a VQ-VAE~\cite{van2017neural}. These models achieve good performance while using a fraction of the GFlops required by pixel space diffusion models~\cite{rombach2022high, podell2023sdxl, peebles2023scalable, chen2023pixartalpha}. Here, we rely on the pretrained Stable Diffusion~\cite{rombach2022high} LDM.

\tit{Conditional Generation} 
Text-to-image diffusion models are able to generate images whose content is specified via a natural language prompt. This is represented by an embedding $\mathbf{e}$ obtained from the text encoder of a pretrained multimodal model (CLIP~\cite{radford2021learning}, in our case). 
In this work, we exploit diffusion models trained with the Classifier-Free Guidance~\cite{ho2022classifier} strategy for conditioning. 
This entails obtaining the noise prediction as a linear combination of conditional and unconditional predictions, with weight $s$, \ie
\begin{equation*}
    \hat{\mathbf{\epsilon}}_{\theta}(\mathbf{x}_t, \mathbf{e}, t) = \mathbf{\epsilon}_{\theta}(\mathbf{x}_t, \emptyset, t) + s(\mathbf{\epsilon}_{\theta}(\mathbf{x}_t, \mathbf{e}, t) - \mathbf{\epsilon}_{\theta}(\mathbf{x}_t, \mathbf{\emptyset}, t)), 
\end{equation*}
where $\emptyset$ is the embedding of the null prompt. 
Note that the conditional generation is implemented by performing cross-attention with the embedding $\mathbf{e}$ inside the noise estimation network. In light of this, by applying our devised MAD operator to the cross-attention layers, we enforce all the latents of the views to jointly attend the prompt, thus favoring semantic coherence.

\tit{Consistency Models} 
Our proposed MAD operator can be applied to different kinds of diffusion models. As an example, in this work, we apply it also to the recently proposed Consistency Models (CMs)~\cite{song2023consistency}, which can achieve impressive generation results. 
In particular, a consistency model $f_{\theta}$ estimates a function that can directly map any intermediate noisy sample $\mathbf{x}_t$ in the forward process to the origin $\mathbf{x}_0$, which is parameterized as
\begin{equation*}
    f_{\theta}(\mathbf{x}, t) = c_{\text{skip}}(t)\mathbf{x} + c_{\text{out}}(t)F_{\theta}(\mathbf{x},t),
\end{equation*}
where $F_{\theta}(\mathbf{x},t)$ is a neural network and $c_{\text{skip}}(t), c_{\text{out}}(t)$ are differentiable functions.
This way, given an arbitrarily small timestep $t_{\mathbf{\delta}}$, the model respects self-consistency $f_{\theta}(\mathbf{x}_t, t) {=} f_{\theta}(\mathbf{x}_t', t') \ \forall  t,t' {\in} [t_{\mathbf{\delta}}, T]$
and boundary condition $f(x_{t_{\mathbf{\delta}}}, \mathbf{\delta})=x_{t_{\mathbf{\delta}}}$, allowing CMs to generate high-quality images in a few steps.
CMs can be either trained from scratch or distilled from pretrained diffusion models. Moreover, Latent Consistency Models (LCMs)~\cite{luo2023latent}, which work on the latent space, are also available. In this work, we use an LCM distilled from Stable Diffusion~\cite{rombach2022high}.

\section{Proposed Approach}
\begin{figure*}[t]
\includegraphics[width=\linewidth]{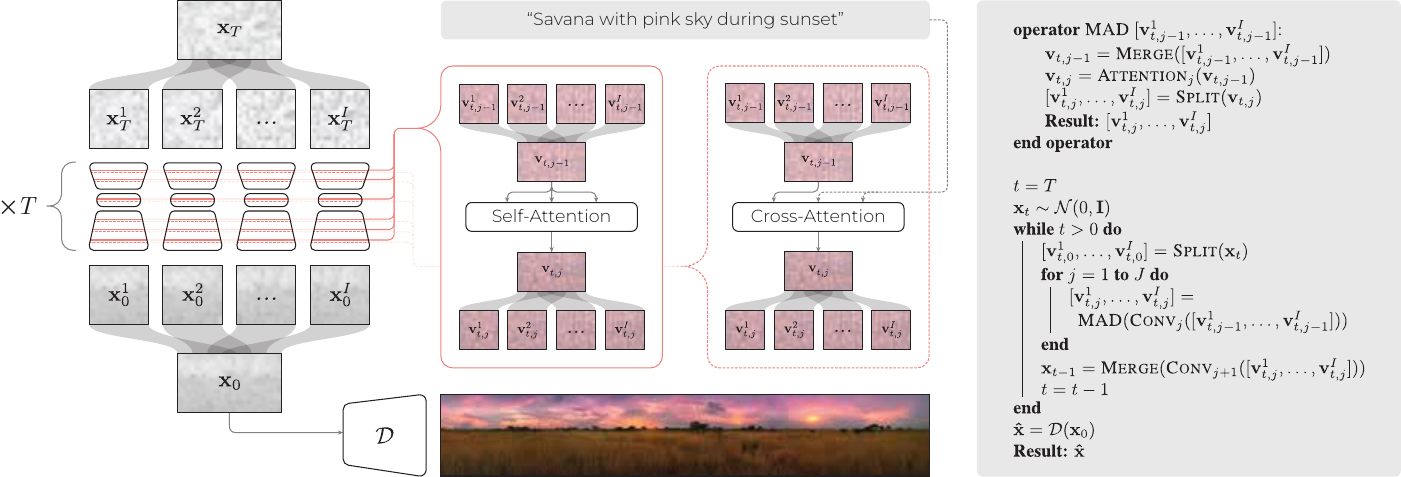}\vspace{-.2em}
\caption{Overview of our inference-time pipeline (left) and its pseudo-code (right). During the diffusion process, the image is split into overlapping views, and each is fed to the model separately. Within the attention layers, MAD provides interaction points between the views, enforcing global coherence in the generated panorama.}
\label{fig:pipeline}\vspace{-1.8em}
\end{figure*}

In this section, we describe our pipeline for generating large images by exploiting a pretrained diffusion model at inference-time (see~\cref{fig:pipeline}). Then, we give the details of our proposed MAD operator, which is the key component of the pipeline for obtaining perceptual and semantic coherence. 

\tit{Joint Diffusion for Panorama Generation}
Diffusion models do not scale well to resolutions and aspect ratios not seen in training~\cite{bar2023multidiffusion,lee2023syncdiffusion}. As a result, generating panorama images by simply applying a model trained on squared ones~\cite{rombach2022high} leads to excessive, unrealistic uniformity in the final output and a lack of variability between images generated from the same prompt, as shown in~\cref{sec:jont_vs_direct} and in~\cite{lee2023syncdiffusion}. 
Thus, for generating larger images, we average the predictions on multiple views at every reverse process step, similar to~\cite{bar2023multidiffusion}. 
Formally, to generate an image $\mathbf{\hat{x}} \in \mathbb{R}^{H {\times} W {\times} C}$, we produce a denoised latent vector $\mathbf{x}_t \in \mathbb{R}^{\frac{H}{n} {\times} \frac{W}{n} {\times} C'}$ where $n$ and $C'$ are, respectively, the scaling factor and the channels input of the decoder $\mathcal{D}$ and $t \in [0,T]$, is the timestep in the denoising process of the latent vector. The final image is obtained as $\mathbf{\hat{x}} = \mathcal{D}(\mathbf{x}_0)$.
To generate panorama images, we split $\mathbf{x}_t$ into $I$ squared overlapped views $\mathbf{x}^i_t \in \mathbb{R}^{L {\times} L {\times} C'}$ where $i \in [1,I]$ denotes the $i$-th view, and $L$ is its width and height.  
We set $L = \frac{H}{n}$ for the horizontal panoramas and $L = \frac{W}{n}$ for the vertical ones.
For simplicity, we consider the noise predictor module as an alternating sequence of convolutional blocks and attention blocks, plus a final convolutional block. Each of the convolutional blocks, dubbed $\textsc{Conv}_{j \in [1,J+1]}$, contains one or more convolutional layers, while each of the attention blocks, dubbed $\textsc{Attention}_{j \in [1,J]}$, contains one or more cross-attention or self-attention layers.
In the following, we use the notation $\textsc{Conv}_{j}([\dots])$ and $\textsc{Attention}_{j}([\dots])$ to indicate that the block is applied to a list of vectors independently.
We define as $\mathbf{v}_{t,j} \in \mathbb{R}^{H_j {\times} W_j {\times} C_j}$ the feature vector output of the $j$-th block and as $\mathbf{v}^i_{t,j} \in \mathbb{R}^{L_j {\times} L_j {\times} C_j}$ its $i$-th view. We remark that $H_j$, $W_j$, $L_j$, and $C_j$ depend on depth and type of the $j$-th block.
Finally, we define the \textsc{Split} and \textsc{Merge} functions. For simplicity, we give their definition by using the latent $\mathbf{x}_t$, but they can be applied also to the $\mathbf{v}_{t,j}$ features. 
$\textsc{Split}: \mathbf{x}_t \rightarrow [\mathbf{x}^1_t, \dots, \mathbf{x}^{I}_t]$ divides its input tensor into multiple overlapping views. 
$\textsc{Merge}: [\mathbf{x}^1_t, \dots, \mathbf{x}^{I}_t] \rightarrow \mathbf{x}_t$, merges its input sequence of overlapping views in a single tensor by averaging the overlapped regions. The number of views depends on the extent of the overlap. 
The more the views overlap, the more seamless the transition between them will be in the final image, at the cost of increasing the inference-time. As a trade-off, we use an overlap of $\frac{3}{4}L$ as in~\cite{lee2023syncdiffusion}.

Note that in previous diffusion approaches for large image generation~\cite{bar2023multidiffusion, lee2023syncdiffusion}, each of the views is processed independently by the noise prediction model. Then, the outputs of the prediction model at the $t$-th step, $[\mathbf{x}^1_t, \dots, \mathbf{x}^{I}_t]$, are combined back into $\mathbf{x}_t$ by averaging the overlapping regions of the views. This averaging procedure allows sharing information between the views. However, in this way, the interactions between the views are infrequent and may be insufficient, which may lead to inconsistent results. 
Different from these approaches, we introduce interaction points between the views also inside the noise prediction model, which are implemented via our proposed MAD operator.

\tit{Merge-Attend-Diffuse Operator}
\label{sec:method} 
We design the MAD operator to be an alternative to the attention layers in the noise prediction module of a pretrained diffusion model when this is used in a joint diffusion setting to generate long images. Through MAD, we enforce consistency between the views while keeping the variability introduced by the convolutional layers, which operate on the views independently. 
Moreover, note that the MAD operator can be integrated at different stages of the noise prediction network and for a varying number of denoising steps starting from $T$, defined by a threshold $\tau$. Note that the views are processed jointly by the attention layers where MAD is applied and independently otherwise. In this way, it is possible to choose the prompt-adequate trade-off between uniformity and variability.
Specifically, MAD takes as input the feature vectors coming from the $(j{-}1)$-th block run over the views, $[\mathbf{v}^1_{t,j-1}, \dots, \mathbf{v}^{I}_{t,j-1}]$, and feeds the $j$-th attention block with the merged latent vector $\mathbf{v}_{t,j-1}$. When the operator is integrated with a cross-attention layer, the cross-attention is performed between the merged tensor and the prompt embedding. When the operator is integrated with a self-attention layer, the self-attention is performed on the merged tensor. In this way, the noise estimation over each view is influenced by the information coming from the other views, thus enforcing global coherence. Finally, the obtained latent vector $\mathbf{v}_{t,j}$ is split back into the views $[\mathbf{v}^1_{t,j}, \dots, \mathbf{v}^{I}_{t,j}]$. The pseudocode of our approach is in~\cref{fig:pipeline}.

\section{Experiments}
\label{sec:experiments}

\tit{Implementation Details}
In our experiments, as the pretrained diffusion models, we use Stable Diffusion 2.0~\cite{rombach2022high} and Latent Consistency Dreamshaper v7 (distilled from Stable Diffusion 1.5)~\cite{luo2023latent} from HuggingFace~\cite{von-platen-etal-2022-diffusers}. 
Both models are based on a U-Net~\cite{ronneberger2015u} architecture, and we apply MAD to all its attention layers. 
For Stable Diffusion, we perform 50 denoising steps and apply MAD for the first $\tau{=}$15 steps. 
For Latent Consistency Dreamshaper, we consider 1, 2, and 4 generation steps and apply MAD with $\tau{=}$2 over 4 steps for the qualitatives. We empirically found that these values of $\tau$ are a good cutoff point, as the global semantic coherence mainly depends on high-level features, defined during the first generation steps.
For generating the panoramas, we consider image views of size 512${\times}$512, which correspond to 64${\times}$64 latents, and a stride of 16 on such latents, following~\cite{lee2023syncdiffusion}.
For the quantitative evaluation, we use the same prompts as for SyncDiffusion~\cite{lee2023syncdiffusion} and MultiDiffusion~\cite{bar2023multidiffusion}, which also exploit our LDM backbone, and report the mean and standard deviation of the performance scores obtained on each prompt. 
For the comparative analyses, we generate the competitors' images with their official codebases. 

\tit{Evaluation Scores}
We adopt the following scores for quantitative evaluation. Mean-CLIP-S (mCLIP)~\cite{hessel2021clipscore}, to evaluate the adherence with the textual prompt; Intra-LPIPS (I-LPIPS)~\cite{zhang2018unreasonable} and Intra-Style Loss (I-StyleL)~\cite{gatys2016image}, to measure the intra-image perceptual coherence; FID~\cite{heusel2017gans}, KID~\cite{binkowski2018demystifying}, and Mean-GIQA (mGIQA)~\cite{gu2020giqa}, to estimate the perceptual similarity and variability with respect to a target distribution.
The I-LPIPS and I-StyleL are computed by splitting each image into non-overlapping squared views and then calculating the average LPIPS and Style Loss of all possible combinations of views (\eg~for a 512${\times}$3072 image, the pairs are 15). 
The FID, KID, and mGIQA are computed by considering, for each prompt, 500 squared images obtained with the baseline LDM or LCM model as reference and single-view-sized random crops from the generated images, one per image. These crops are also used to compute the mCLIP with the input textual prompt. For comparison, we compute the FID, mGIQA, and KID scores for the baseline LDM and LCM by comparing two random halves of the generated images and the average LPIPS and Style Loss by comparing 1000 random pairs of images for each prompt. 
Note that these scores capture different, complementary, and somewhat contrasting characteristics of the generated images. As a result, methods that perform well in terms of some scores might achieve worse values for other scores.\vspace{-1.em}

\begin{figure}[t]
\begin{minipage}[b]{0.485\linewidth}
\centering
    \scriptsize
    \setlength{\tabcolsep}{0.15em}
    \begin{tabular}{m{0.6em} c}
         \rotatebox[origin=l]{90}{\textbf{None}} &
         \makecell{\includegraphics[width=5.5cm]{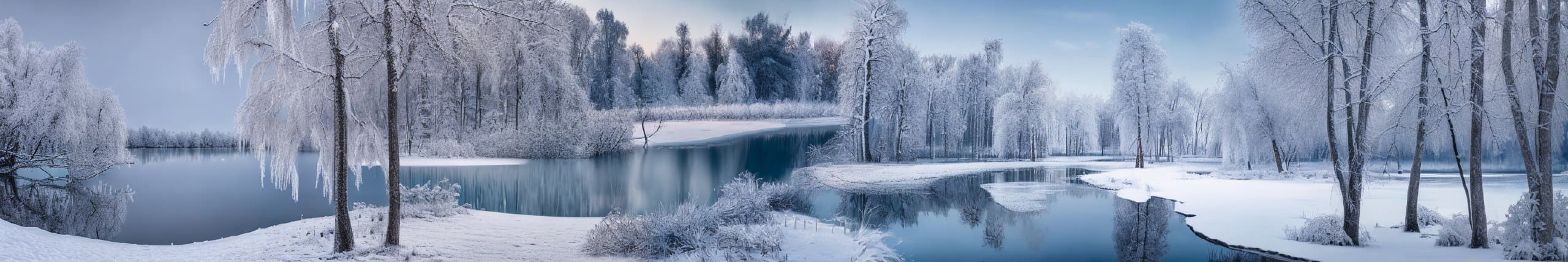}}\\
         \rotatebox[origin=l]{90}{\textbf{Mid}} &
         \makecell{\includegraphics[width=5.5cm]{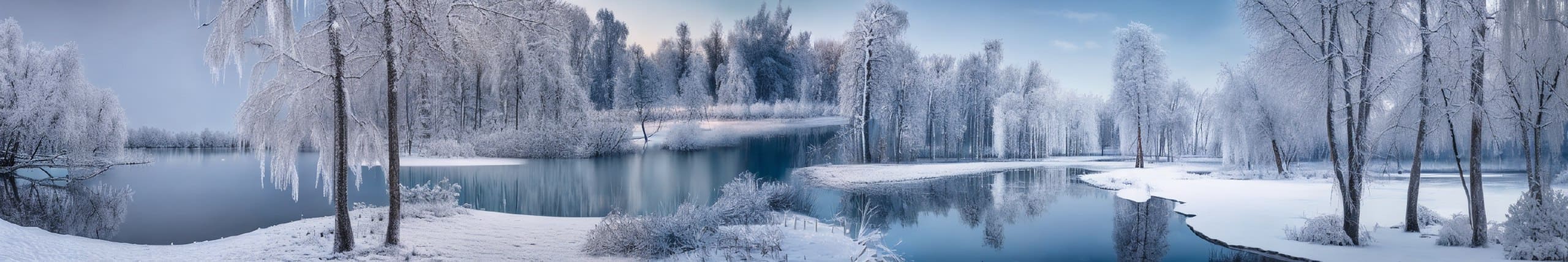}}\\
         \rotatebox[origin=l]{90}{\textbf{Down}} & 
         \makecell{\includegraphics[width=5.5cm]{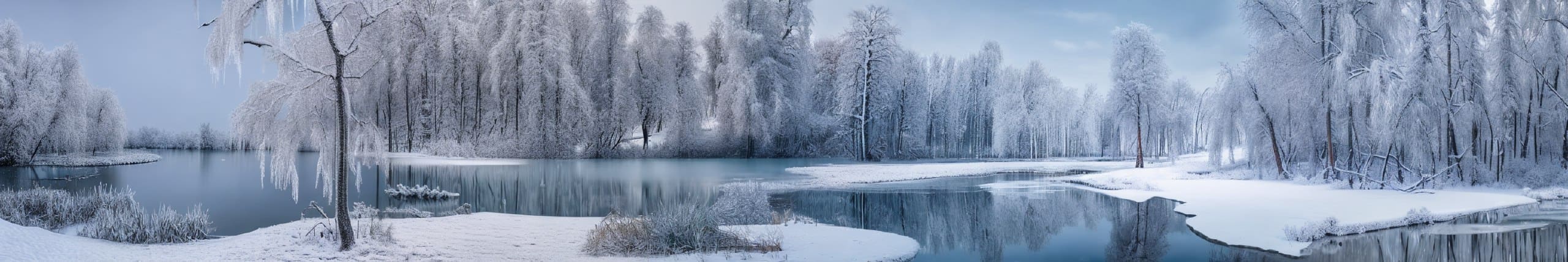}}\\
         \rotatebox[origin=l]{90}{\textbf{Up}} & 
         \makecell{\includegraphics[width=5.5cm]{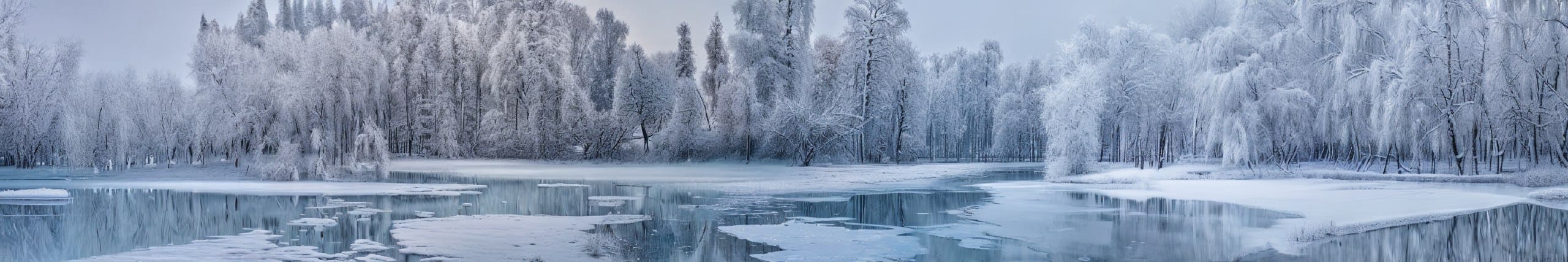}}\\
         \rotatebox[origin=l]{90}{\textbf{All}} &
         \makecell{\includegraphics[width=5.5cm]{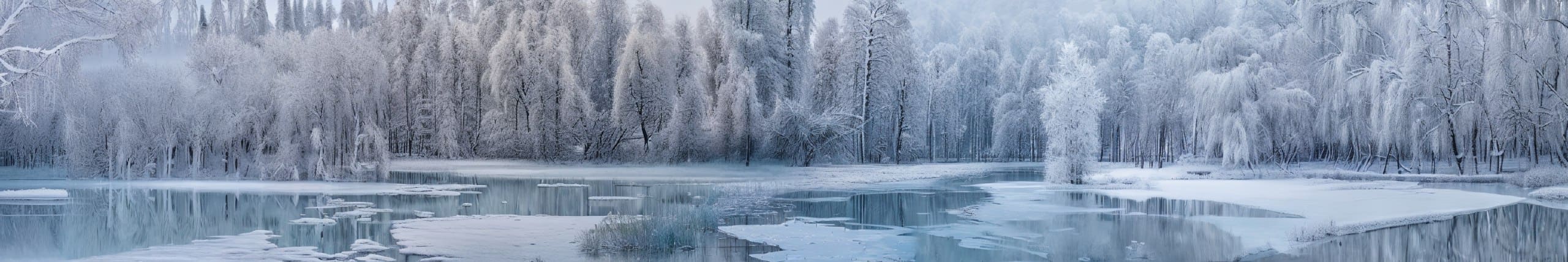}}\\
    \end{tabular}\vspace{-.5em}
\caption{\strut Long images generated by the considered LDM with MAD, with $\tau{=}$15, applied in different blocks of the noise prediction model for the prompt \textit{A snowy winter landscape with frosted trees and a frozen lake}. \textbf{None} is the setting where MAD is never applied.}
\label{fig:ldm_mad_blocks}\vspace{-2.em}
\end{minipage}
\hfill
\begin{minipage}[b]{0.485\linewidth}
    \centering
    \scriptsize
    \setlength{\tabcolsep}{0.15em}
    \begin{tabular}{m{0.6em} c}
         \rotatebox[origin=l]{90}{\textbf{$\tau{=}$0} } & \makecell{\includegraphics[width=5.5cm]{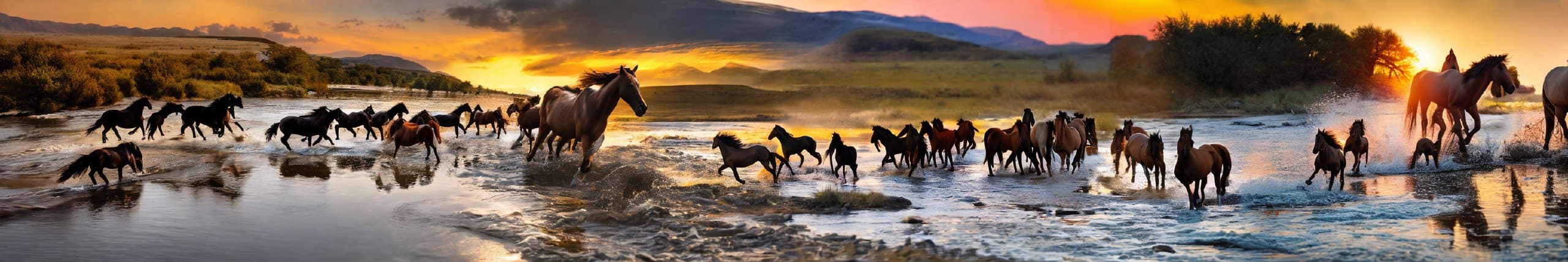}}\\
         \rotatebox[origin=l]{90}{\textbf{$\tau{=}$5} } & \makecell{\includegraphics[width=5.5cm]{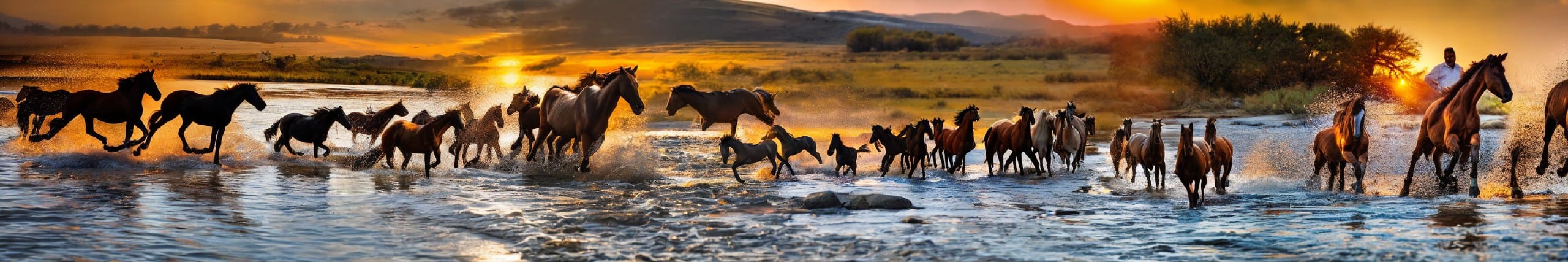}}\\
         \rotatebox[origin=l]{90}{\textbf{$\tau{=}$15}} & \makecell{\includegraphics[width=5.5cm]{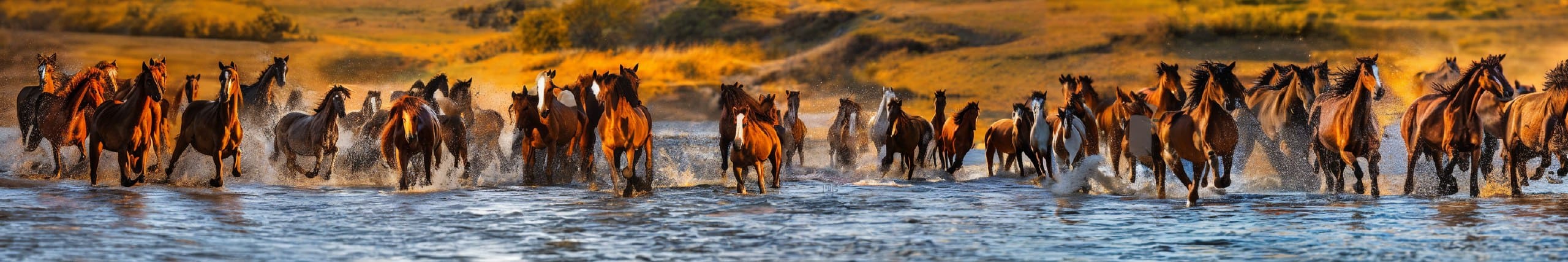}}\\
         \rotatebox[origin=l]{90}{\textbf{$\tau{=}$25}} & \makecell{\includegraphics[width=5.5cm]{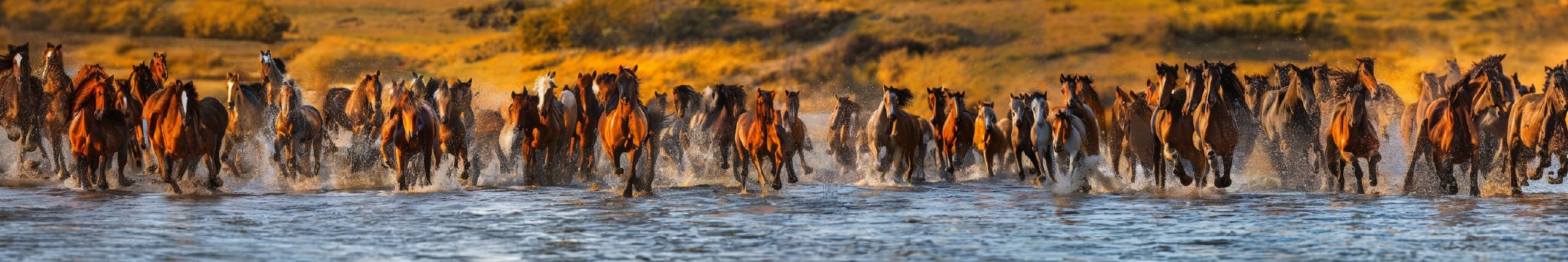}}\\
         \rotatebox[origin=l]{90}{\textbf{$\tau{=}$50}} & \makecell{\includegraphics[width=5.5cm]{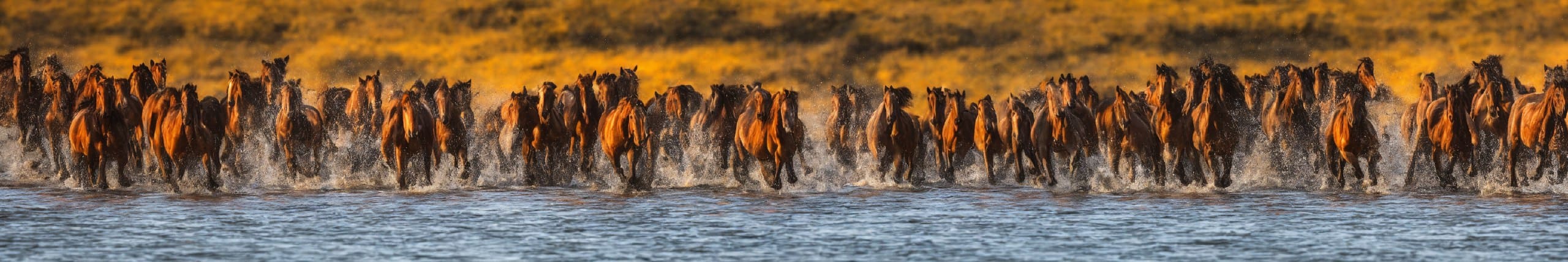}}\\
    \end{tabular}\vspace{-.5em}
\caption{\strut Long images generated by the considered LDM with MAD applied, in all the U-Net attention layers, up to different numbers of inference steps for the prompt \textit{A herd of Mustang horses crossing a river at sunset}. \textbf{$\tau$=0} is the setting where MAD is never applied.}
\label{fig:ablation_threshold}\vspace{-2.em}
\end{minipage}
\end{figure}

\subsection{Results}

\tit{Variety-Uniformity Trade-off} 
The noise prediction model of the considered backbone diffusion models is a U-Net, which features multiple attention layers in the downsampling blocks (referred to as \textit{Down} blocks), bottleneck block (referred to as \textit{Mid} block), and upsampling blocks (referred to as \textit{Up} Blocks). As stated above, we apply MAD in all the U-Net blocks (referred to as \textit{All} blocks). Nonetheless, we design our approach to be modular so that it can be applied to selected layers and for a desired number of timesteps, thus obtaining the desired variety-uniformity trade-off. 
To explore this aspect, in~\cref{fig:ldm_mad_blocks}, we show 512$\times$3072 images generated by applying MAD in all the attention layers of different blocks in the LDM backbone, thus varying the number of interaction points and the stage at which these are performed.
\cref{fig:ldm_mad_blocks} shows different levels of global coherence, ranging from a very coherent scene when MAD is applied on all the blocks (which is our selected setting) to a more varied image when MAD is applied only in the Mid block. 
Quantitative results on the application of MAD at different stages of the U-Net are reported in the supplementary.
Moreover, in~\cref{fig:ablation_threshold}, we show 512$\times$3072 images generated by applying MAD up to a different number of inference steps, defined by the threshold $\tau$ (recall that $\tau{=}$15 in our selected setting). With a low threshold, the images are not globally coherent because the views interactions are insufficient. With increasing thresholds, the images become more and more coherent. We give a quantitative analysis in~\cref{tab:quantitatives} and further substantiate this claim in the supplementary.
Both~\cref{fig:ldm_mad_blocks,fig:ablation_threshold} reflect the trade-off between variety and uniformity. When the views interactions are reduced, either by applying MAD to a few attention layers or for a few timesteps, the resulting image contains several perceptually varying views. When such interactions between the views increase, the global semantic layout and the perceptual patterns are more coherent.\vspace{-.4em}

\begin{table}[t]
\renewcommand{\arraystretch}{.95}
    \footnotesize
        \centering
        \setlength{\tabcolsep}{.28em}
        \caption{Quantitative comparison on 512$\times$3072 panorama generation using the LDM. I-StyleL, KID, and mGIQA values are scaled by 10$^{\text{3}}$.}
        \label{tab:quantitatives}\vspace{-.8em}
        \resizebox{\linewidth}{!}{
        \begin{tabular}{lc c c c c c c c}
        \toprule
        && \textbf{mCLIP $\uparrow$} & \textbf{I-LPIPS $\downarrow$} & \textbf{I-StyleL $\downarrow$} & \textbf{FID $\downarrow$} & \textbf{KID $\downarrow$} & \textbf{mGIQA $\uparrow$} & \textbf{Runtime}\\
        \midrule
        \textbf{SD}                && 31.63${\pm}$1.89 & 0.74${\pm}$0.07 & 0.74${\pm}$0.07 & ~28.31${\pm}$10.89 & ${<}$0.01${\pm}$0.13 & 26.70${\pm}$6.90 & -- \\
        \midrule
        \textbf{SD-L}              && 32.01${\pm}$1.67 & 0.50${\pm}$0.11 & 0.58${\pm}$0.40 & ~87.64${\pm}$30.25 &   ~76.83${\pm}$30.52 & 27.72${\pm}$7.83 & ~88.2${\pm}$0.1s\\ 
        \midrule
        \textbf{MD}                && 31.77${\pm}$2.32 & 0.69${\pm}$0.09 & 2.98${\pm}$2.41 & ~33.52${\pm}$12.43 &   ~~9.04${\pm}$4.23~ & 28.54${\pm}$7.99 & ~41.8${\pm}$0.2s\\
        \textbf{SyncD}             && 31.84${\pm}$2.19 & 0.56${\pm}$0.06 & 1.39${\pm}$1.15 & ~44.60${\pm}$18.45 &   ~21.00${\pm}$11.06 & 27.17${\pm}$6.66 & 392.4${\pm}$0.6s\\
        \midrule 
        \textbf{MAD ($\tau{=}$0)}  && 31.65${\pm}$2.17 & 0.64${\pm}$0.10 & 2.65${\pm}$2.33 & ~34.51${\pm}$13.92 & ~9.19${\pm}$4.51 & 27.59${\pm}$6.83 & ~41.9${\pm}$0.1s\\
        \textbf{MAD ($\tau{=}$5)}  && 31.86${\pm}$2.22 & 0.59${\pm}$0.10 & 2.07${\pm}$1.31 & ~38.10${\pm}$13.71 & 13.75${\pm}$4.00~ & 28.32${\pm}$7.64 & ~42.9${\pm}$0.2s\\
        \textbf{MAD ($\tau{=}$15)} && 32.03${\pm}$2.29 & 0.56${\pm}$0.10 & 1.90${\pm}$1.32 & ~48.52${\pm}$17.14 & 27.15${\pm}$9.10~ & 28.32${\pm}$7.76 & ~44.3${\pm}$0.2s\\
        \textbf{MAD ($\tau{=}$25)} && 32.15${\pm}$2.25 & 0.53${\pm}$0.10 & 1.43${\pm}$1.04 & ~61.76${\pm}$23.73 & 43.31${\pm}$17.41 & 28.10${\pm}$7.84 & ~45.6${\pm}$0.2s\\
        \textbf{MAD ($\tau{=}$40)} && 32.16${\pm}$1.95 & 0.50${\pm}$0.10 & 0.88${\pm}$0.61 & ~86.20${\pm}$37.23 & 76.15${\pm}$37.73 & 27.59${\pm}$7.61 & ~48.3${\pm}$0.1s\\
        \textbf{MAD ($\tau{=}$50)} && 32.14${\pm}$1.72 & 0.49${\pm}$0.10 & 0.71${\pm}$0.47 & ~98.01${\pm}$43.64 & 91.51${\pm}$49.95 & 27.05${\pm}$7.28 & ~49.8${\pm}$0.1s\\        
        \bottomrule
        \end{tabular}
        }
        \vspace{-1.8em}
    \end{table}

\tit{Quantitative Comparison}
We compare the performance of our method on 512$\times$3072 images with different values of $\tau$ against applying Stable Diffusion to generate the whole panorama image directly (which we refer to as SD-L) and the joint diffusion methods MultiDiffusion (MD)~\cite{bar2023multidiffusion} and SyncDiffusion (SyncD)~\cite{lee2023syncdiffusion}, which are the closest to our approach. 
For numerical reference, we also compute the scores on the squared images from Stable Diffusion (SD). The results of this analysis are reported in~\cref{tab:quantitatives}.
From the table, we observe that MAD achieves the best or second-best performance on all the scores, depending on the parameter $\tau$, which determines how many times it is applied in the denoising process. 
In particular, with respect to perceptual and semantic coherence of the generated images, we observe that the more MAD is applied, the more the images are compatible with the textual prompt, as measured by the mCLIP. The obtained values are always superior to those obtained with the competitors and slightly inferior \wrt~SD-L only in the case of a few application steps. Moreover, the I-LPIPS and I-StyleL values indicate that MAD also leads to increasingly perceptually coherent images the more it is applied. It is worth noting that MAD performs on par or better than SyncD in terms of the I-LPIPS, although SyncD specifically entails optimizing this score. 
As for FID, KID, and mGIQA, note that these measure different aspects of the generated images \wrt~the other three considered scores. In particular, I-LPIPS and I-StyleL evaluate the perceptual similarity within each image, while FID, KID, and mGIQA the distance of the distributions of random crops from the images generated by the approach under evaluation and SD. 
If a long image is composed of many perceptually varying views (as those by MD), it is more likely that its crops resemble the squared SD reference images. However, this characteristic does not ensure perceptual coherence between the views in the same long image (measured by I-LPIPS and I-StyleL) nor adherence with the textual prompt (measured by the mCLIP). 
In summary, the results in~\cref{tab:quantitatives} suggest that our approach is able to generate semantically and perceptually coherent panoramas (which is the goal of our work), regardless of the similarity of their single crops to SD squared images. 
\begin{figure*}[t]
\centering
\scriptsize
    \setlength{\tabcolsep}{.2em}
    \begin{tabular}{m{.6em}c c c}
        & \tiny{\textit{A photo of a city skyline at night}} & \tiny{\textit{A natural landscape in anime style illustration}} \\
        \rotatebox[origin=l]{90}{\textbf{MD   }} &
         \makecell{\includegraphics[height=.95cm]{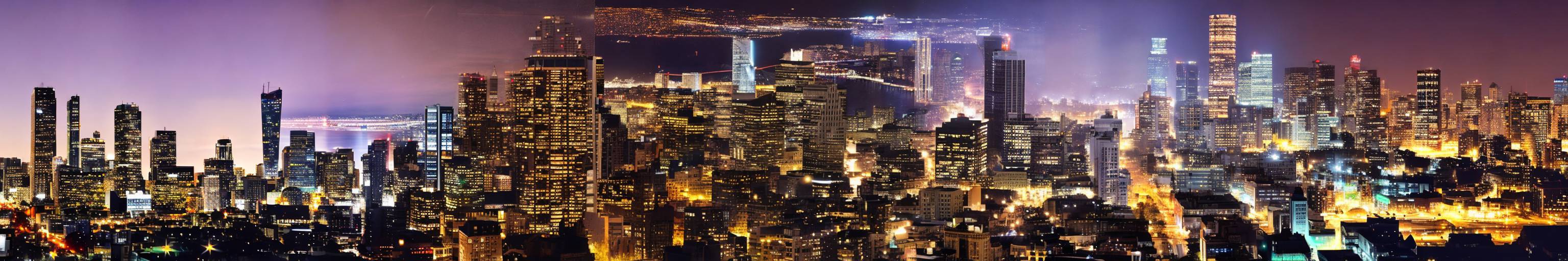}} & 
         \makecell{\includegraphics[height=.95cm]{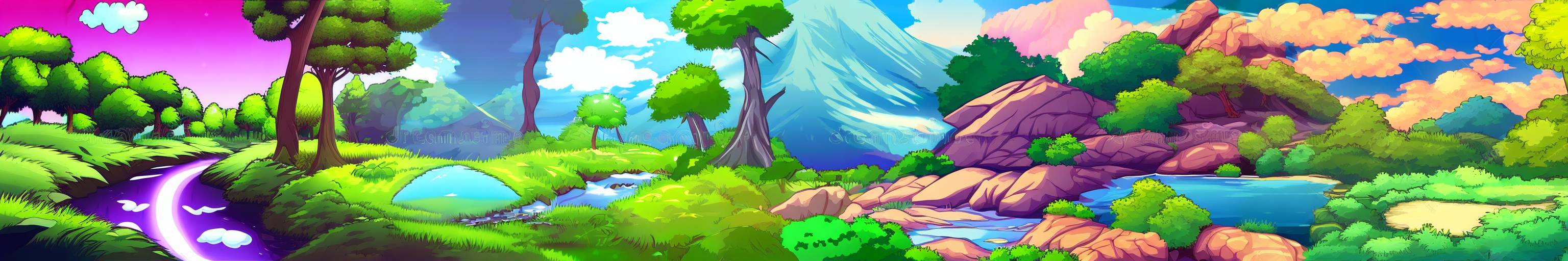}}\\
        \rotatebox[origin=l]{90}{\textbf{SyncD}} &  
         \makecell{\includegraphics[height=.95cm]{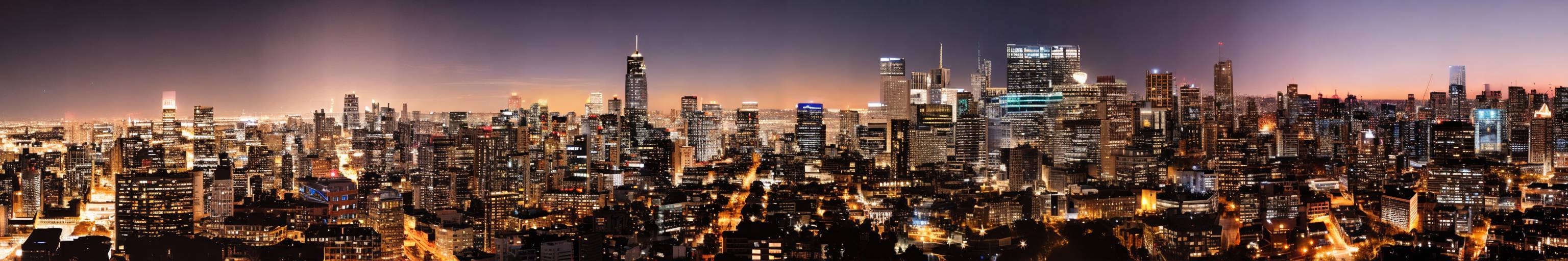}} & 
         \makecell{\includegraphics[height=.95cm]{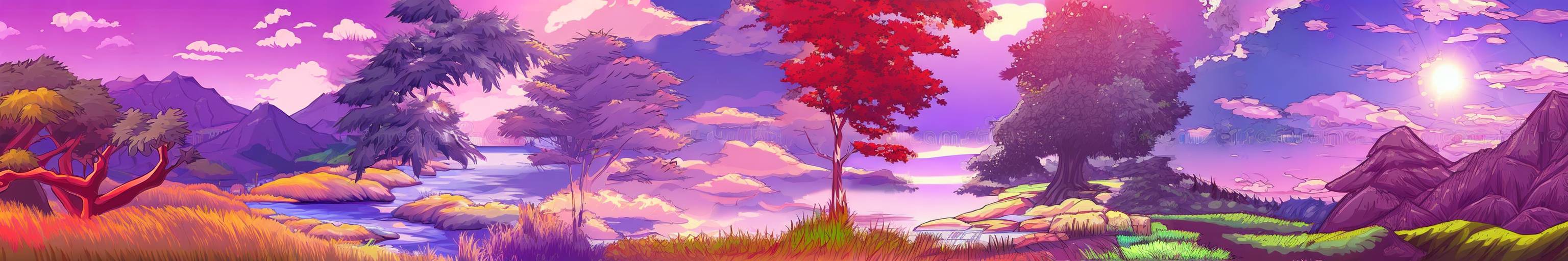}}\\
        \rotatebox[origin=l]{90}{\textbf{MAD}} &
         \makecell{\includegraphics[height=.95cm]{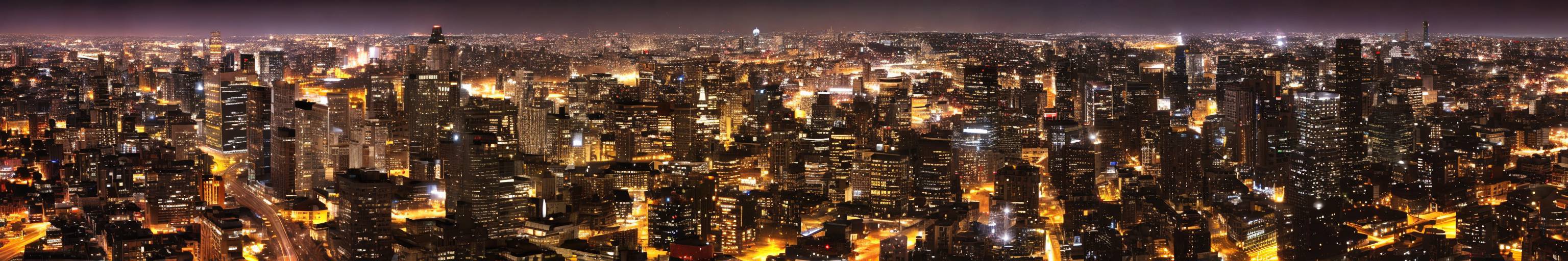}} & 
         \makecell{\includegraphics[height=.95cm]{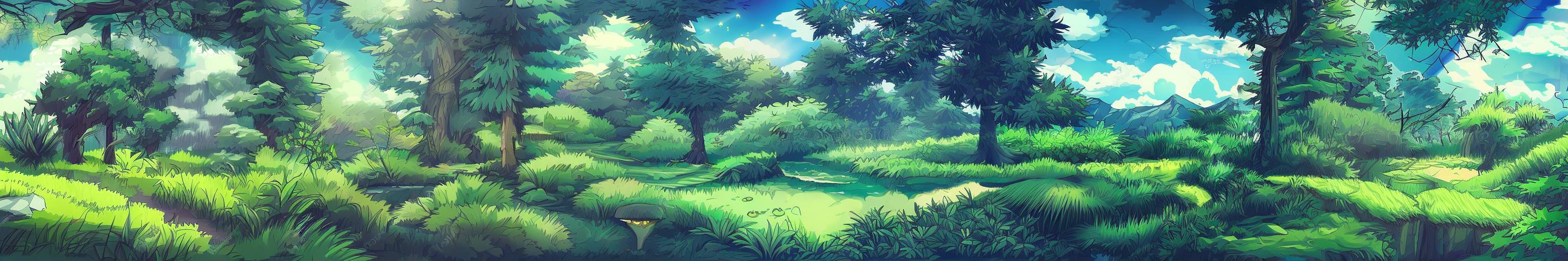}} 
    \end{tabular}\vspace{-.5em}
\caption{Qualitative results of our approach, MD, and SyncD. MD generates panoramas with smooth but incoherent transitions. SyncD increases perceptual coherence but lacks semantic coherence. MAD generates perceptually and semantically coherent images.}
\label{fig:ldm_six_prompts}
\vspace{-1.5em}
\end{figure*}

\tit{Qualitative Comparison}
\cref{fig:ldm_six_prompts} shows a qualitative comparison among the joint diffusion methods on two of the six evaluation prompts (more qualitatives are provided in the supplementary). We generate the images using the same seed for each comparison to ensure consistency. 
As we can see, MD generates images with smooth but incoherent transitions. 
SyncD generally enhances the perceptual coherence of the panoramas but does not address semantic coherence. For instance, in the anime-style landscape of the last row, we can see a lake morphing into clouds. 
Conversely, our approach generates perceptually and semantically coherent panoramas with these prompts. 

\begin{figure}[t]
\begin{minipage}[b]{0.485\linewidth}
    \centering
    \scriptsize
    \arrayrulecolor{white}  
    \setlength{\tabcolsep}{0.1cm}
    \begin{tabular}{m{0.25em} r}
         \rotatebox[origin=l]{90}{\textbf{MD}} &
         \makecell{
         \includegraphics[width=1.8cm]{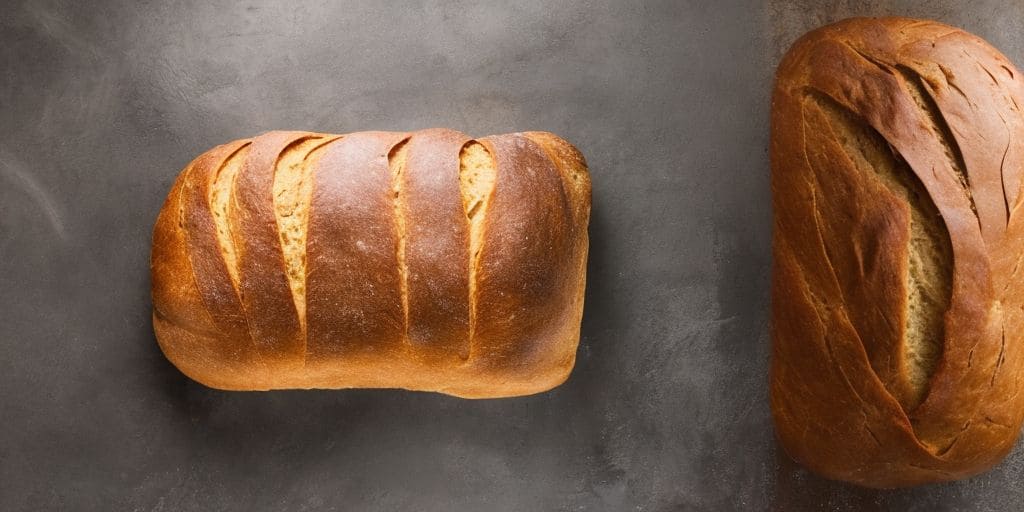}
         \includegraphics[width=3.6cm]{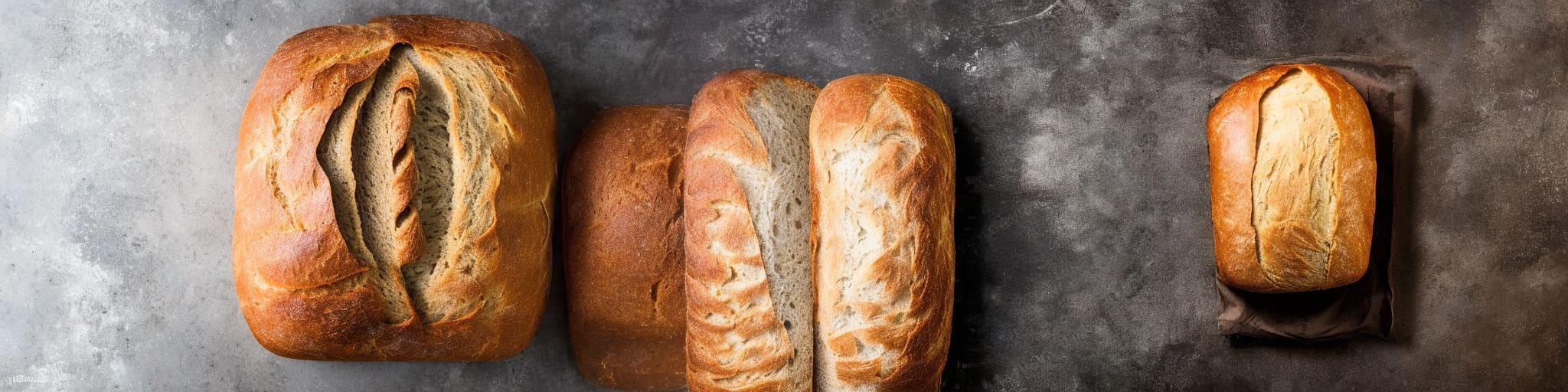} \\
         \includegraphics[width=5.5cm]{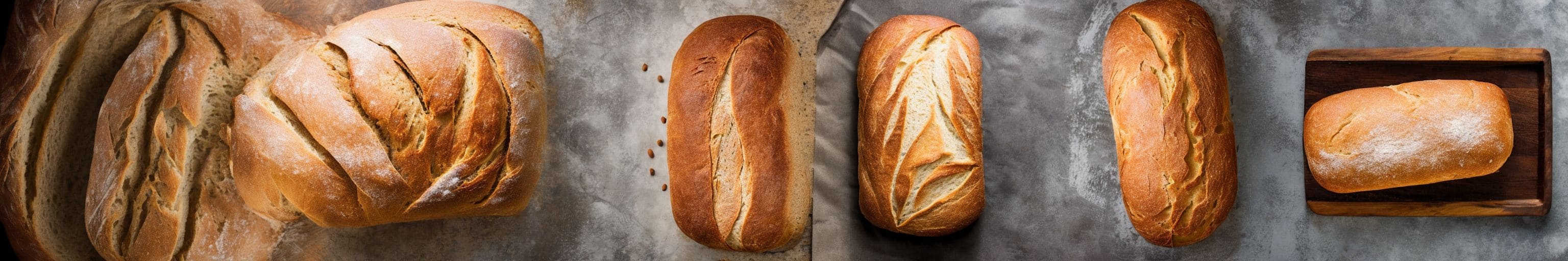}}  \\
        \hline
        \hline
         \rotatebox[origin=l]{90}{\textbf{SyncD}} &
         \makecell{
         \includegraphics[width=1.8cm]{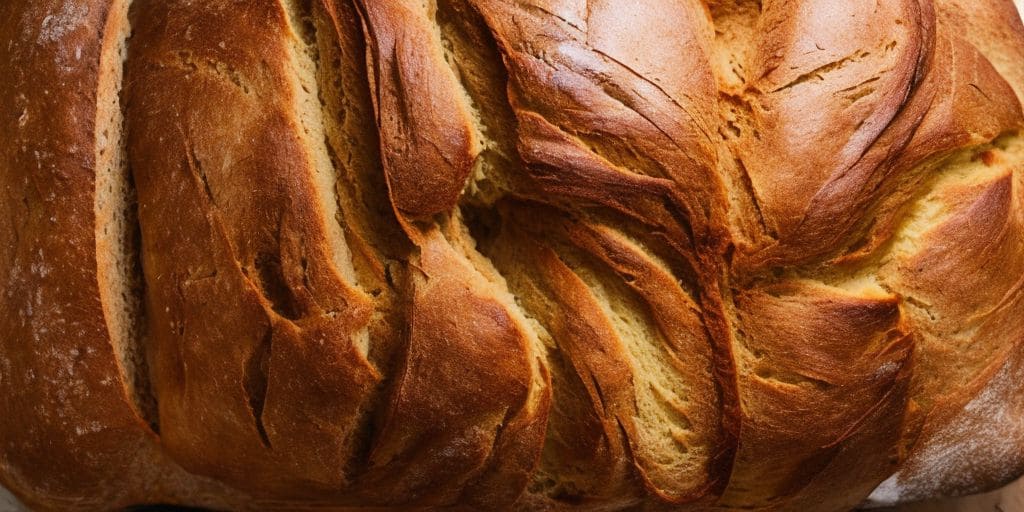}
         \includegraphics[width=3.6cm]{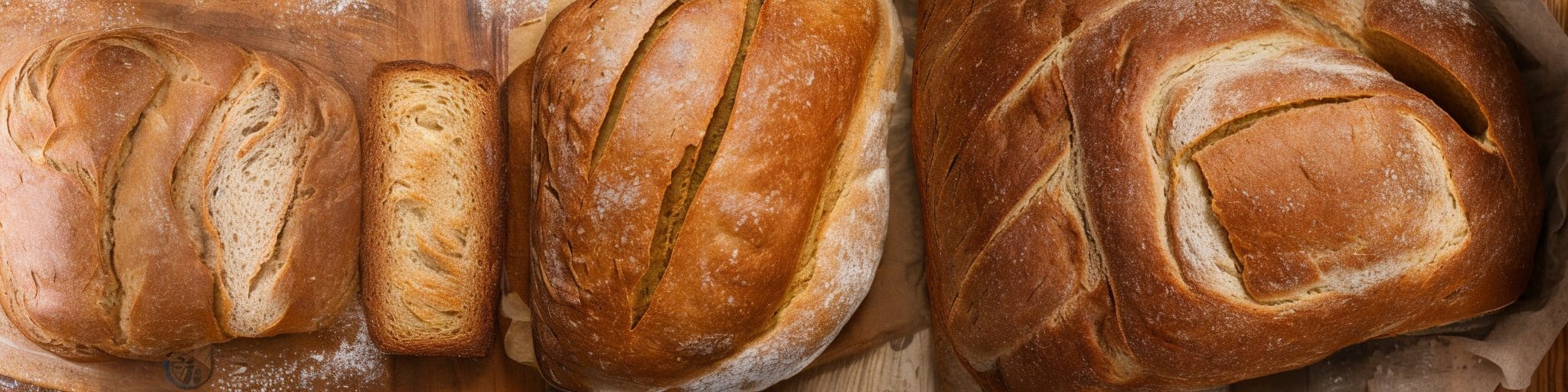} \\
         \includegraphics[width=5.5cm]{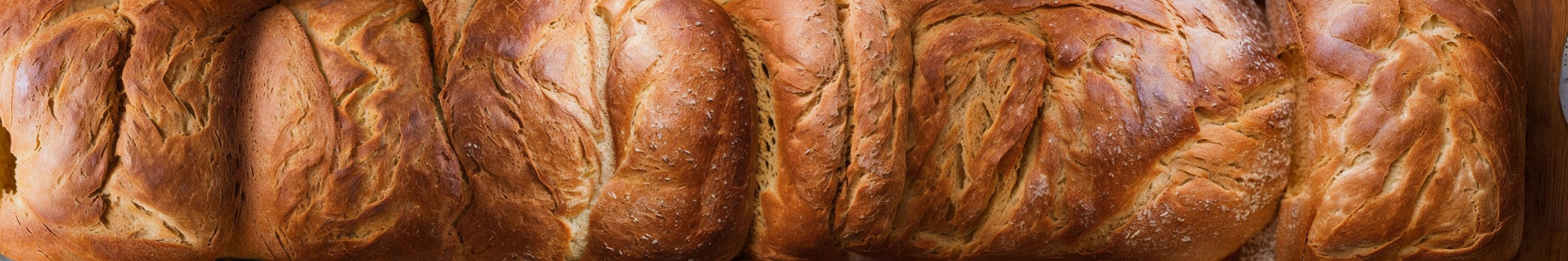}} \\
        \hline
        \hline  
         \rotatebox[origin=l]{90}{\textbf{MAD}} & 
         \makecell{
         \includegraphics[width=1.8cm]{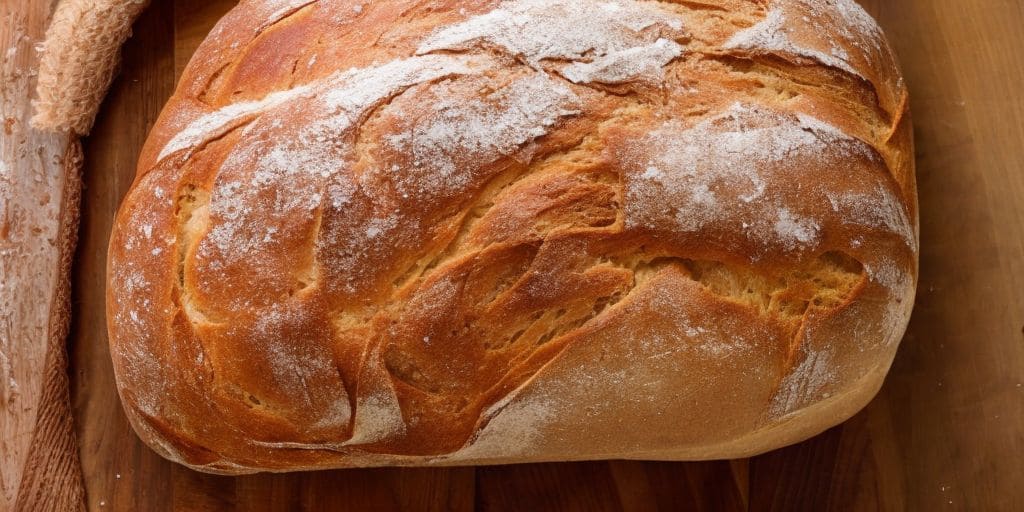}
         \includegraphics[width=3.6cm]{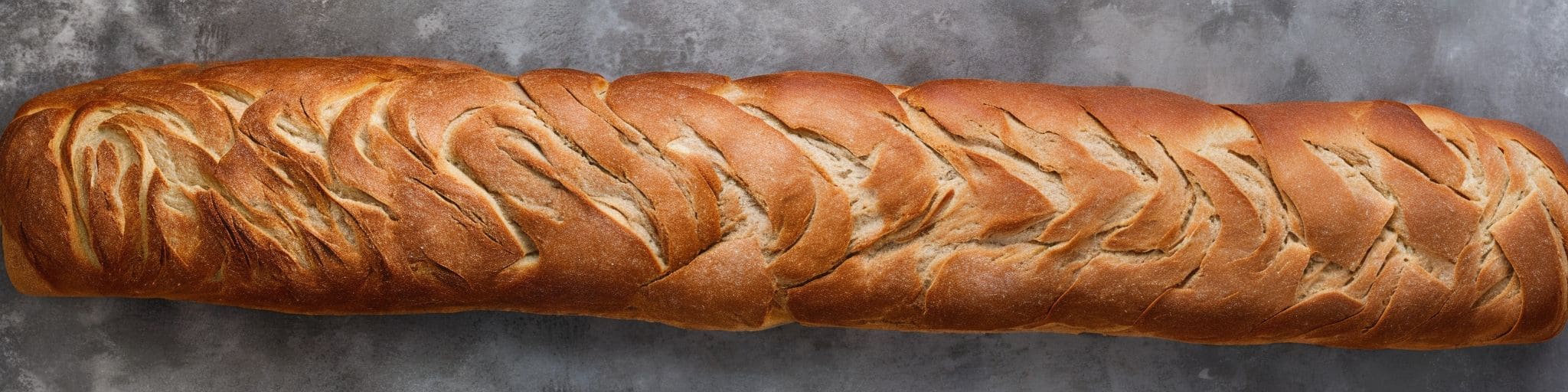} \\
         \includegraphics[width=5.5cm]{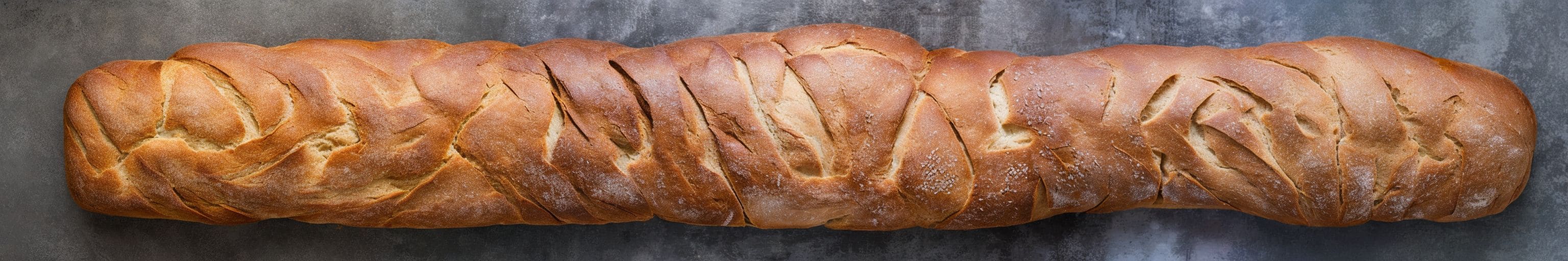}} \\
    \end{tabular}\vspace{.03em}
    \arrayrulecolor{black}
    \vspace{-.5em}
\caption{Qualitative comparisons between our approach with LDM, MD, and SyncD at different aspect ratios for the prompt \textit{Top view of a loaf of bread}.}
\label{fig:aspect_ratios}
\end{minipage}
\hfill
\begin{minipage}[b]{0.485\linewidth}
\centering
    \scriptsize
    \setlength{\tabcolsep}{0.1em}
    \begin{tabular}{cc c cc}
        \textbf{SyncD} & \textbf{MAD} && \textbf{SyncD} & \textbf{MAD} \\
        \includegraphics[width=1.4cm]{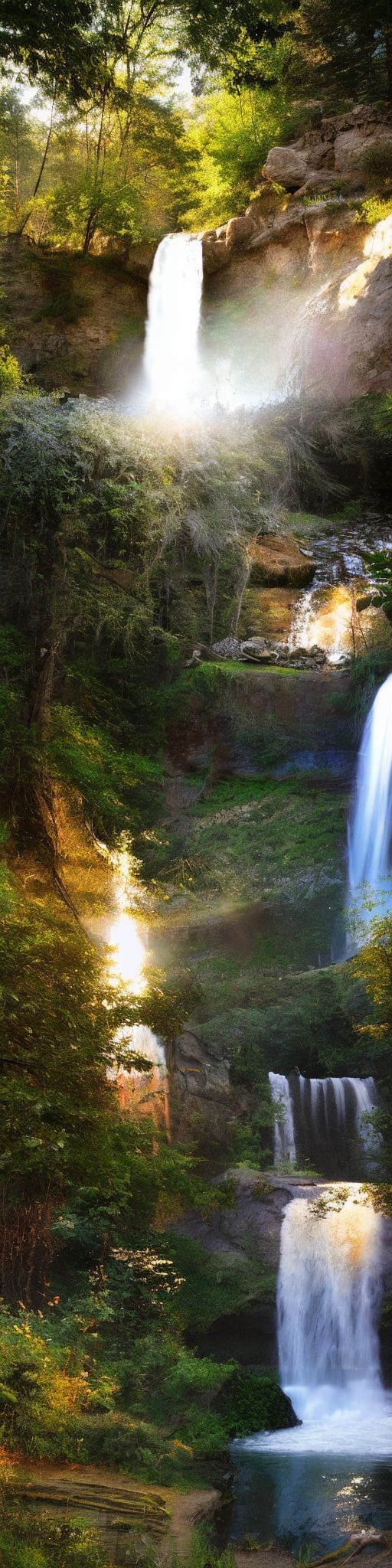} &
        \includegraphics[width=1.4cm]{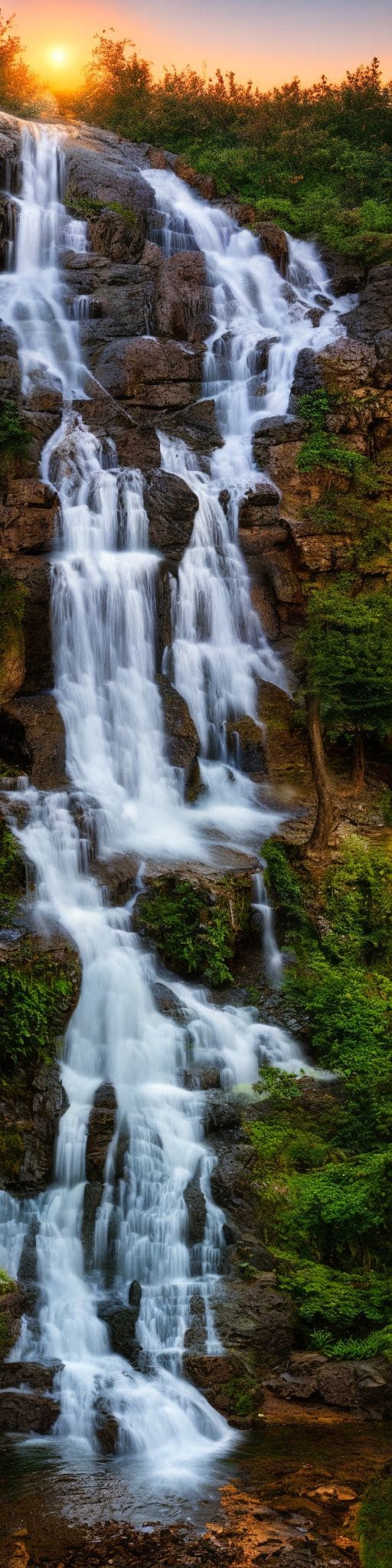} &&
        \includegraphics[width=1.4cm]{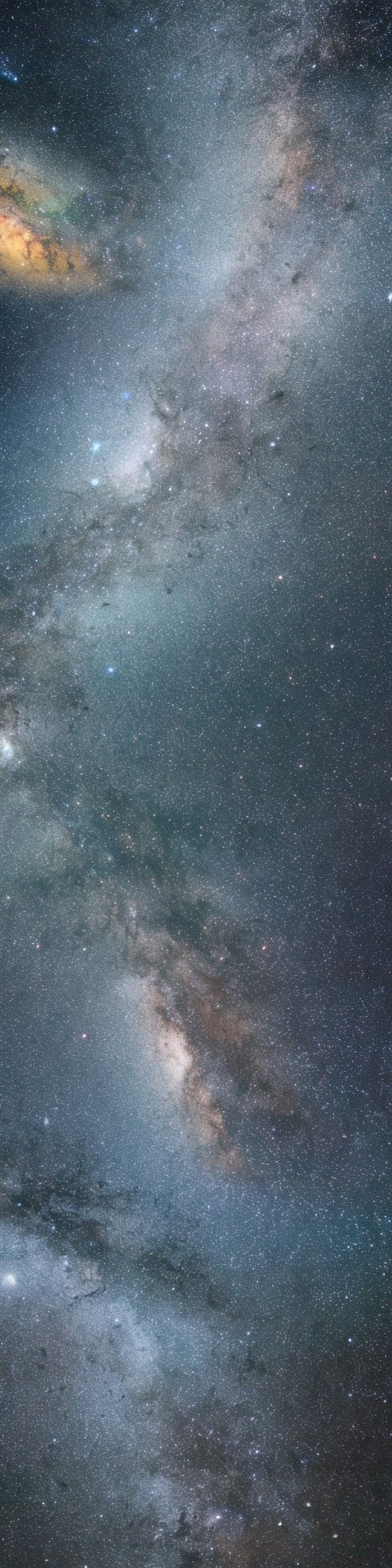} &
        \includegraphics[width=1.4cm]{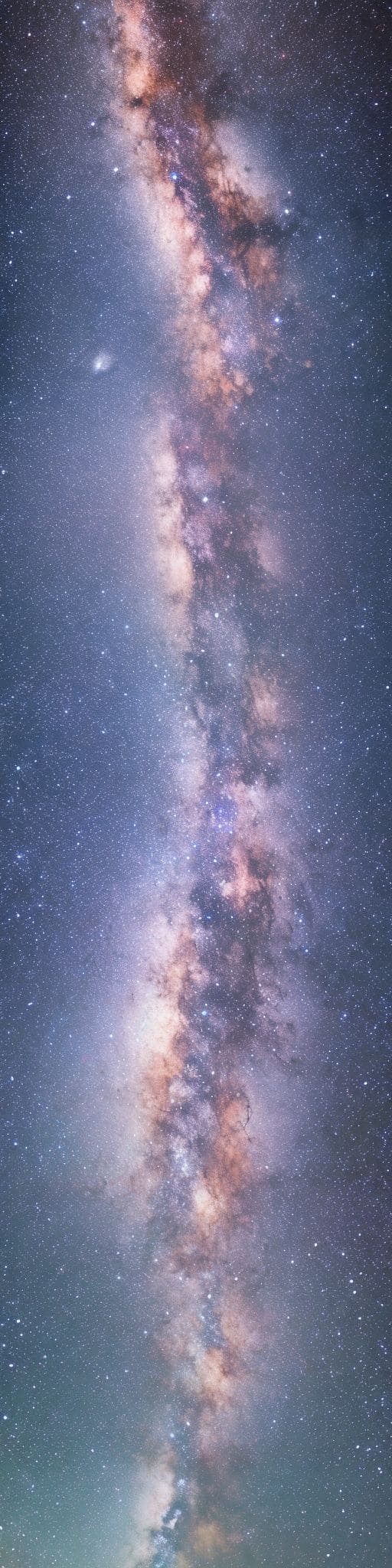} \\
        \multicolumn{2}{c}{\tiny \textit{A waterfall at sunset}} &&
        \multicolumn{2}{c}{\tiny \textit{Milky way}} \\
    \end{tabular}
    \vspace{-.5em}
    \caption{\strut Qualitative comparison on vertical 512{$\times$}2048 images between our approach applied to the LDM and SyncD for prompts requiring semantic coherence.}
    \label{fig:qualitatives_vertical}
\end{minipage}
\vspace{-.4em}
\end{figure}

\begin{table}[h!]
\renewcommand{\arraystretch}{.95}
\footnotesize
    \centering
    \setlength{\tabcolsep}{.28em}
        \caption{Quantitative comparison on the generation of images at different aspect ratios with the LDM. The I-StyleL, KID, and mGIQA are scaled by 10$^{\text{3}}$.}
    \label{tab:quantitatives_aspect_ratios}\vspace{-.8em}
    \resizebox{.8\linewidth}{!}{
    \begin{tabular}{lc c c c c c c }
        \toprule
        && \textbf{mCLIP $\uparrow$}& \textbf{I-LPIPS $\downarrow$} & \textbf{I-StyleL $\downarrow$} & \textbf{FID $\downarrow$} & \textbf{KID $\downarrow$} & \textbf{mGIQA $\uparrow$} \\
    \midrule
    \multicolumn{8}{c}{\textbf{512$\times$1024}}\\
    \midrule
     \textbf{MD}                    && 31.73${\pm}$2.22 & 0.66${\pm}$0.09 & 2.57${\pm}$1.97 & 30.66${\pm}$11.79 & ~5.24${\pm}$3.04~ & 28.17${\pm}$7.54 \\
     \textbf{SyncD}                 && 31.71${\pm}$2.01 & 0.53${\pm}$0.06 & 1.09${\pm}$0.77 & 40.35${\pm}$16.43 & 13.20${\pm}$7.61~ & 26.41${\pm}$6.38 \\
     \textbf{MAD}     && 31.82${\pm}$2.09 & 0.55${\pm}$0.10 & 1.47${\pm}$0.89 & 37.53${\pm}$14.72 & 12.63${\pm}$5.74~ & 27.58${\pm}$7.04 \\ 
    \midrule
    \multicolumn{8}{c}{\textbf{512$\times$2048}}\\
    \midrule
     \textbf{MD}                    && 31.77${\pm}$2.14 & 0.69${\pm}$0.09 & 2.96${\pm}$2.41 & 33.07${\pm}$12.38 & ~8.58${\pm}$3.99~ & 28.33${\pm}$7.79 \\
     \textbf{SyncD}                 && 31.77${\pm}$2.14 & 0.55${\pm}$0.06 & 1.39${\pm}$1.19 & 43.33${\pm}$17.98 & 18.77${\pm}$10.19 & 27.08${\pm}$6.65 \\
     \textbf{MAD}     && 31.97${\pm}$2.23 & 0.56${\pm}$0.10 & 1.79${\pm}$1.21 & 44.26${\pm}$16.26 & 21.80${\pm}$7.54~ & 28.13${\pm}$7.61 \\ 
     \midrule
    \multicolumn{8}{c}{\textbf{512$\times$4096}}\\
    \midrule
     \textbf{MD}                    && 31.64${\pm}$2.18 & 0.64${\pm}$0.10 & 2.18${\pm}$1.16 & 34.87${\pm}$13.56 & ~9.12${\pm}$3.72~ & 27.89${\pm}$7.08 \\
     \textbf{MAD}     && 32.05${\pm}$2.30 & 0.56${\pm}$0.10 & 1.91${\pm}$1.35 & 49.61${\pm}$16.98 & 28.34${\pm}$8.05~ & 28.33${\pm}$7.94 \\ 
    \bottomrule
    \end{tabular}
    }
    \vspace{-1.2em}
\end{table} 
    
\tit{Different Aspect Ratios}
We also evaluate our approach for the generation of images with different aspect ratios. \cref{tab:quantitatives_aspect_ratios} shows the performance of our method and the joint diffusion competitors when generating 512$\times$1024, 512$\times$2048, and 512$\times$4096 images.
As we can see, MAD is consistently better in terms of mCLIP and competitive in I-LPIPS. 
The visual quality of the generated images can also be appreciated in~\cref{fig:aspect_ratios}, where we report 512$\times$1024, 512$\times$2048, and 512$\times$3072 images for the same prompt and seed. It can be observed that our approach can generate perceptually and semantically coherent images at different aspect ratios, while the competitors struggle to achieve this capability even in shorter images. 
We also compare our method with SyncD on long vertical images in~\cref{fig:qualitatives_vertical} by generating 2048$\times$512 images. We chose to compare only with SyncD because, compared to MD, it enforces perceptual coherence. Nonetheless, it struggles to maintain semantic coherence and to generate long images with subjects that should have a beginning and an end, such as the waterfall.
Please refer to the supplementary for more qualitative results at different aspect ratios.

\begin{figure}[t]
\begin{minipage}[b]{0.315\linewidth}
\centering
    \includegraphics[width=\linewidth]{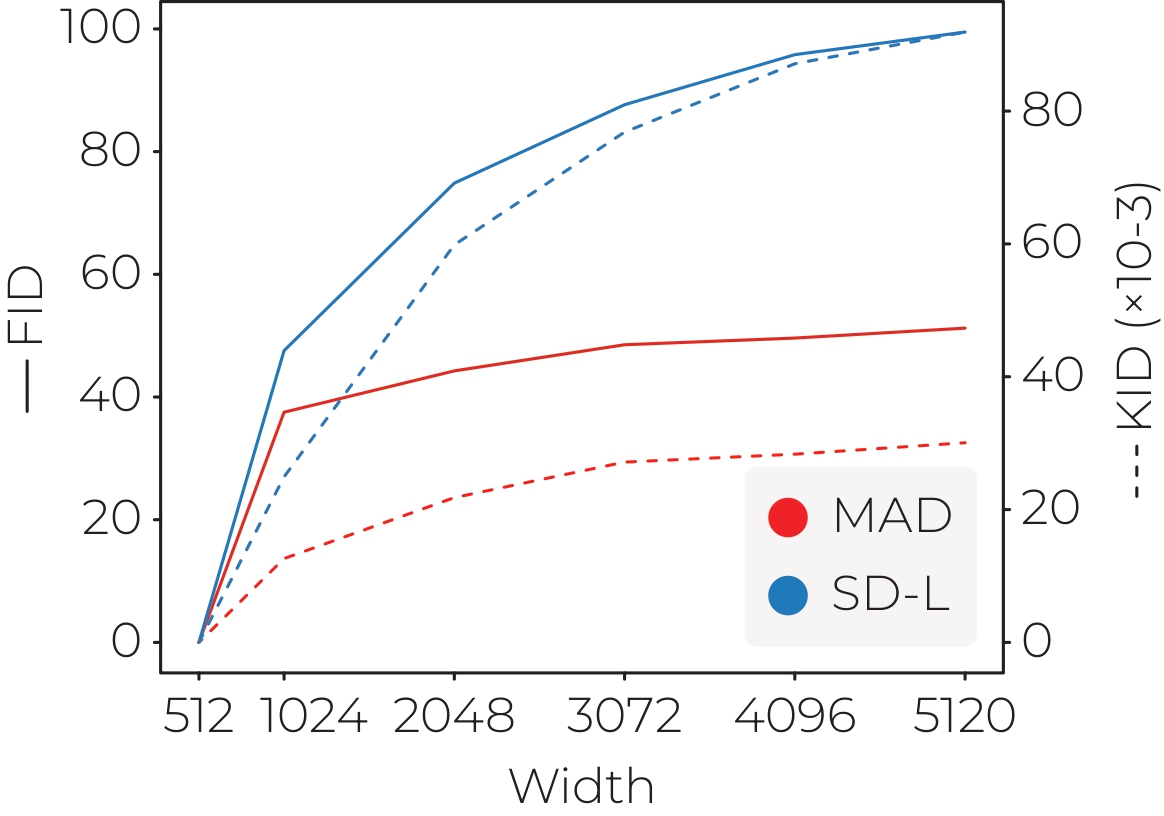}
    \vspace{-.6cm}
    \caption{Mean FID and KID on images of different widths by MAD and SD-L for the prompts used in the quantitative analysis. }\label{fig:sdlong_vs_mad}
\end{minipage}
\hfill
\begin{minipage}[b]{0.315\linewidth}
\centering
    \includegraphics[width=\linewidth]{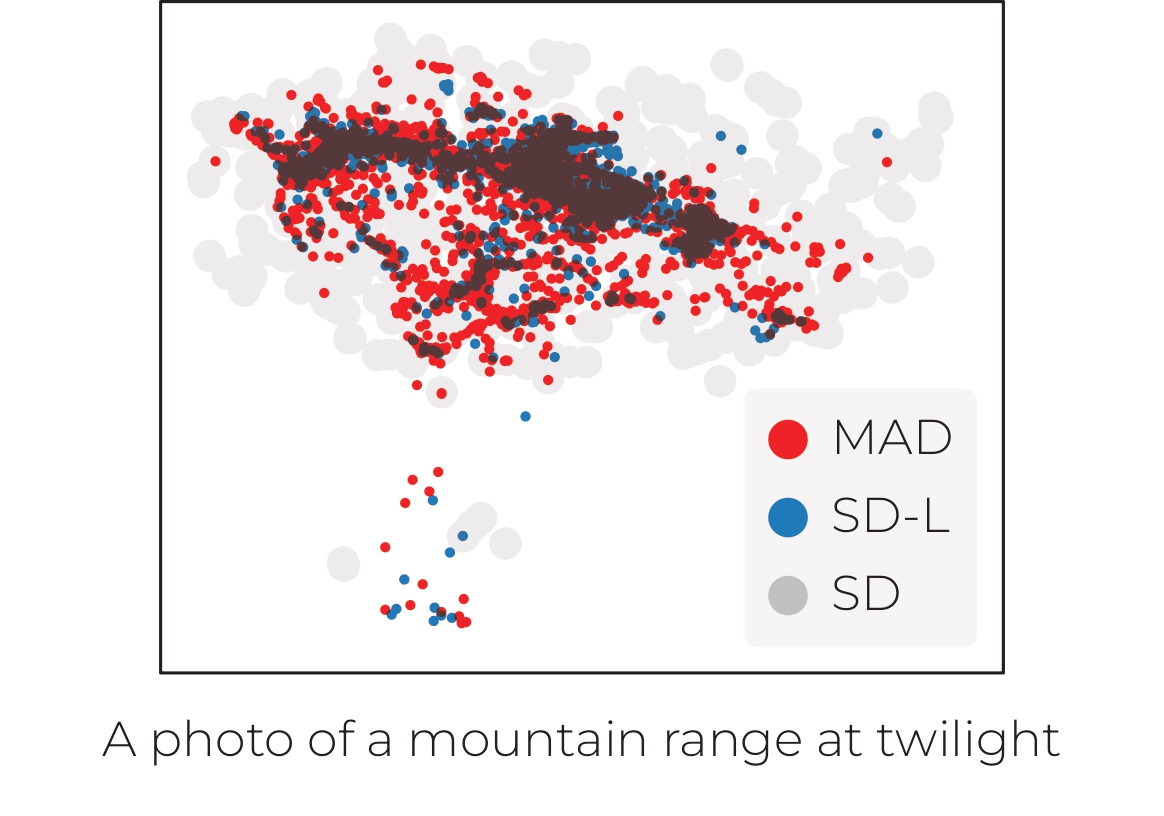}
    \vspace{-.6cm}
    \caption{T-SNE analysis of crops from images by MAD and SD-L \wrt~squared images from SD for the same prompt.}\label{fig:sdlong_vs_mad-tsne}
\end{minipage}
\hfill
\begin{minipage}[b]{0.315\linewidth}
\centering
    \includegraphics[width=\linewidth]{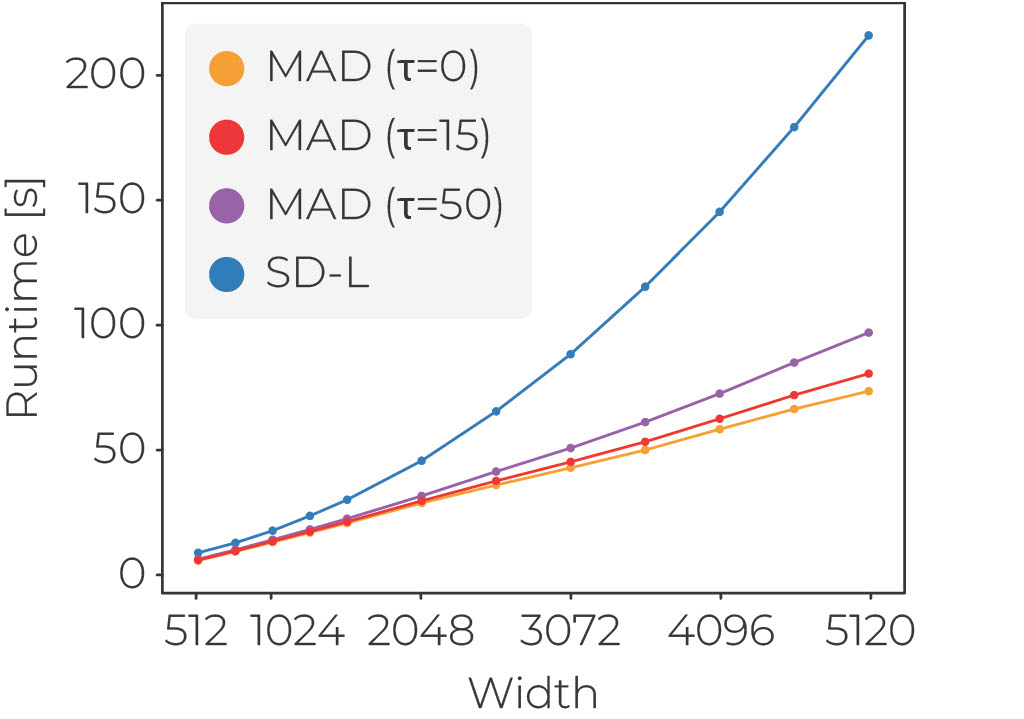}
    \vspace{-.6cm}
    \caption{Inference time at increasing output widths for SD-L and MAD: for $\tau{=}0$, MAD is not applied, $\tau{=}50$ is the upper bound.}\label{fig:runtimes}
\end{minipage}
\vspace{-1.5em}
\end{figure}

\tit{Joint Diffusion \textit{vs} Direct Inference}\label{sec:jont_vs_direct}
When generating long images at inference-time directly with models trained on squared images, a train-test discrepancy can occur, which is more evident as the desired output image width increases. This can be seen in~\cref{fig:sdlong_vs_mad}, where we compare the FID and KID on long image generation for MAD and SD-L. 
As the width increases, the generation performance of SD-L is increasingly impacted. This effect is reduced with our approach. 
Another issue with directly generating long images by simply applying a model trained on squared images is that this can lead to excessive, unrealistic uniformity in the final output and a lack of variability between images generated from the same prompt as their size increases~\cite{lee2023syncdiffusion}.
To explore this aspect, we run the t-SNE analysis~\cite{van2008visualizing} reported in~\cref{fig:sdlong_vs_mad-tsne} and in the supplementary. Specifically, we extract inception features from square images by SD and compute their t-SNE representations~\cite{van2008visualizing}. Then, we consider all the squared crops obtainable from 512${\times}$5120 images by SD-L and MAD and represent all their inception features in the same space using openTSNE~\cite{linderman2019fast, polivcar2019opentsne}. As we can see, images generated using MAD (represented by the red dots) cover a bigger portion of the SD distribution (the grey circles) compared to SD-L (the blue dots). Note that in~\cref{fig:sdlong_vs_mad-tsne}, the dark areas represent an overlap between MAD and SD-L.
The t-SNE analysis and the perceptual and semantic scores in~\cref{tab:quantitatives} suggest that MAD leads to both intra-image coherence and inter-image variability.\vspace{-.3em}

\tit{Runtime}
We also analyze the computation times of our approach and the competitors, all run on the same NVIDIA RTXA5000-24GB for generating 512${\times}$3072 images with the LDM backbone (reported in~\cref{tab:quantitatives}).
MAD runtime ranges from $\sim$43s to $\sim$50s, depending on the threshold $\tau$, which is comparable to MD (41.9s). The small computational overhead is due to the multiple merging and splitting operations and the larger tensors on which self- and cross-attention are performed. Compared to SyncD, which runs in 391.9${\pm}$0.1s, MAD is more than 8$\times$ faster for not requiring the expensive gradient computation. 
Moreover, in~\cref{fig:runtimes}, we report the runtime analysis at increasing output width \wrt~performing direct inference (SD-L) and never applying MAD in joint diffusion ($\tau$=0). Note that convolutions and attentions are linear \wrt~the output width when applied to split views and quadratic when applied to the merged latent. MAD entails convolutions on split views and attentions on merged latents only for $t<\tau$ (on splits afterward) and thus the computation grows slower than for SD-L.\vspace{-.2em}

\tit{Complex Scenes}
Moreover, we evaluate on the generation of 512$\times$3072 images from a set of 1000 prompts describing complex scenes, which we automatically obtain with ChatGPT-4 (referred to as GPT1k). We argue that such scenes allow to better highlight the challenges of the panorama generation task.
The results are reported in Table~\ref{tab:gpt1k} and highlight the superiority of our approach over the competitors. They also demonstrate that the task is far from being solved, and thus, we release the GPT1k set for fostering its exploration. Note that the qualitatives in this paper (excluding those in~\cref{fig:ldm_six_prompts}) are obtained with prompts in GPT1k. For more qualitative comparisons, see the supplementary.
\vspace{-.2em}

\begin{table}[t]
    \footnotesize
    \renewcommand{\arraystretch}{.95}
    \centering
    \setlength{\tabcolsep}{.2em}
    \caption{Quantitative comparison on 512$\times$3072 GPT-1k panorama generation using the LDM. I-StyleL, KID, and mGIQA values are scaled by 10$^{\text{3}}$.}
    \label{tab:gpt1k}\vspace{-.8em}
    \resizebox{.66\linewidth}{!}{
    \begin{tabular}{l c c c c c c c}
    \toprule
    & \textbf{mCLIP $\uparrow$} & \textbf{I-LPIPS $\downarrow$} & \textbf{I-StyleL $\downarrow$} & \textbf{FID $\downarrow$} & \textbf{KID $\downarrow$} & \textbf{mGIQA $\uparrow$}\\
    \midrule
    \textbf{SD}                & 32.45 & 0.67 & 12.76 & 49.32 & $<$0.01 & 12.98 \\
    \midrule
    \textbf{SD-L}              & 31.89 & 0.52 & ~0.73 & 68.03 & ~6.61 & 12.58 \\
    \textbf{MD}                & 32.46 & 0.63 & ~4.81 & 49.41 & ~0.42 & 13.22 \\
    \textbf{SyncD}             & 32.34 & 0.55 & ~4.24 & 53.52 & ~1.11 & 12.98 \\
    \textbf{MAD}               & 32.47 & 0.58 & ~3.89 & 54.44 & ~1.28 & 13.03 \\
    \bottomrule
    \end{tabular}}
\end{table}

\begin{figure}[t]
\begin{minipage}[b]{0.485\linewidth}
\centering
    \includegraphics[width=\linewidth]{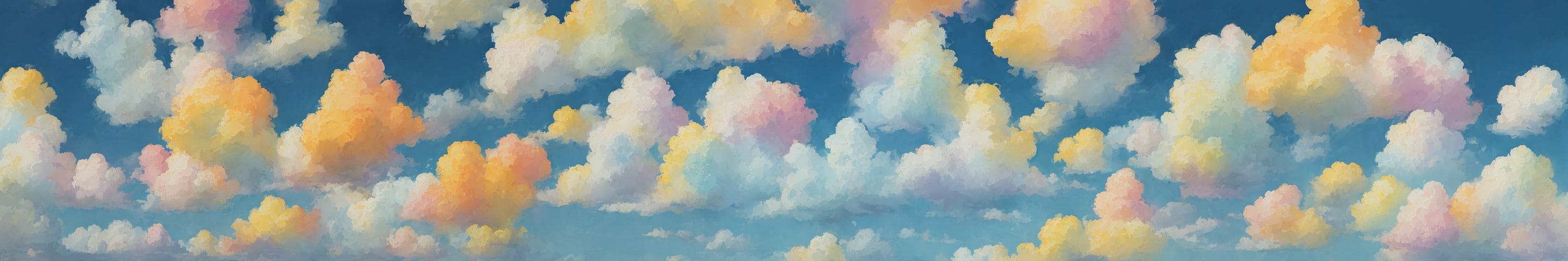} \\ 
    \includegraphics[width=\linewidth]{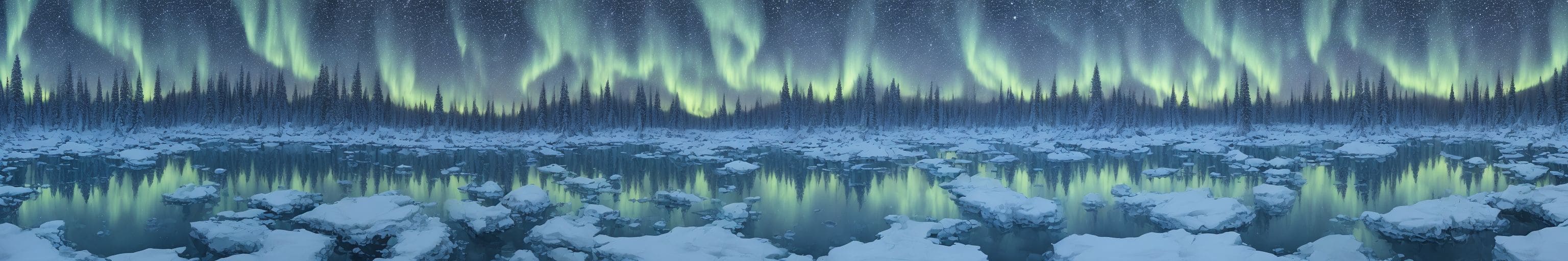} \\ 
    \includegraphics[width=\linewidth]{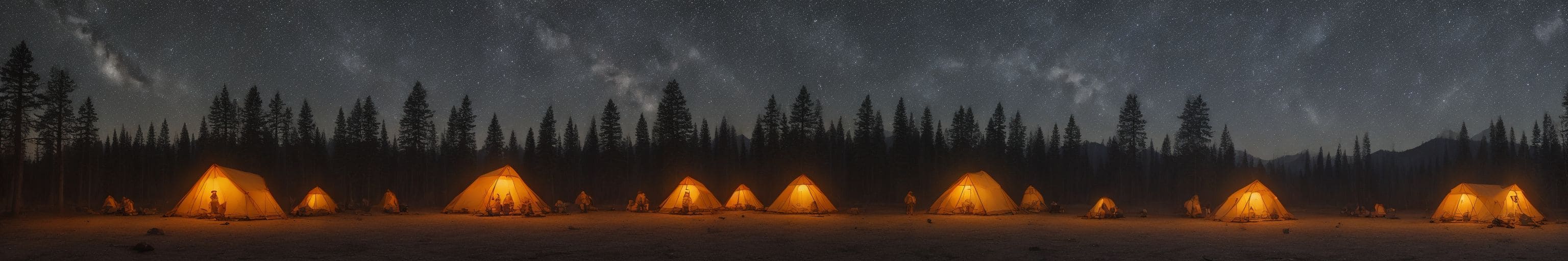} \\ 
    \includegraphics[width=\linewidth]{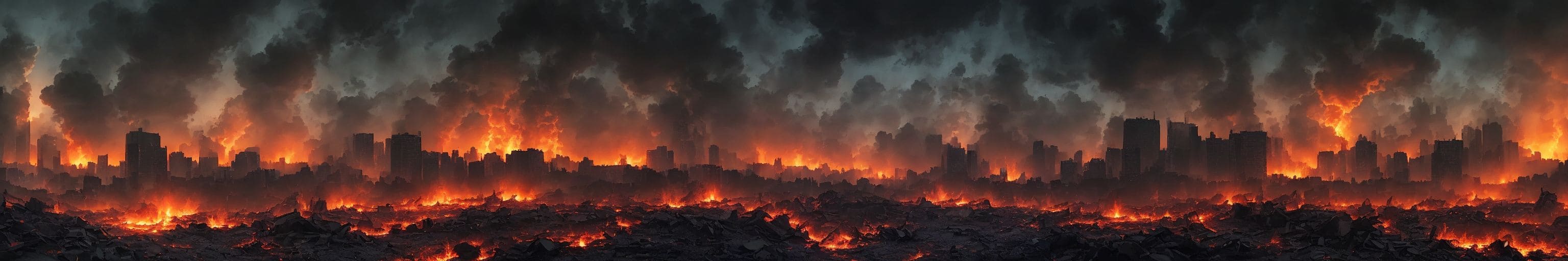} \\ 
    \includegraphics[width=\linewidth]{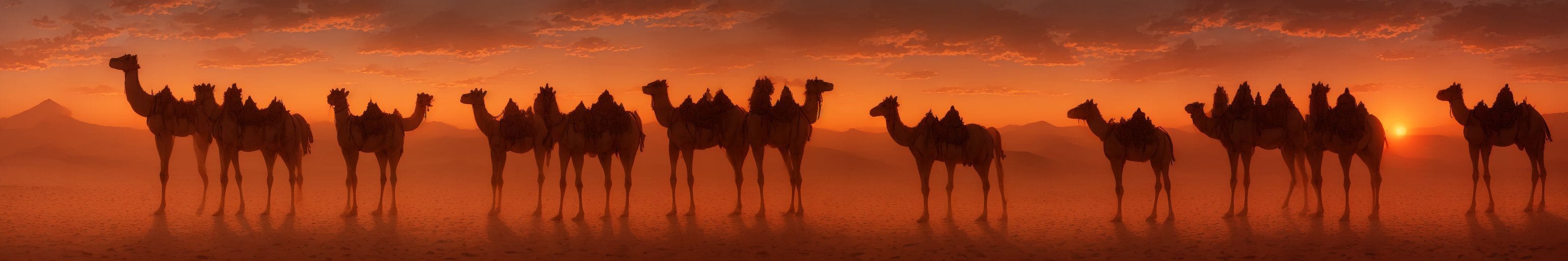} \\ 
    \includegraphics[width=\linewidth]{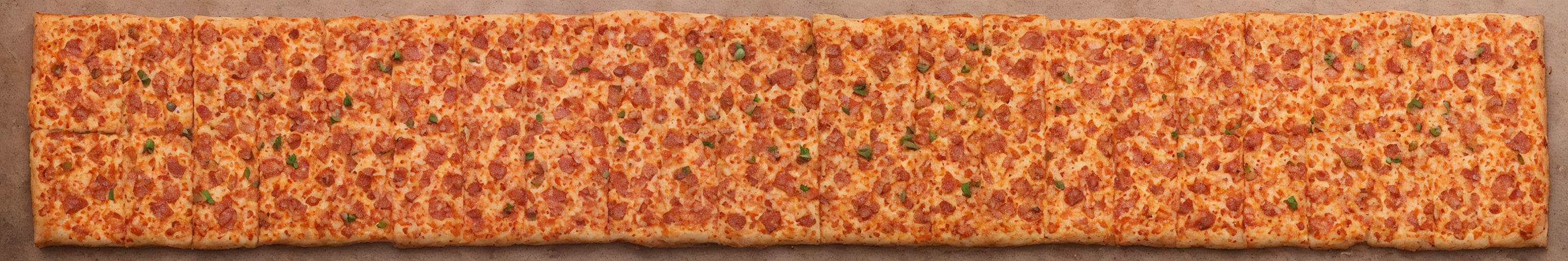} \\  \vspace{0.115cm}   
    \vspace{-.5em}
\caption{Qualitative results on different prompts of our approach by exploiting LCD as backbone.} \label{fig:qualitatives_ldm_lcm}
\end{minipage}
\hfill
\begin{minipage}[b]{0.485\linewidth}
    \centering
    \scriptsize
    \setlength{\tabcolsep}{0.1em}
    \begin{tabular}{cc c cc}
        \textbf{Baseline} & \textbf{MAD} && 
        \textbf{Baseline} & \textbf{MAD} \\
        \includegraphics[width=1.4cm]{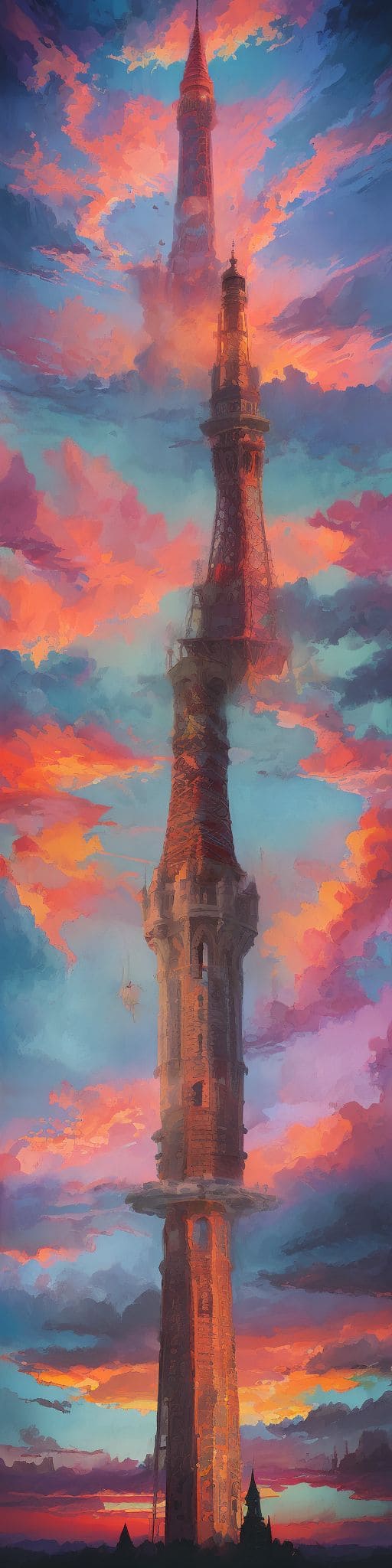} &
        \includegraphics[width=1.4cm]{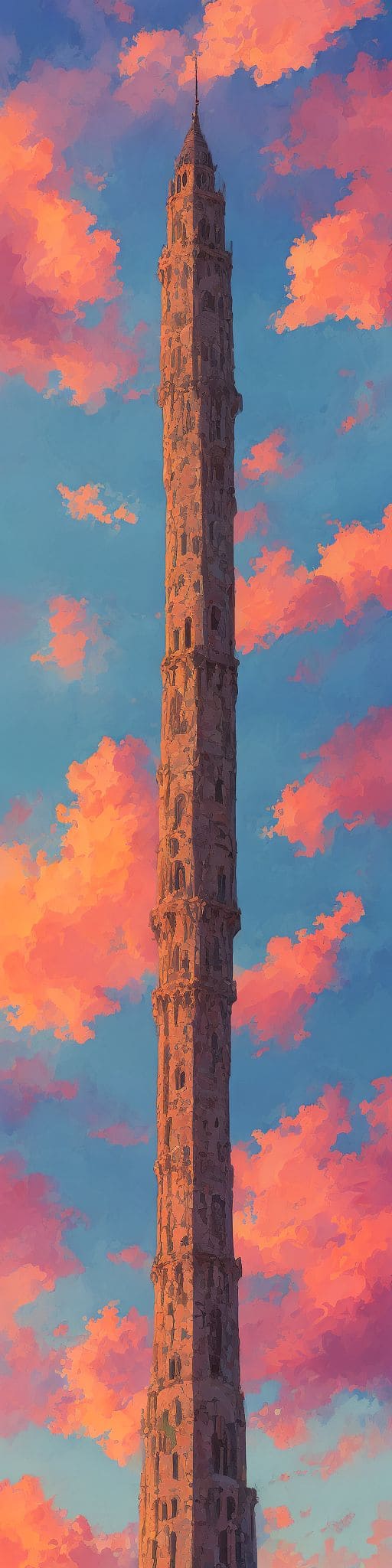} &&
        \includegraphics[width=1.4cm]{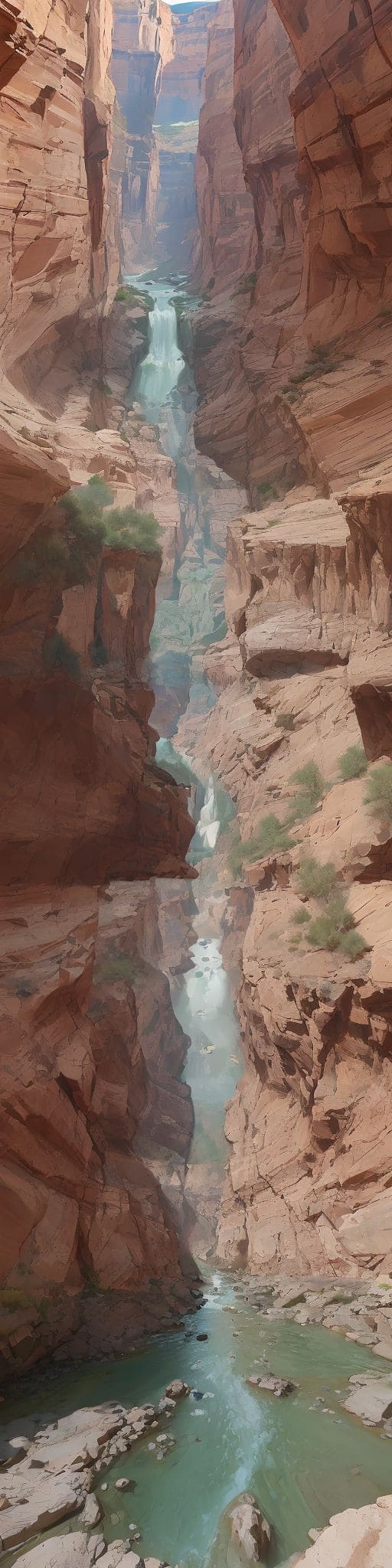} &
        \includegraphics[width=1.4cm]{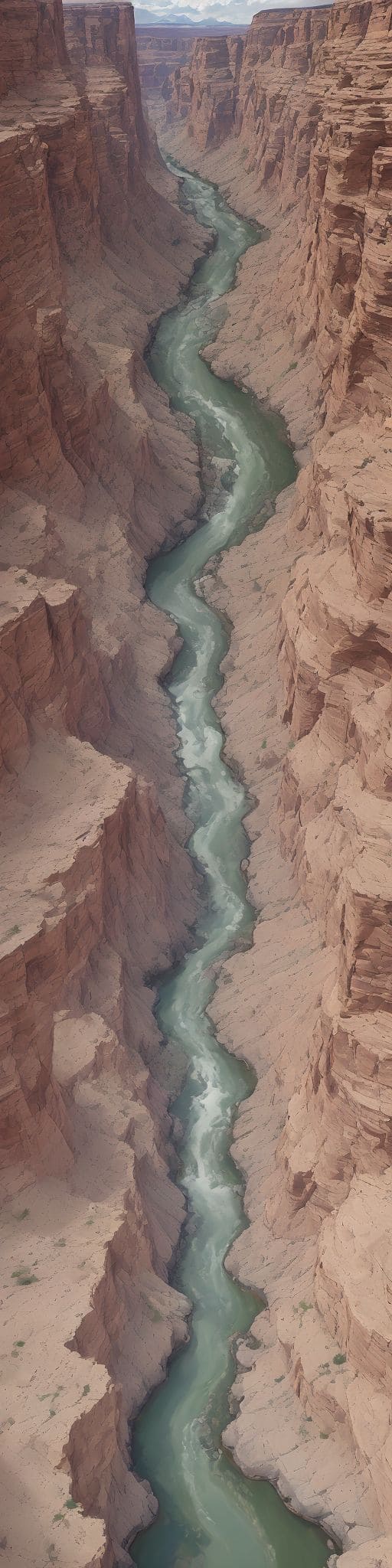} \\
        \multicolumn{2}{c}{\tiny \textit{A tower in a colorful sky}} &&
        \multicolumn{2}{c}{\tiny \textit{A river inside a canyon}} \\
    \end{tabular}
    \vspace{-.5em}
    \caption{\strut Qualitative comparison on vertical images with LCD. \textbf{Baseline} entails never applying MAD ($\tau{=}$0).    }
    \label{fig:qualitatives_vertical_lcm}
\end{minipage}
\vspace{-1.8em}
\end{figure}

\begin{table}[]
\renewcommand{\arraystretch}{.95}
\footnotesize
    \centering
    \setlength{\tabcolsep}{.28em}
    \caption{Quantitative comparison on 512$\times$3072 panorama generation with the LCM for different numbers of inference steps. I-StyleL, KID, and mGIQA are scaled by 10$^{\text{3}}$. }\vspace{-.8em}
    \resizebox{.9\linewidth}{!}{
    \label{tab:lcm_3072}\vspace{-2.8em}
    \begin{tabular}{lc c c c c c c }
        \toprule
        && \textbf{mCLIP $\uparrow$}& \textbf{I-LPIPS $\downarrow$} & \textbf{I-StyleL $\downarrow$} & \textbf{FID $\downarrow$} & \textbf{KID $\downarrow$} & \textbf{mGIQA $\uparrow$} \\
    \midrule
    \multicolumn{8}{c}{\textbf{1 Inference Step}}\\
    \midrule
    \textbf{LCD}              && 28.10${\pm}$1.65 & 0.42${\pm}$0.06 & 0.23${\pm}$0.12 & 26.95${\pm}$3.92~ & <0.01${\pm}$0.33~ & 42.52${\pm}$3.97~ \\
    \textbf{LCD-L}            && 29.50${\pm}$1.48 & 0.40${\pm}$0.08 & 0.21${\pm}$0.18 & 67.54${\pm}$15.53 & 66.86${\pm}$21.22 & 32.21${\pm}$4.50~ \\  
    \midrule
    \textbf{MAD ($\tau{=}$0)} && 28.42${\pm}$1.75 & 0.45${\pm}$0.06 & 0.25${\pm}$0.17 & 54.65${\pm}$24.64 & 50.56${\pm}$33.30 & 32.27${\pm}$4.38~ \\
    \textbf{MAD ($\tau{=}$1)} && 29.01${\pm}$1.66 & 0.40${\pm}$0.07 & 0.37${\pm}$0.28 & 72.43${\pm}$23.75 & 75.47${\pm}$33.24 & 32.23${\pm}$4.95~ \\
    \midrule
    \multicolumn{8}{c}{\textbf{2 Inference Steps}}\\
    \midrule
    \textbf{LCD}              && 30.57${\pm}$1.81 & 0.53${\pm}$0.06 & 1.43${\pm}$0.48 & 26.82${\pm}$11.51 & <0.01${\pm}$0.32~ & 44.30${\pm}$14.31 \\
    \textbf{LCD-L}            && 30.77${\pm}$2.09 & 0.47${\pm}$0.06 & 0.56${\pm}$0.29 & 49.98${\pm}$25.20 & 44.29${\pm}$26.88 & 35.45${\pm}$11.94 \\  
    \midrule
    \textbf{MAD ($\tau{=}$0)} && 30.75${\pm}$2.16 & 0.52${\pm}$0.05 & 0.84${\pm}$0.35 & 27.95${\pm}$12.07 & 13.39${\pm}$7.40~ & 35.78${\pm}$10.70 \\
    \textbf{MAD ($\tau{=}$1)} && 30.97${\pm}$2.15 & 0.50${\pm}$0.06 & 0.85${\pm}$0.34 & 35.69${\pm}$17.88 & 23.74${\pm}$15.82 & 35.32${\pm}$11.49 \\
    \textbf{MAD ($\tau{=}$2)} && 30.87${\pm}$2.18 & 0.47${\pm}$0.06 & 0.60${\pm}$0.23 & 52.88${\pm}$26.12 & 48.03${\pm}$27.75 & 35.48${\pm}$11.77 \\
    \midrule
    \multicolumn{8}{c}{\textbf{4 Inference Steps}}\\
    \midrule
    \textbf{LCD}              && 31.37${\pm}$1.60 & 0.55${\pm}$0.05 & 1.84${\pm}$0.81 & 29.05${\pm}$13.59 & <0.01${\pm}$0.23~ & 45.00${\pm}$12.97 \\  
    \textbf{LCD-L}            && 31.30${\pm}$1.63 & 0.50${\pm}$0.06 & 0.58${\pm}$0.24 & 55.52${\pm}$32.82 & 51.71${\pm}$37.79 & 35.44${\pm}$12.94 \\  
    \midrule
    \textbf{MAD ($\tau{=}$0)} && 31.36${\pm}$1.83 & 0.55${\pm}$0.05 & 0.90${\pm}$0.35 & 31.35${\pm}$15.42 & 13.27${\pm}$10.05 & 35.44${\pm}$12.38 \\
    \textbf{MAD ($\tau{=}$2)} && 31.48${\pm}$1.87 & 0.52${\pm}$0.05 & 0.71${\pm}$0.26 & 38.77${\pm}$19.85 & 27.41${\pm}$16.19 & 35.23${\pm}$13.15 \\
    \textbf{MAD ($\tau{=}$4)} && 31.41${\pm}$1.94 & 0.48${\pm}$0.06 & 0.46${\pm}$0.19 & 62.82${\pm}$36.24 & 61.69${\pm}$42.07 & 35.06${\pm}$13.61 \\
    \bottomrule
    \end{tabular}
    }
\end{table} 

\tit{User Study} 
Considering the sometimes subjective differences in image quality and the partial nature of quantitative evaluations, we also conduct a user study to evaluate the coherence of the generated images. Participants, presented with pairs of images generated with our approach, MD, and SyncD, were asked which was most coherent and best matched the textual prompt. 
Every user answered six control questions to assess their understanding of the task. We collected responses from 201 users and kept those with at least five correct answers to the control questions (139 users). These provided 7807 answers, from which it has emerged that MAD was preferred 80.92\% of the times over MD and 66.54\% over SyncD (note that SyncD was preferred over MD 68.19\% of the times, which is in line with what found in~\cite{lee2023syncdiffusion}). For further details, please see the supplementary.

\tit{Results with LCMs}
Finally, we apply our method with different $\tau$ to Latent Consistency Dreamshaper (LCD) at varying generation steps with resolution 512$\times$3072. 
The quantitative results are in~\cref{tab:lcm_3072}, where we also report the performance obtained by applying LCD to generate the whole panorama image directly (referred to as LCD-L) and the scores on the squared images from LCD, for reference. Applying MAD improves the mCLIP, generating more coherent images (see also the qualitatives in~\cref{fig:qualitatives_ldm_lcm,fig:qualitatives_vertical_lcm} and in the supplementary).

\section{Conclusions and Discussion}
\label{sec:conclusions}
In this work, we have presented the \textbf{M}erge-\textbf{A}ttend-\textbf{D}iffuse operator, which can be applied to the attention layers of a pretrained convolutional-attentive diffusion model for zero-shot generation of long images that are both perceptually and semantically coherent. We have conducted extensive experimental analysis, whose results demonstrate the effectiveness of the proposed approach especially in terms of visual quality and adherence to the input prompt.

\tit{Limitations} Similar to other inference-time joint diffusion approaches~\cite{bar2023multidiffusion, lee2023syncdiffusion}, a limitation of our method is that it relies on the diffusion paths of the base model for realistic results and can be hindered by a bad seed or inappropriate text prompts. Performing the attention operations on the merged tensor mitigates this issue. For example, in~\cref{fig:mountain_comparison_method,fig:ldm_mad_blocks,fig:ablation_threshold}, the incoherent horizon lines of independent diffusion paths (in the top images) are corrected by applying MAD. Further discussion can be found in the supplementary.
\tit{Potential Negative Societal Impacts} Image generative models can produce harmful content like deepfakes and NSFW images, violate copyrights, and reflect existing societal biases. Further research is necessary to address these issues. 

\section*{Acknowledgement}
This work was supported by the ``AI for Digital Humanities'' project funded by ``Fondazione di Modena'' and the PNRR project Italian Strengthening of ESFRI RI Resilience (ITSERR) funded by the European Union – NextGenerationEU.

\clearpage
\bibliographystyle{splncs04}
\bibliography{main}

\title{Supplementary Material for:\\Merging and Splitting Diffusion Paths for Semantically Coherent Panoramas}

\author{Fabio Quattrini\orcidlink{0009-0004-3244-6186} \and Vittorio Pippi\orcidlink{0009-0001-7365-6348} \and \\ Silvia Cascianelli\orcidlink{0000-0001-7885-6050} \and Rita Cucchiara\orcidlink{0000-0002-2239-283X}}

\authorrunning{F.~Quattrini et al.}
\titlerunning{Suppl.: Merging and Splitting Diffusion Paths for Coherent Panoramas}

\institute{University of Modena and Reggio Emilia, Modena, Italy\\
\email{\{name.surname\}@unimore.it}}

\maketitle

In this document, we report additional results of our proposed MAD approach for inference-time adaptation of diffusion models for generating long images.
We explore the effect of changing the application stage and varying the number of application steps, present more detailed information regarding the user study, and offer further comparisons with SD-L. Additionally, we report other qualitative results, also at different aspect ratios, and further discuss the limitations. 

\section{Effect of the Application Stage}
In~\Cref{tab:quantitatives_ldm_blocks,tab:quantitatives_lcm_blocks}, we report a quantitative analysis of the effects of applying MAD at different stages in the noise prediction model of the considered LDM and LCM. We observe that when MAD is applied only in the bottleneck (Mid blocks), FID, KID, and mGIQA values are comparable to those obtained with the baseline in which MAD is never applied and the perceptual coherence is lower than in the other settings. On the other hand, I-LPIPIS, I-StyleL, and mCLIP get better when MAD is applied more (\ie~in all blocks) and in later stages of the U-Net (\ie~in the upsampling blocks). The same trend can be observed from the LDM qualitatives in~\cref{fig:ablation_blocks_sup}.
    
\section{Effect of the Number of Application Steps} 
We report a quantitative analysis (\cref{tab:ablation_threshold_quantitative}) and a more exhaustive qualitative analysis (\cref{fig:ablation_threshold_qualitative}) on the effect of the number of inference timesteps in which the proposed MAD operator is applied (defined by the hyperparameter $\tau$). In this experiment, we use the considered LDM run for 50 reverse process steps and apply MAD at each timestep from the first one up to a certain threshold. As we can see, 15 steps lead to a good trade-off between variety and uniformity.

\section{Details of the User Study} 
Here, we report further details on the user study conducted to assess human preference between images generated with our approach or with two other State-of-the-Art ones (namely, MultiDiffusion~\cite{bar2023multidiffusion} and SyncDiffusion~\cite{lee2023syncdiffusion}). For each approach, we generated 60 images for each of the six prompts used for the quantitative evaluation (for a total of 360 images per method). We then formed pairs of images with matching prompts and presented 60 random pairs to the users (plus 6 additional pairs as a vigilance task), asking them to select the image that they found to be ``\textit{the most coherent, and that conforms the most to the prompt}''. As mentioned in the main paper, 139 out of 201 participants passed the vigilance task and provided a total of 7807 preference answers. In~\cref{fig:user_study}, we plot the overall and the per-prompt results. As we can see, our method is consistently preferred over the competitors.

\section{Further comparison with Direct Inference}
\begin{figure*}[t]
    \centering
    \scriptsize
    \includegraphics[width=0.9\linewidth]{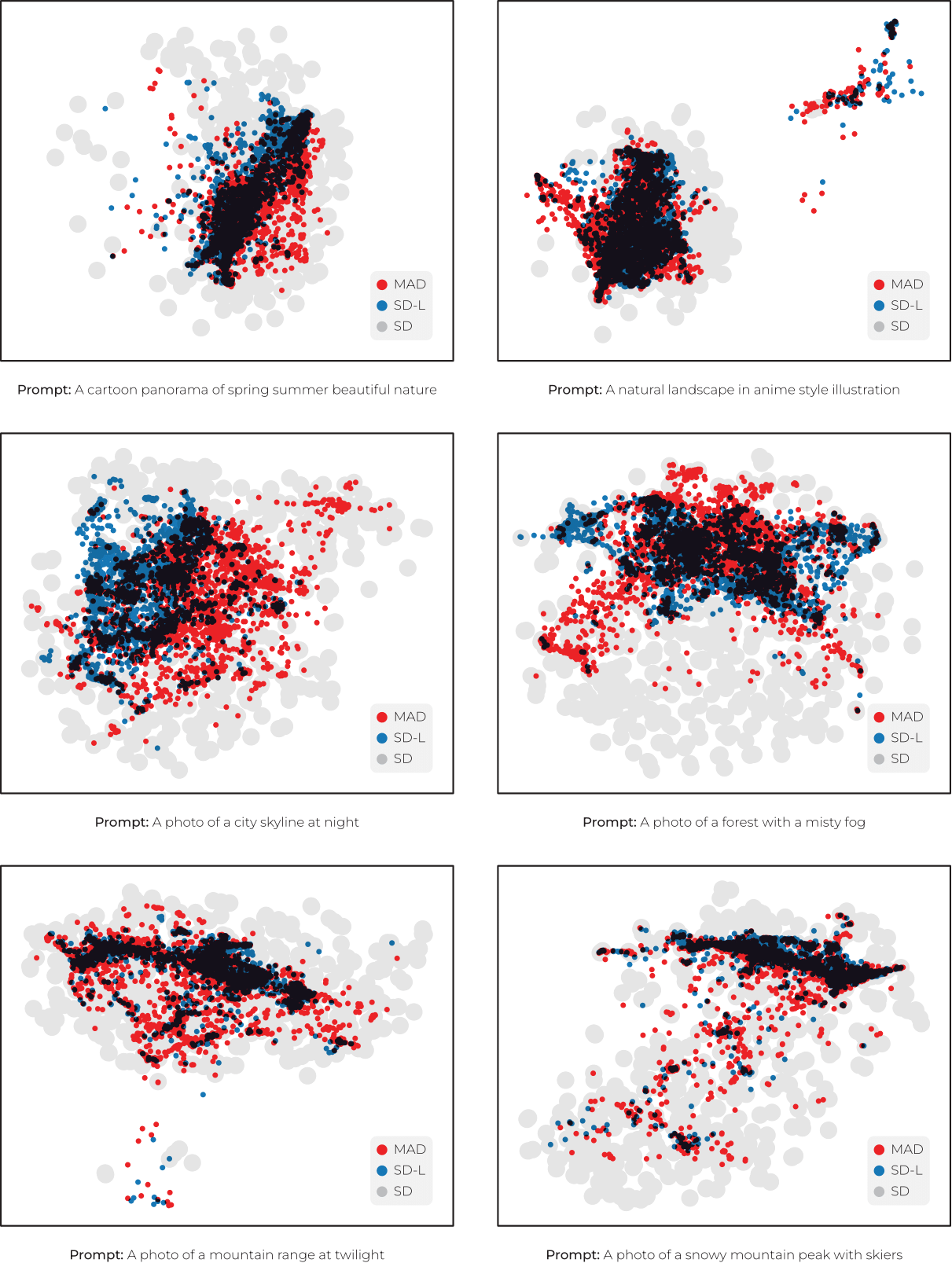}
\caption{Comparison on the distribution coverage of panoramas generated by MAD (red) and SD-L (blue) with respect to square images generated by SD (gray). Darker areas indicate an overlap between the embeddings of MAD and SD-L images.}
\label{fig:tsne_all_six_prompts}
\end{figure*}

Here, we report further comparison on the variability between the generated images using MAD and SD-L. In particular, for each one of the six evaluation prompts, we exploit the same network used for computing the FID, KID, and mGIQA (an Inception-v3 trained on Imagenet) to extract features from square images generated with SD and embed them using t-SNE~\cite{van2008visualizing}. Then, we generate 512$\times$5120 images with SD-L and MAD and embed all the squared crops using openTSNE~\cite{linderman2019fast, polivcar2019opentsne}. In~\cref{fig:tsne_all_six_prompts}, we report the embeddings, showing that images generated by MAD have more variability with respect to the images generated by the vanilla SD. Thus, MAD has increased inter-image variability compared to SD-L. For each prompt, we plot the SD embeddings using grey circles, the MAD embeddings using red dots, and the SD-L embeddings using blue dots. Darker areas indicate an overlap between the embeddings of MAD and SD-L images.

\section{Combination with Attention Scaling}
To adapt Stable Diffusion to the generation of larger images, a recent work has proposed to re-scale the attention operations in the U-Net~\cite{jin2024training} (which we refer to as Attn-S). We evaluate this approach both in combination with a method that generates the whole image directly and MAD (we refer to this combination as MAD-L+Attn-S). To this end, we consider the generation of 512$\times$3072 images.
The result of generation with LDMs and LCMs are reported in~\Cref{tab:ldm_attns} and \Cref{tab:lcm_attns}, respectively. It can be noticed that the combination with the Attn-S has limited to no benefit in terms of generation performance. This can be explained by looking at the small differences in terms of qualitative results, reported in~\Cref{fig:ldm_attns,fig:ldm_attns}.
    
\section{Further Results at Different Aspect Ratios}
\begin{figure*}[t]
    \centering
    \scriptsize
    \includegraphics[width=0.92\linewidth]{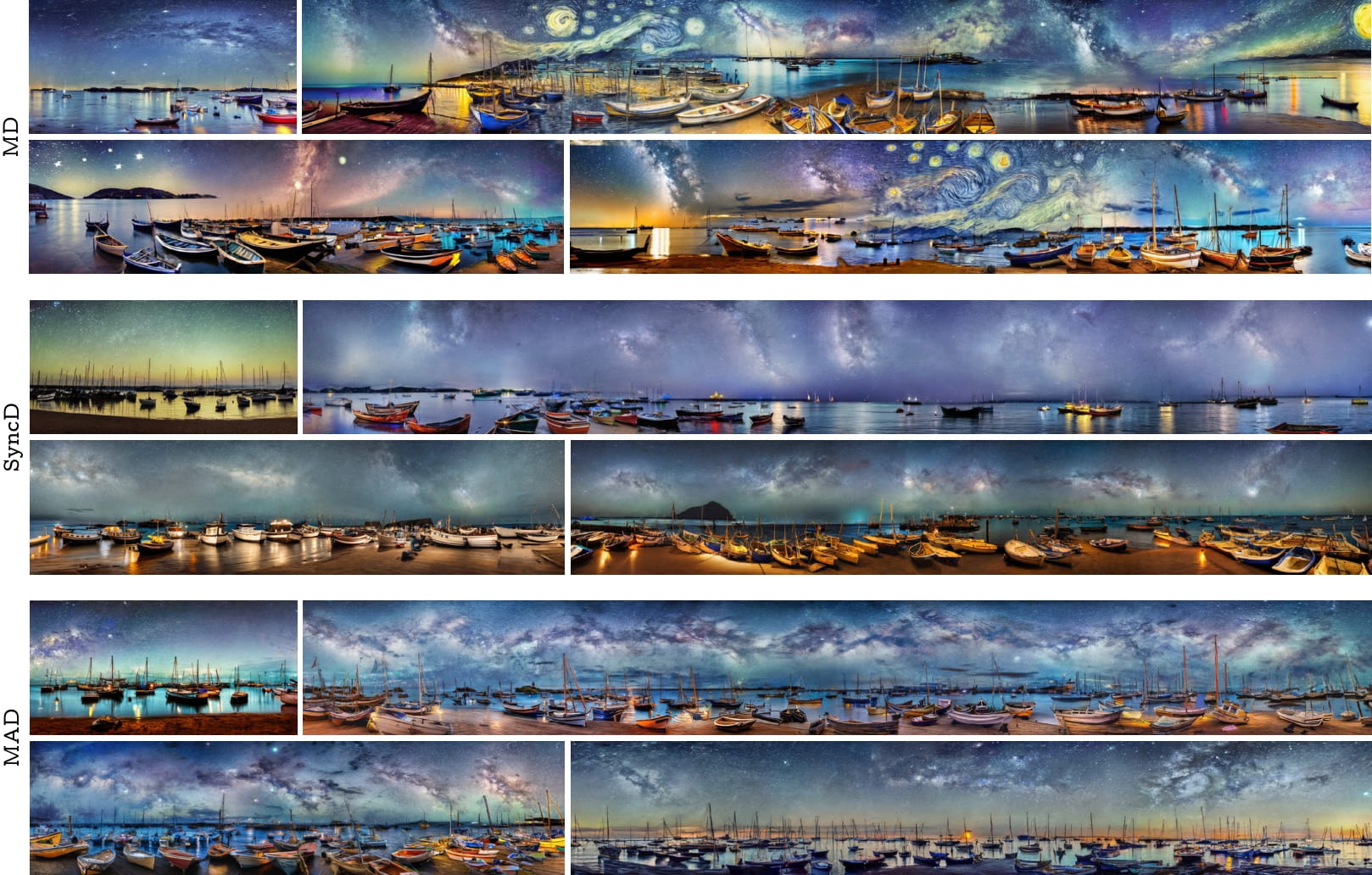}
\caption{Qualitative comparisons with MD and SyncD on the generation of horizontal images at different aspect ratios with the LDM model and the prompt \textit{A photo of the seaside with some boats in a starry night}. These images are 512$\times$1024 (top-left), 512$\times$4096 (top-right), 512$\times$2048 (bottom-left), and 512$\times$3072 (bottom-right).}
\label{fig:supp_aspect_ratios_h_ldm}
\end{figure*}

\begin{figure*}[t]
    \centering
    \scriptsize
    \includegraphics[width=0.92\linewidth]{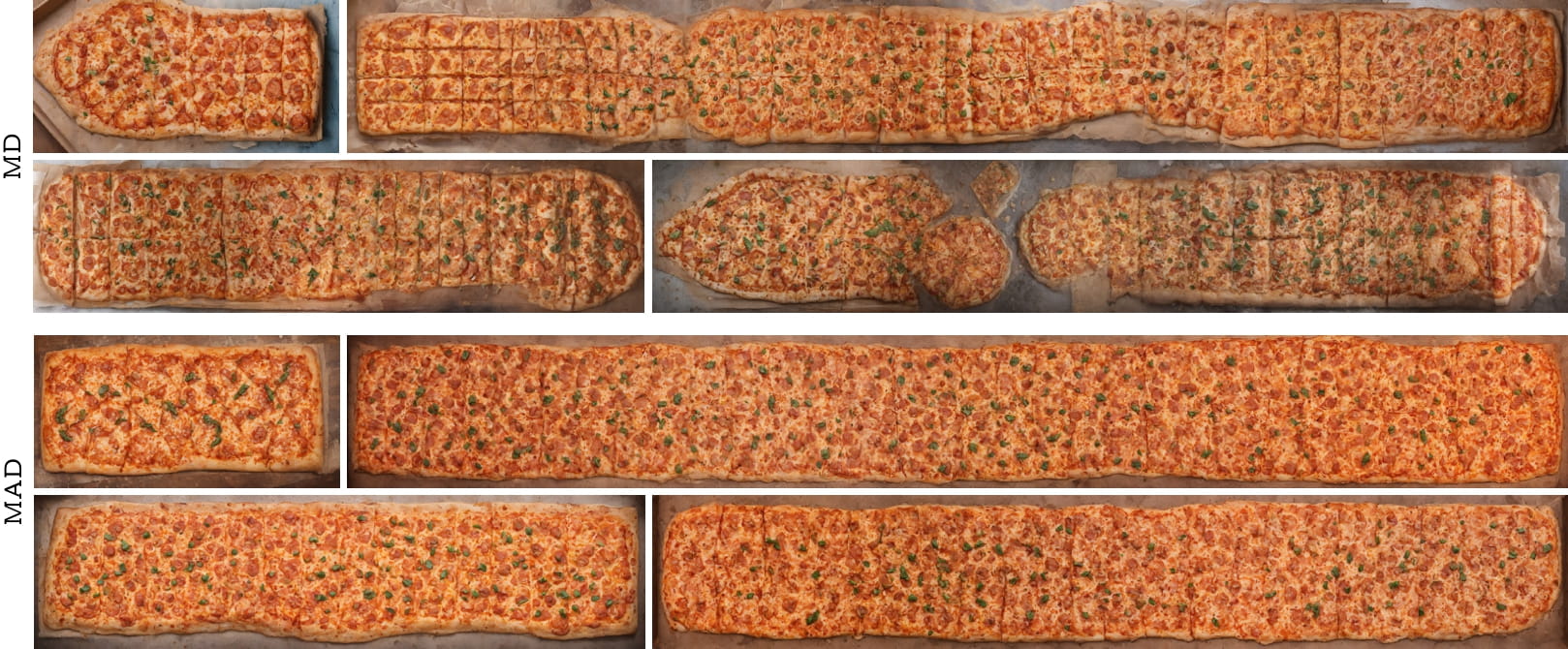}
\caption{Qualitative comparisons with respect to a baseline on the generation of horizontal images at different aspect ratios with the LCM model and the prompt \textit{Top-view of a squared long pizza}. The images are 512$\times$1024 (top-left), 512$\times$4096 (top-right), 512$\times$2048 (bottom-left), and 512$\times$3072 (bottom-right).}
\label{fig:supp_aspect_ratios_h_lcm}
\end{figure*}

\begin{figure*}[t]
    \centering
    \scriptsize
    \includegraphics[height=0.92\linewidth]{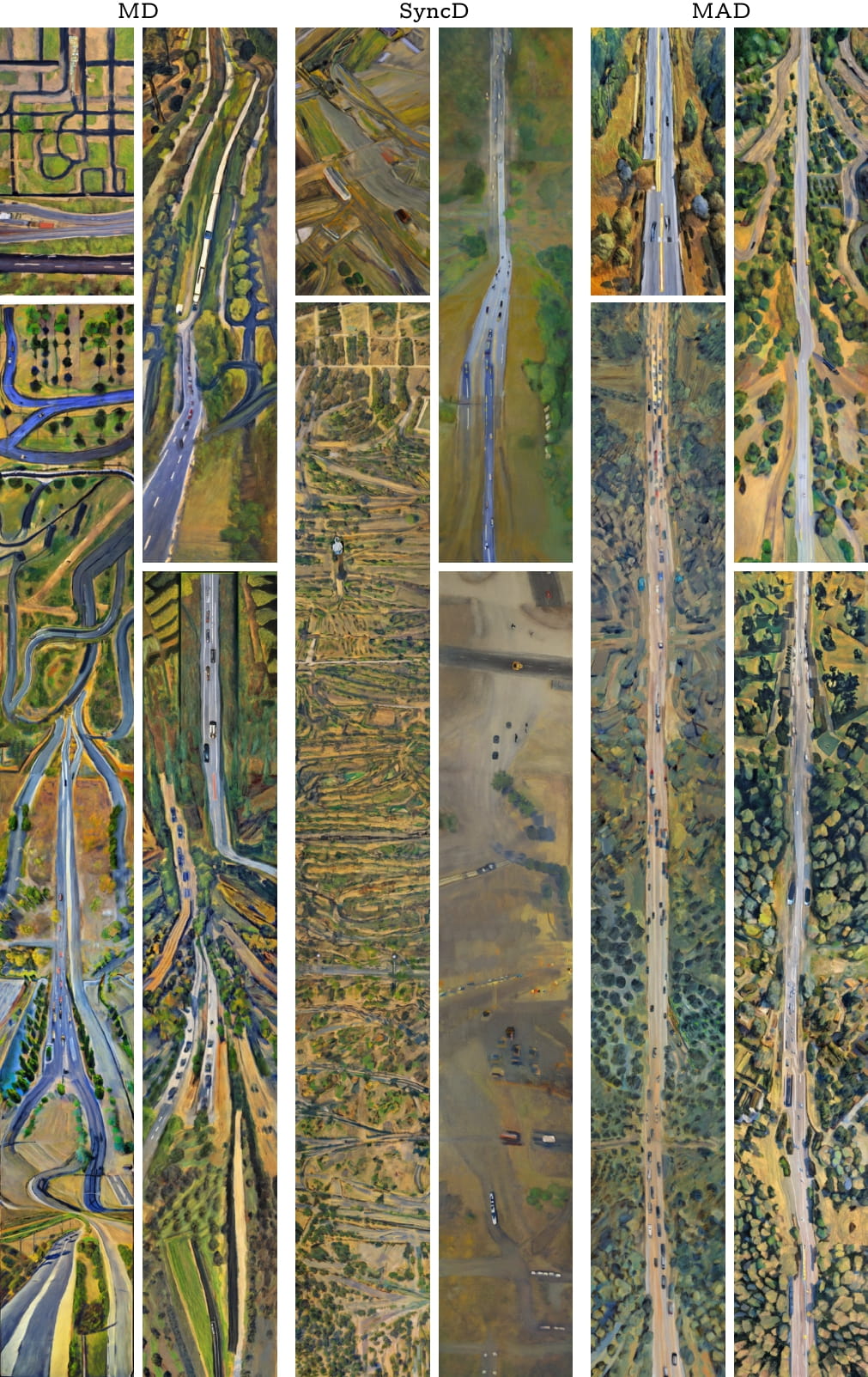}
\caption{Qualitative comparisons with MD and SyncD on the generation of vertical images at different aspect ratios with the LDM model and the prompt \textit{Top view of a long road with the Manet style}. The images are 1024$\times$512 (top-left), 2048$\times$512 (top-right), 4096$\times$512 (bottom-left), and 3072$\times$512 (bottom-right).}
\label{fig:supp_aspect_ratios_v}
\end{figure*}

\begin{figure*}[t]
    \centering
    \scriptsize
    \includegraphics[height=0.92\linewidth]{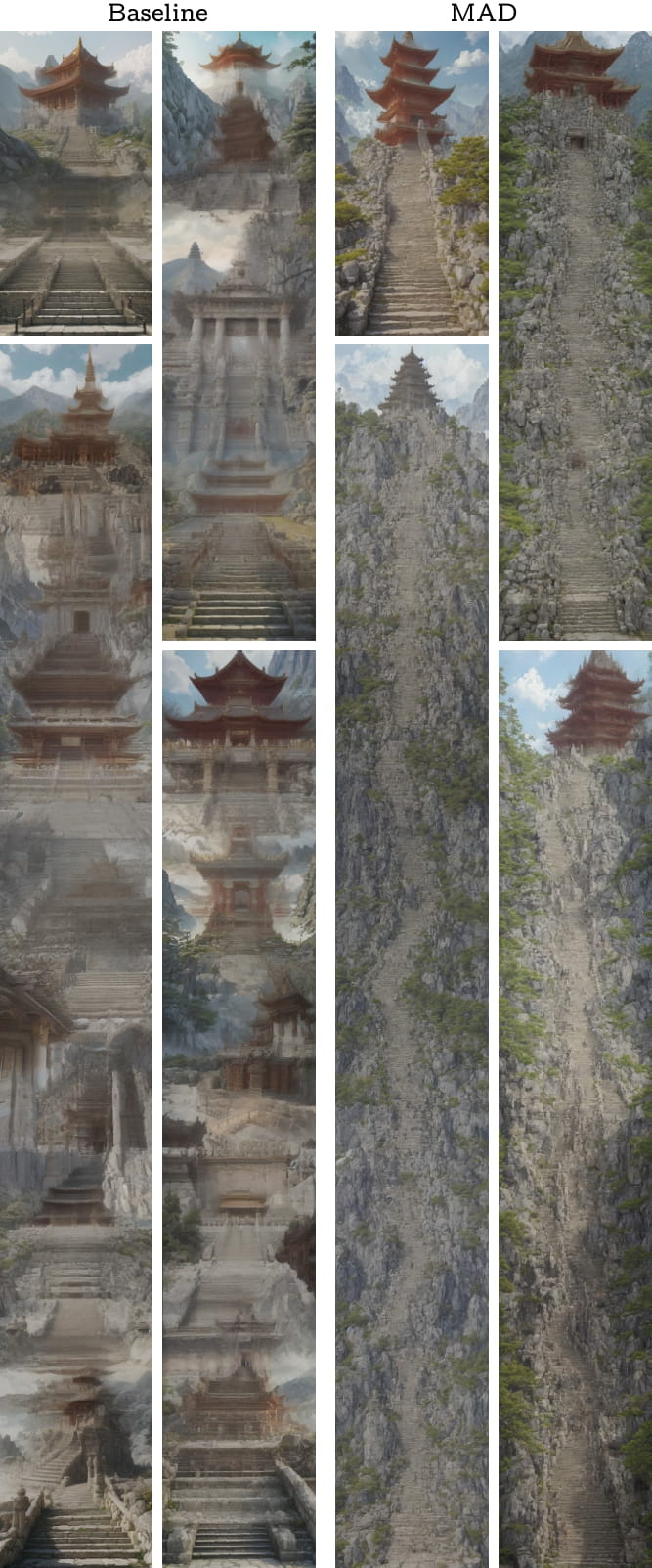}
\caption{Qualitative comparisons with respect to a baseline on the generation of vertical images at different aspect ratios with the LCM model and the prompt \textit{Stairs on a mountain with a temple on top}. The images are 1024$\times$512 (top-left), 2048$\times$512 (top-right), 4096$\times$512 (bottom-left), and 3072$\times$512 (bottom-right).}
\label{fig:supp_aspect_ratios_v_lcm}
\end{figure*}

We report a qualitative comparison with the State-of-the-Art approaches (MultiDiffusion~\cite{bar2023multidiffusion} and SyncDiffusion~\cite{lee2023syncdiffusion}) on the generation of images with different aspect ratios. In particular, we consider the generation of horizontal images with the LDM (\cref{fig:supp_aspect_ratios_h_ldm}) and with the LCM (\cref{fig:supp_aspect_ratios_h_lcm}) and of vertical images with the same models (\cref{fig:supp_aspect_ratios_v} and~\cref{fig:supp_aspect_ratios_v_lcm}, respectively).

\section{Further Qualitative Results}
\begin{figure*}[ht]
\centering
\scriptsize
    \begin{tabular}{m{0.75em}c}
    \arrayrulecolor{white}
         \rotatebox[origin=l]{90}{\textbf{SD-L}} & \makecell{\includegraphics[width=0.8\linewidth, height=0.135\linewidth]{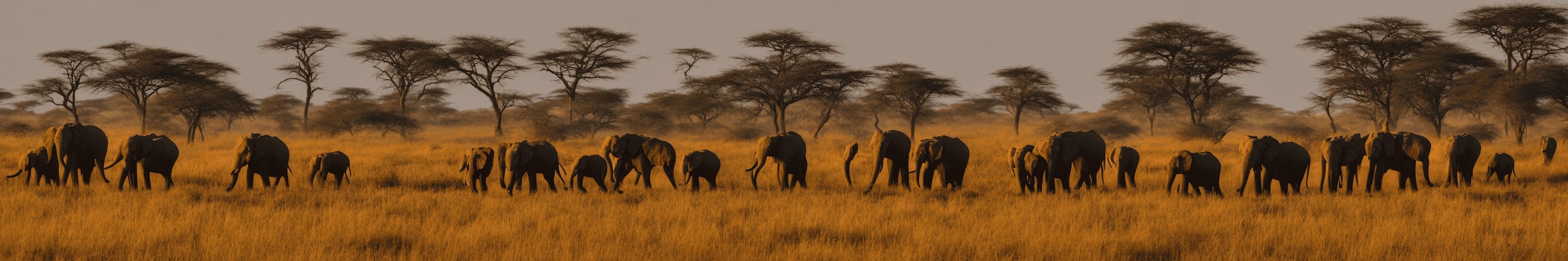}  }\\
         \rotatebox[origin=l]{90}{\textbf{MD   }} & \makecell{\includegraphics[width=0.8\linewidth, height=0.135\linewidth]{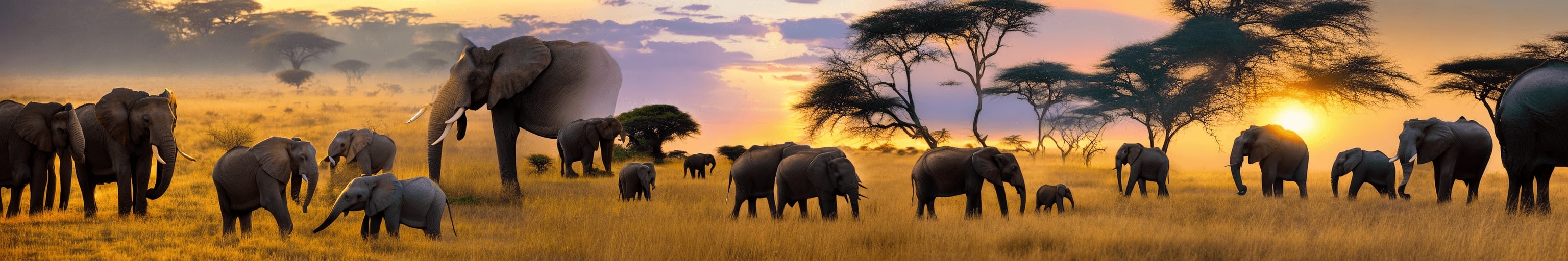}  }\\
         \rotatebox[origin=l]{90}{\textbf{SyncD}} & \makecell{\includegraphics[width=0.8\linewidth, height=0.135\linewidth]{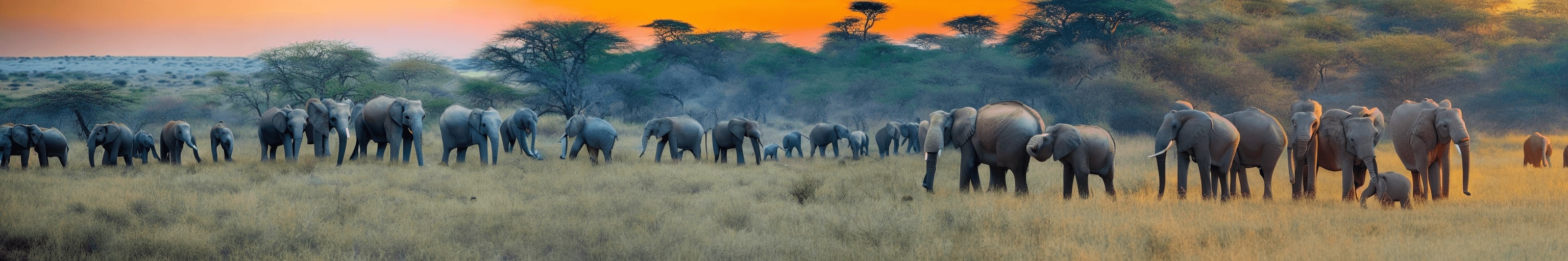}  }\\
         \rotatebox[origin=l]{90}{\textbf{MAD }} & \makecell{\includegraphics[width=0.8\linewidth, height=0.135\linewidth]{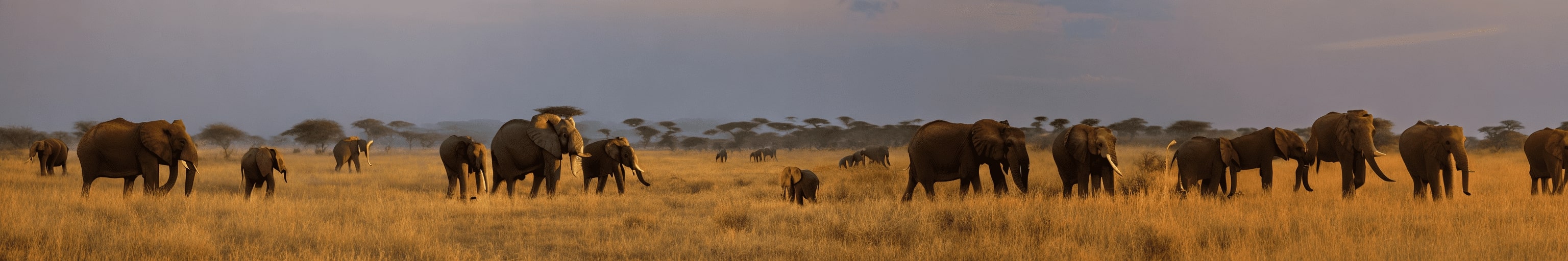}  }\\
    \end{tabular}
\caption{Qualitative comparisons with respect to SD-L, MD, and SyncD for the prompt \textit{A group of elephants grazing on the Savannah at sunset} from GPT1k, using the LDM model.}
\label{fig:qualitative_supp_gpt1k_1}
\end{figure*}

\begin{figure*}[ht]
\centering
\scriptsize
    \begin{tabular}{m{0.75em}c}
    \arrayrulecolor{white}
         \rotatebox[origin=l]{90}{\textbf{SD-L}} & \makecell{\includegraphics[width=0.8\linewidth, height=0.135\linewidth]{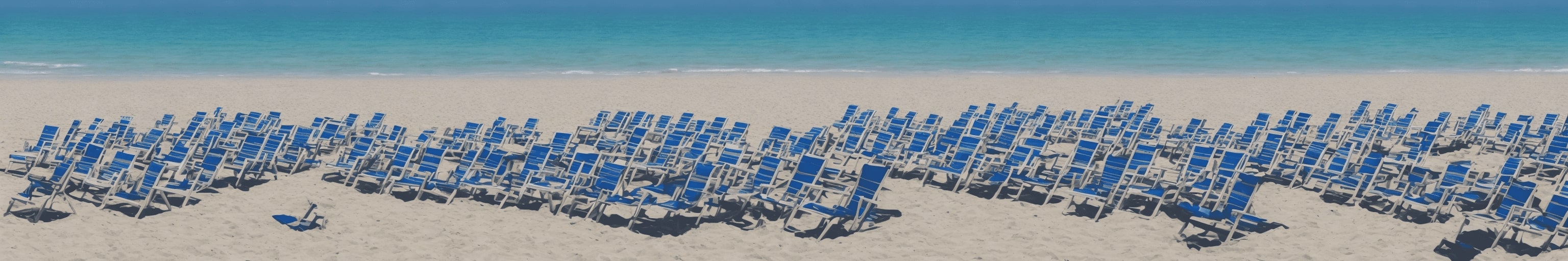}  }\\
         \rotatebox[origin=l]{90}{\textbf{MD   }} & \makecell{\includegraphics[width=0.8\linewidth, height=0.135\linewidth]{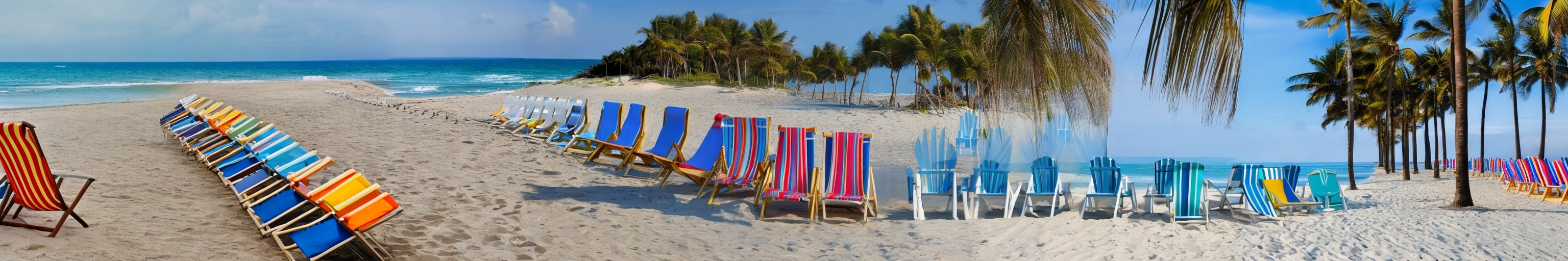}  }\\
         \rotatebox[origin=l]{90}{\textbf{SyncD}} & \makecell{\includegraphics[width=0.8\linewidth, height=0.135\linewidth]{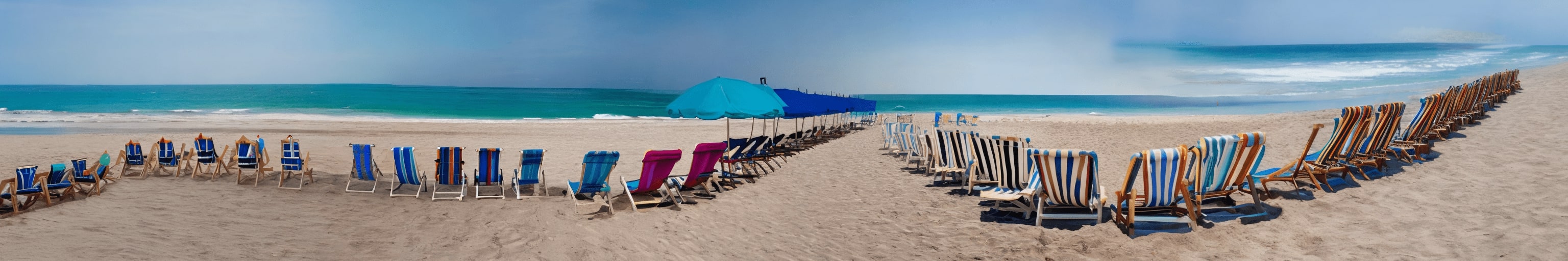}  }\\
         \rotatebox[origin=l]{90}{\textbf{MAD }} & \makecell{\includegraphics[width=0.8\linewidth, height=0.135\linewidth]{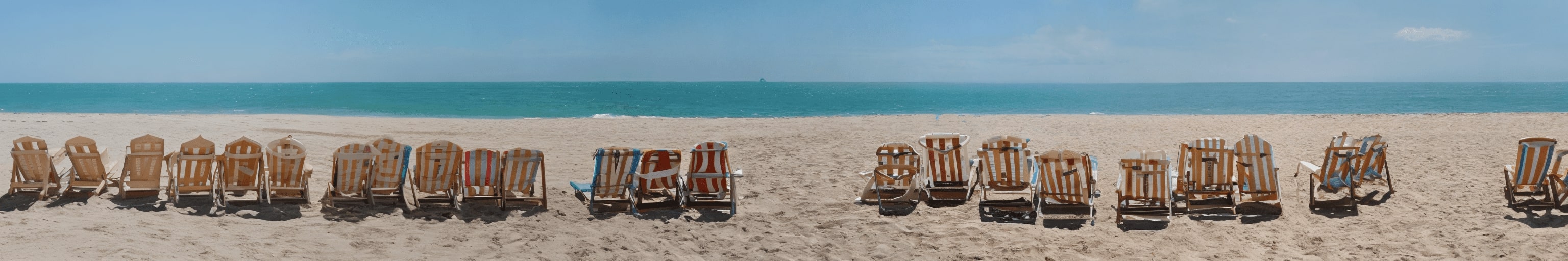}  }\\
    \end{tabular}
\caption{Qualitative comparisons with respect to SD-L, MD, and SyncD for the prompt \textit{A line of beach chairs facing the ocean} from GPT1k, using the LDM model.}
\label{fig:qualitative_supp_gpt1k_2}
\end{figure*}

\begin{figure*}[ht]
\centering
\scriptsize
         \begin{tabular}{m{0.75em}c}
     \arrayrulecolor{white}
          \rotatebox[origin=l]{90}{\textbf{LCD-L   }} & \makecell{\includegraphics[width=0.8\linewidth, height=0.135\linewidth]{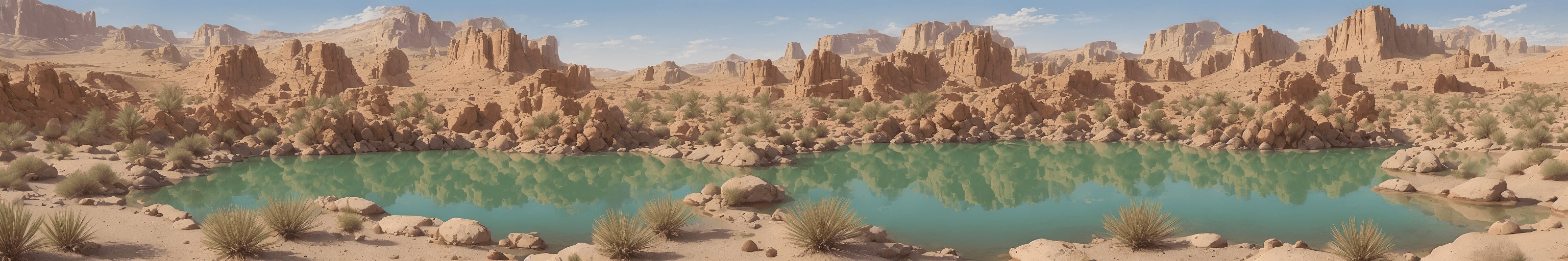}  }\\
          \rotatebox[origin=l]{90}{\textbf{Baseline}} & \makecell{\includegraphics[width=0.8\linewidth, height=0.135\linewidth]{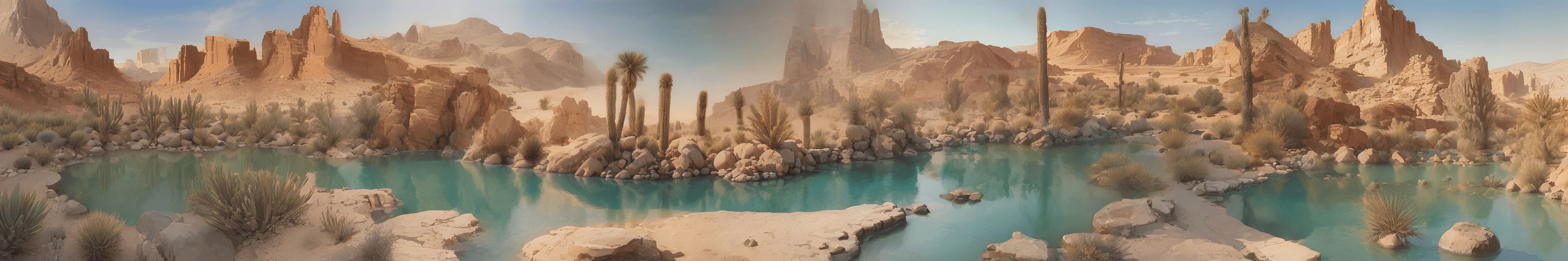}  }\\
          \rotatebox[origin=l]{90}{\textbf{MAD     }} & \makecell{\includegraphics[width=0.8\linewidth, height=0.135\linewidth]{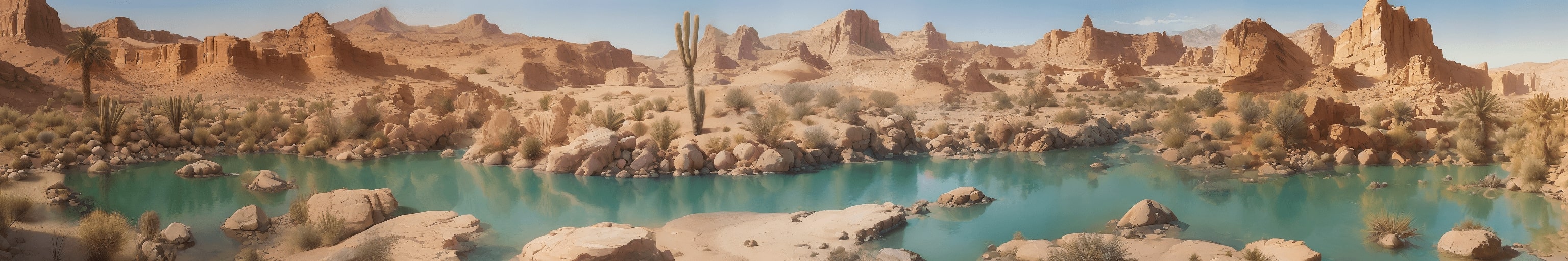}  }\\
     \end{tabular}
 \caption{Qualitative comparisons with respect to LCD-L, MD, and SyncD for the prompt \textit{A tranquil desert oasis} from GPT1k, using the LCM model.}
\label{fig:qualitative_supp_gpt1k_lcm_1}
\end{figure*}

\begin{figure*}[ht]
\centering
\scriptsize
    \begin{tabular}{m{0.75em}c}
    \arrayrulecolor{white}
         \rotatebox[origin=l]{90}{\textbf{LCD-L   }} & \makecell{\includegraphics[width=0.8\linewidth, height=0.135\linewidth]{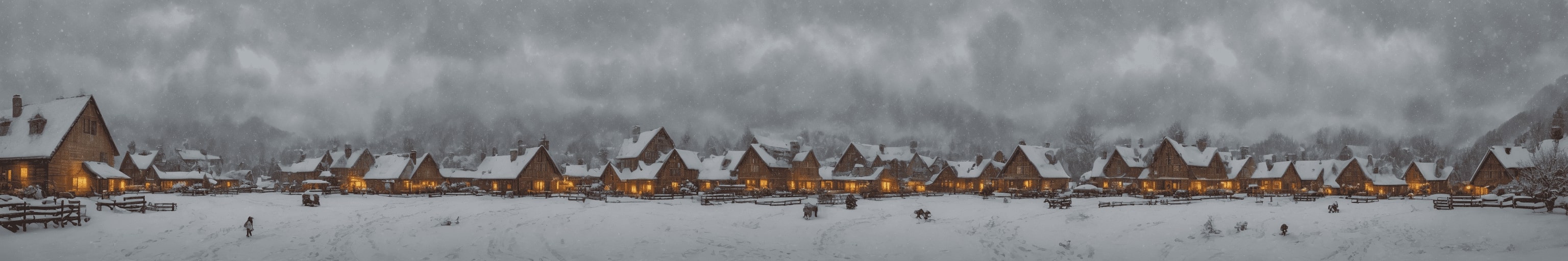}  }\\
         \rotatebox[origin=l]{90}{\textbf{Baseline}} & \makecell{\includegraphics[width=0.8\linewidth, height=0.135\linewidth]{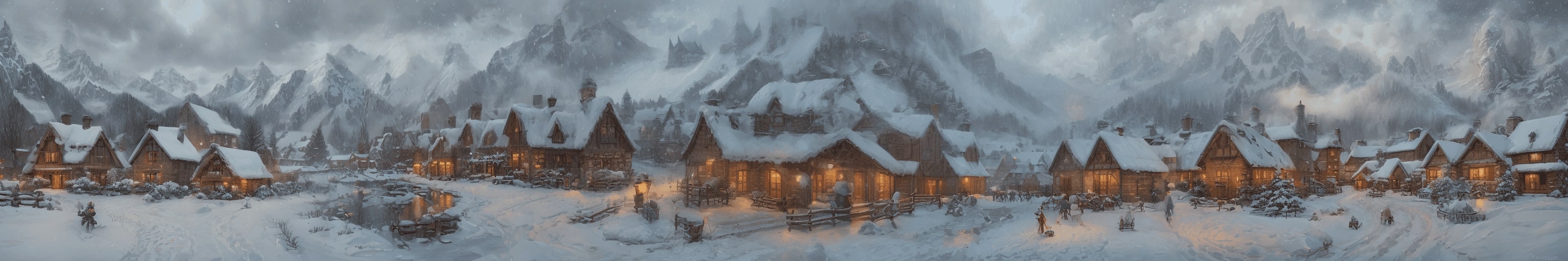}  }\\
         \rotatebox[origin=l]{90}{\textbf{MAD     }} & \makecell{\includegraphics[width=0.8\linewidth, height=0.135\linewidth]{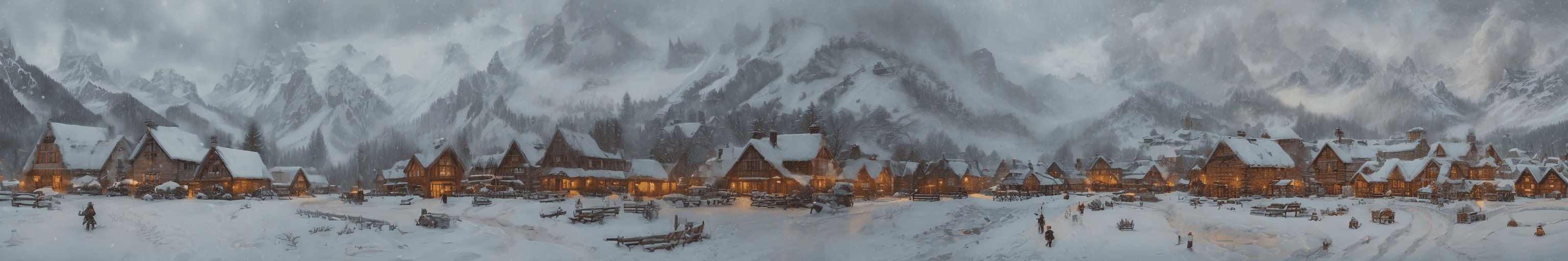}  }\\
    \end{tabular}
\caption{Qualitative comparisons with respect to LCD-L and a baseline for the prompt \textit{A snowstorm blanketing a small village} from GPT1k, using the LCM model.}
\label{fig:qualitative_supp_gpt1k_lcm_2}
\end{figure*}

\begin{figure*}[ht]
\centering
\scriptsize
    \includegraphics[width=0.8\linewidth]{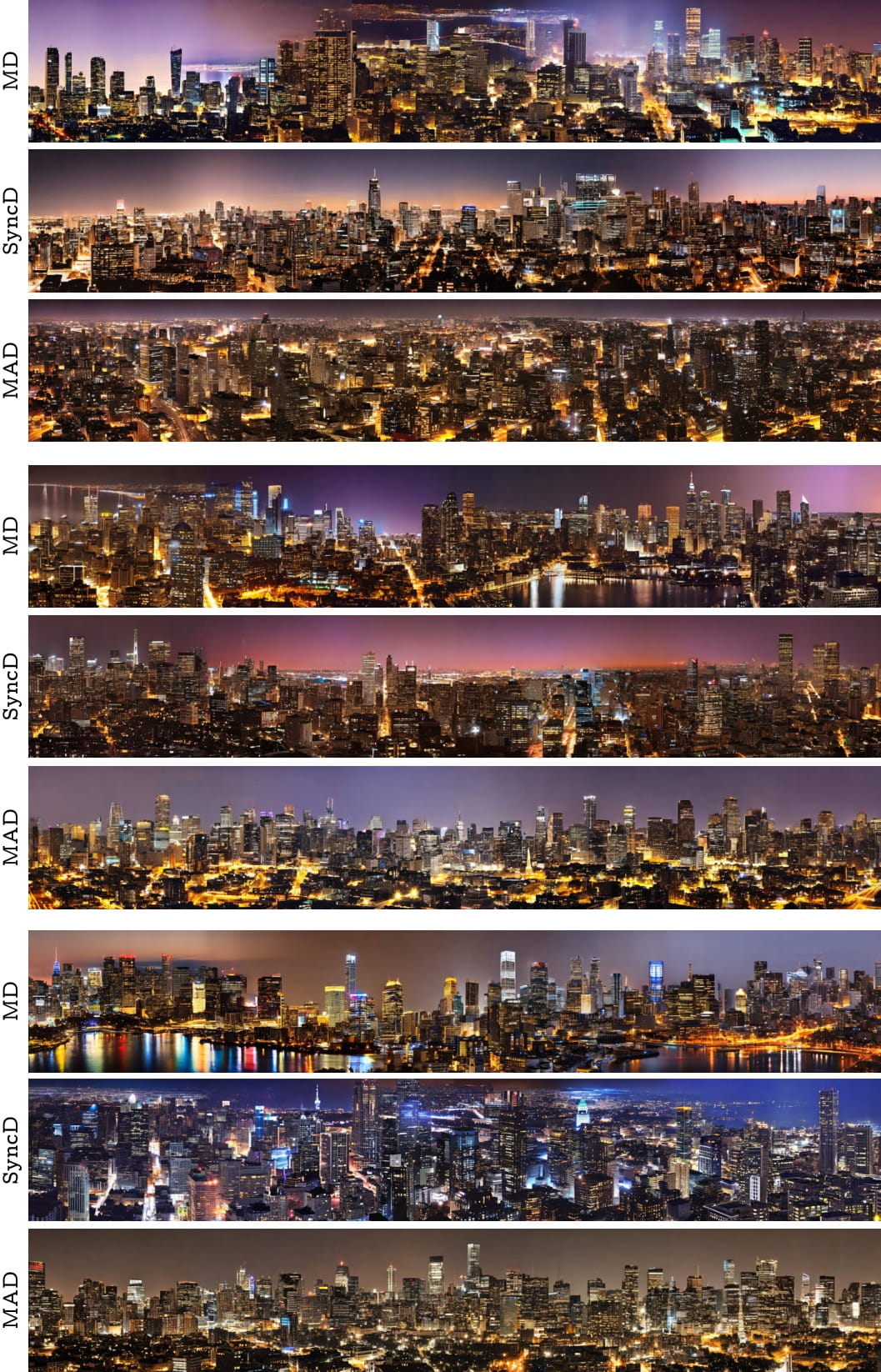}
\caption{Qualitative comparisons with respect to MD and SyncD for the prompt \textit{A photo of a city skyline at night}, using the LDM model.}
\label{fig:qualitative_supp_city}
\end{figure*}

\begin{figure*}[ht]
\centering
\scriptsize
 \includegraphics[width=0.8\linewidth]{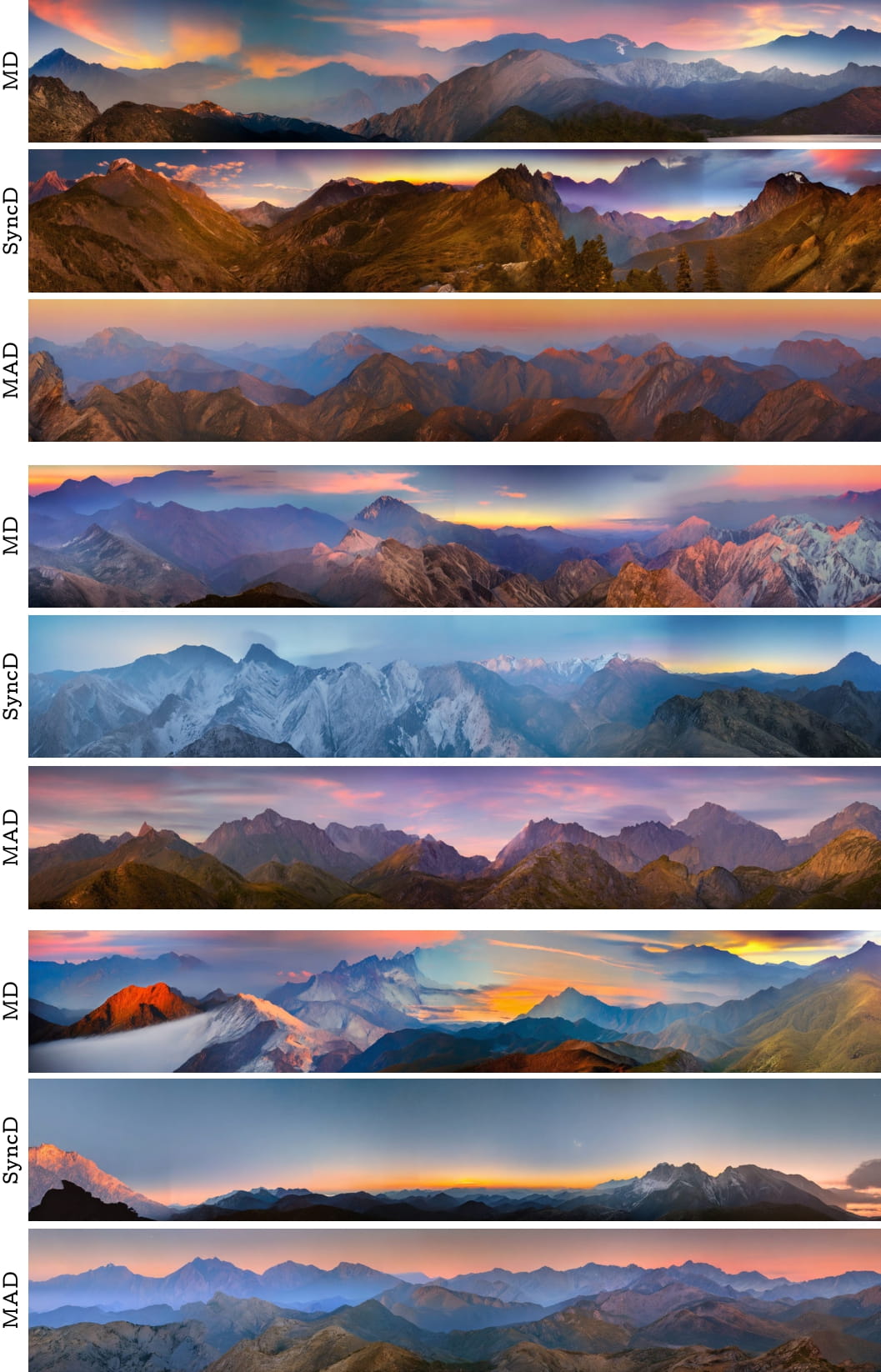}
\caption{Qualitative comparisons with respect to MD and SyncD for the prompt \textit{A photo of a mountain range at twilight}, using the LDM model.}
\label{fig:qualitative_supp_mount}
\end{figure*}

\begin{figure*}[ht]
\centering
\scriptsize
\includegraphics[width=0.8\linewidth]{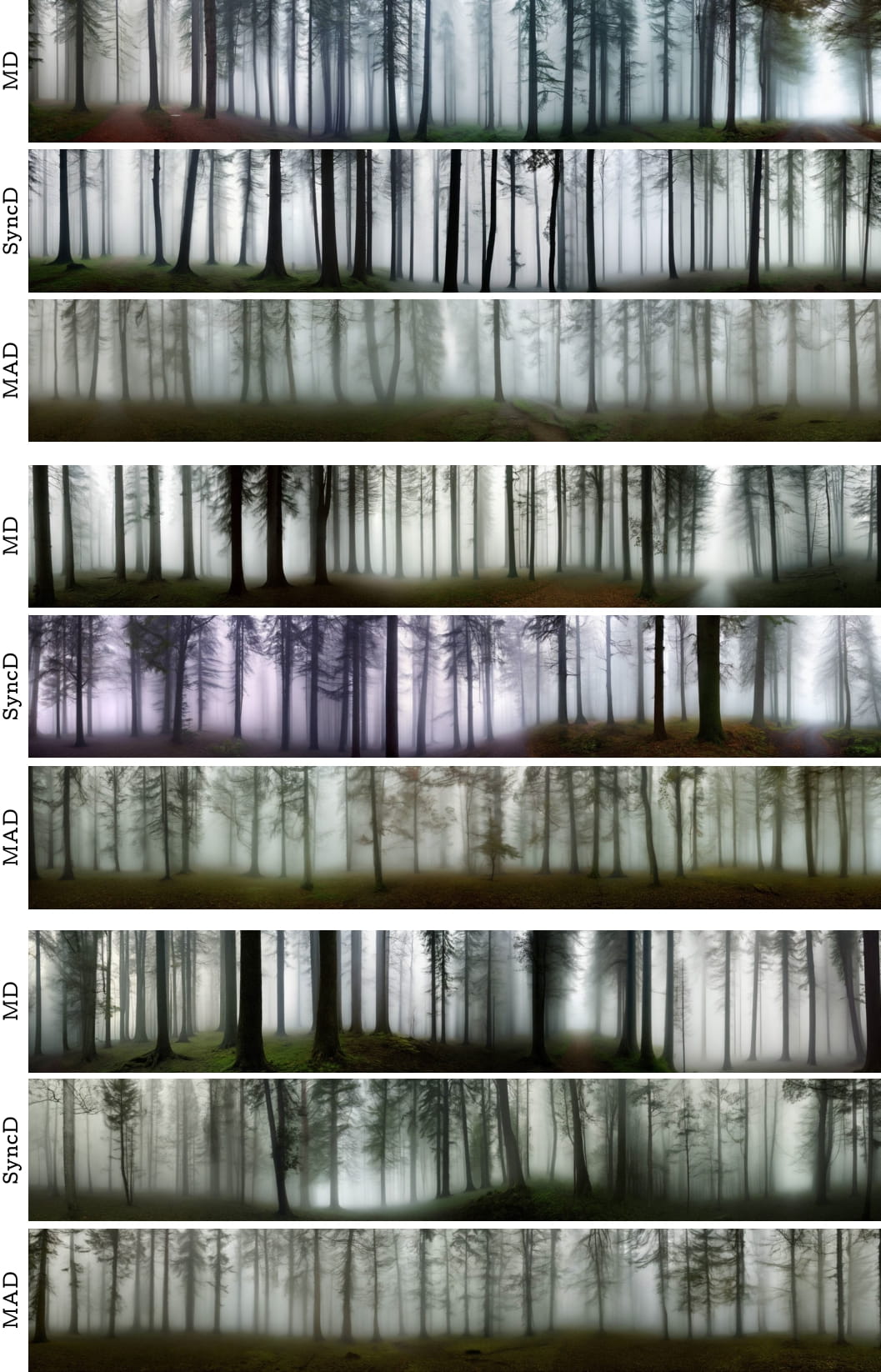}
\caption{Qualitative comparisons with respect to MD and SyncD for the prompt \textit{A photo of a forest with a misty fog}, using the LDM model.}
\label{fig:qualitative_supp_forest}
\end{figure*}

\begin{figure*}[ht]
\centering
\scriptsize
\includegraphics[width=0.8\linewidth]{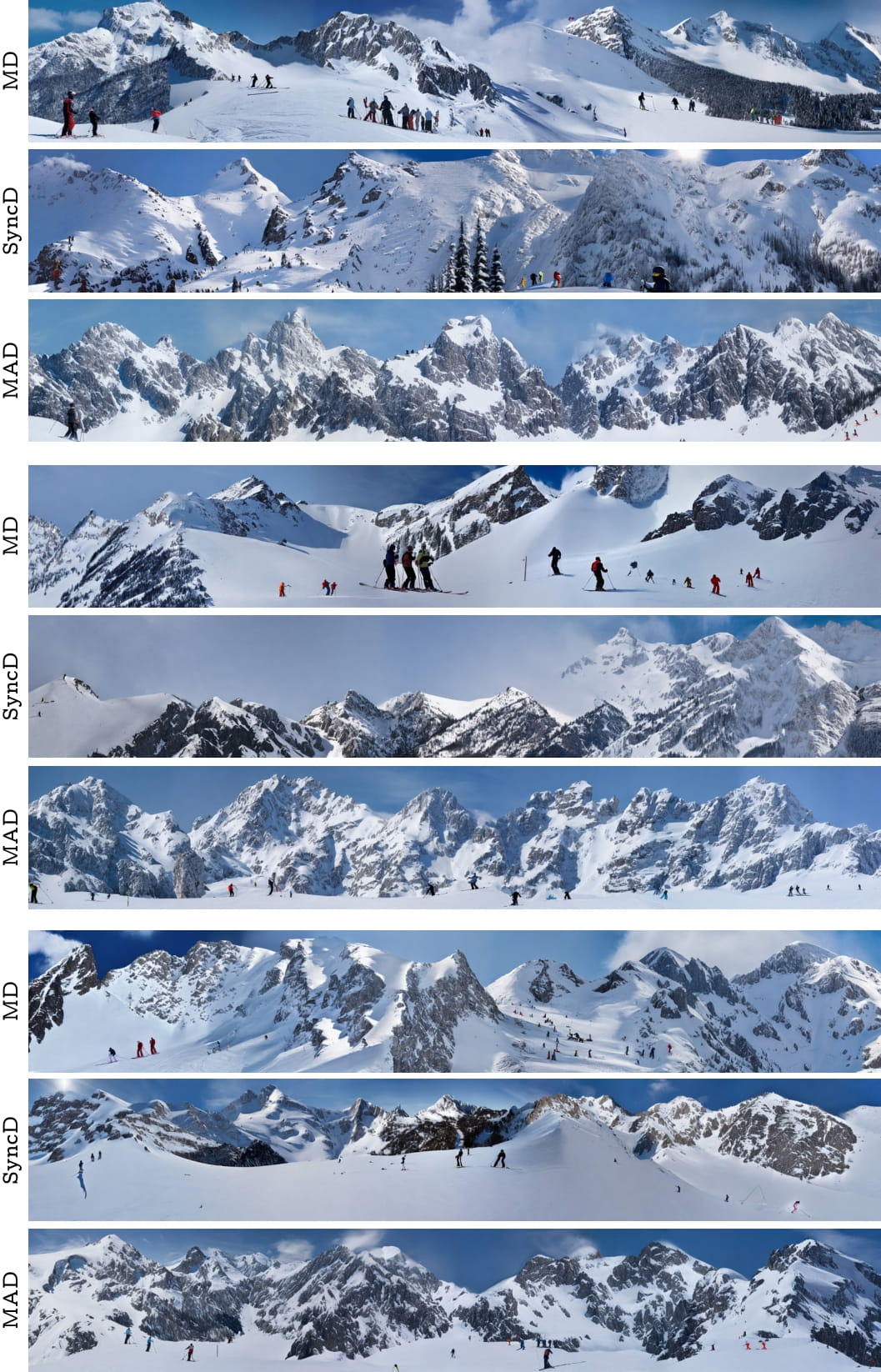}
\caption{Qualitative comparisons with respect to MD and SyncD for the prompt \textit{A photo of a snowy mountain peak with skiers}, using the LDM model.}
\label{fig:qualitative_supp_ski}
\end{figure*}

\begin{figure*}[ht]
\centering
\scriptsize
\includegraphics[width=0.8\linewidth]{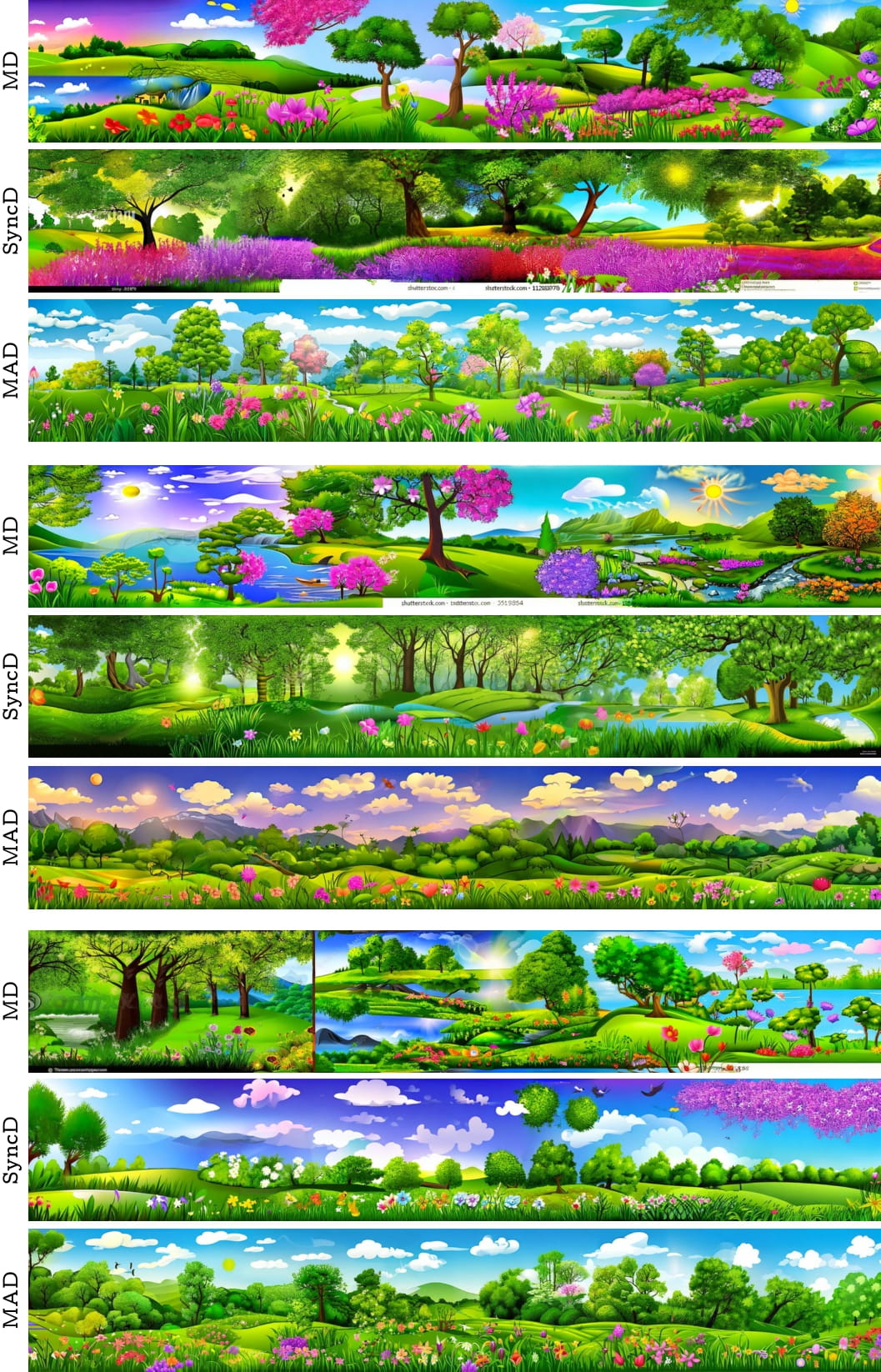}
\caption{Qualitative comparisons with respect to MD and SyncD for the prompt \textit{A cartoon panorama of spring summer beautiful nature}, using the LDM model.}
\label{fig:qualitative_supp_cartoon}
\end{figure*}

\begin{figure*}[ht]
\centering
\scriptsize
 \includegraphics[width=0.8\linewidth]{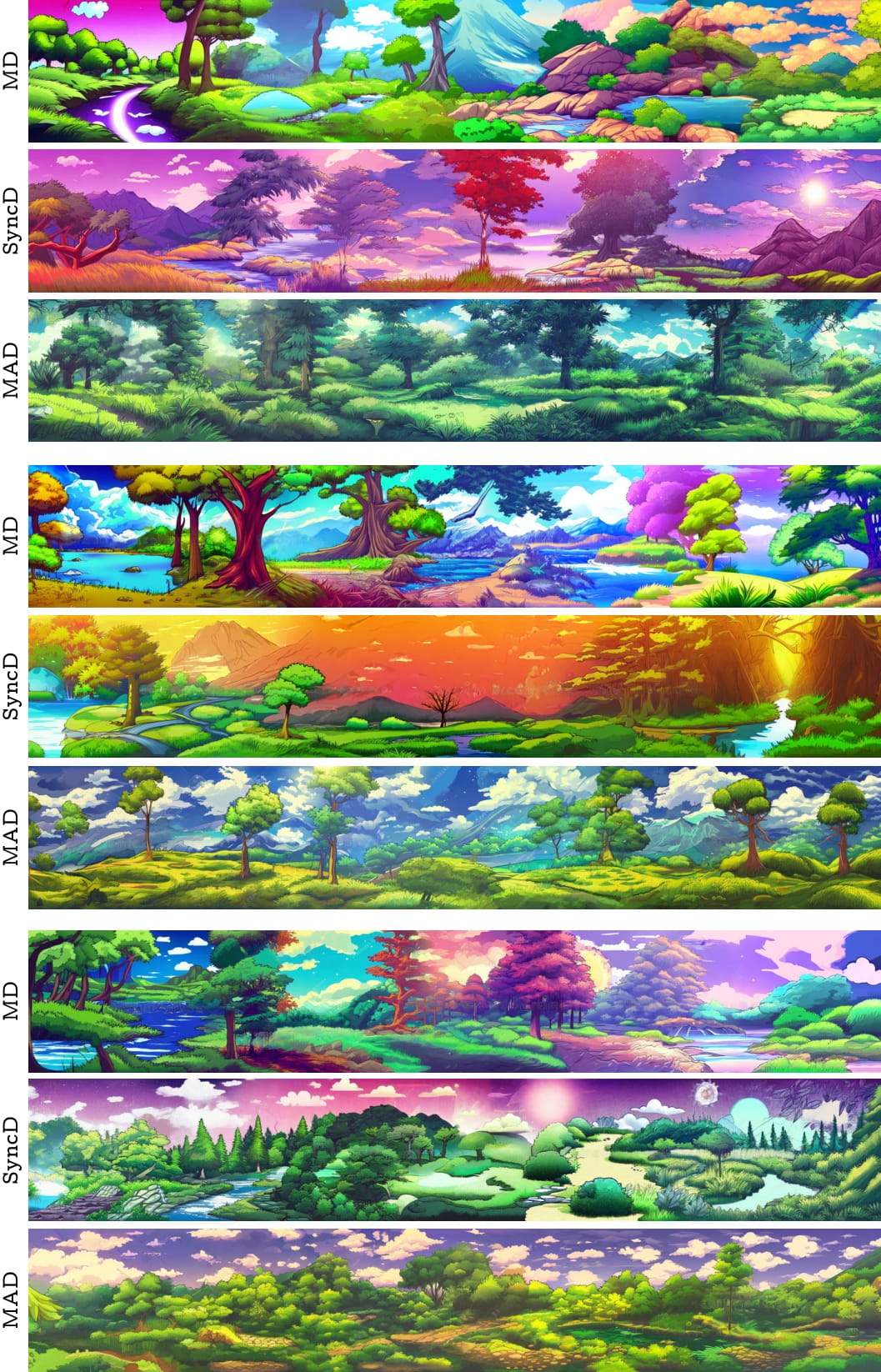}
\caption{Qualitative comparisons with respect to MD and SyncD for the prompt \textit{A natural landscape in anime style illustration}, using the LDM model.}
\label{fig:qualitative_supp_anime}
\end{figure*}

\begin{figure*}[ht]
\centering
\scriptsize
\includegraphics[width=0.8\linewidth]{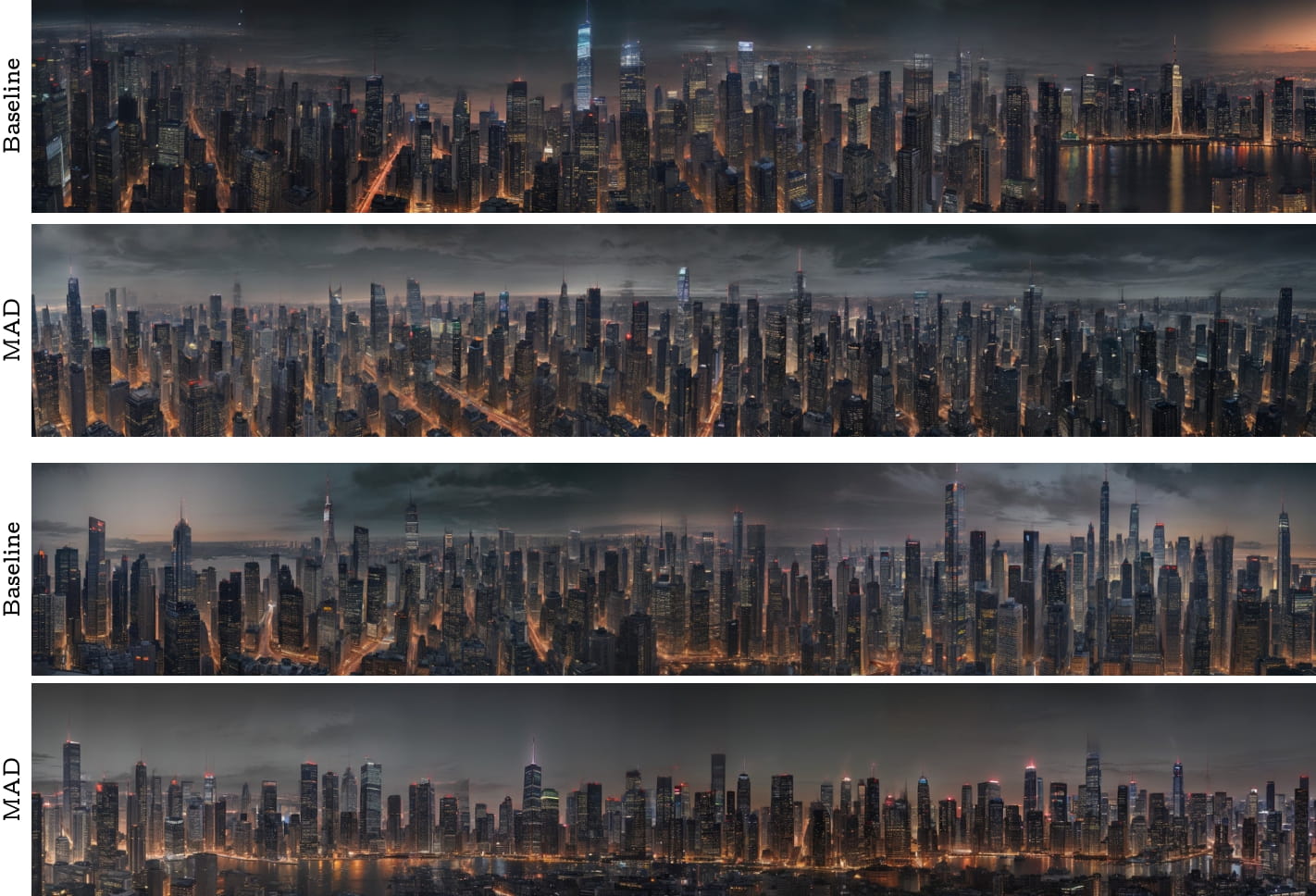}
\caption{Qualitative comparisons with respect to a baseline for the prompt \textit{A photo of a city skyline at night}, using the LCM model.}
\label{fig:qualitative_supp_city_lcm}
\end{figure*}

\begin{figure*}[ht]
\centering
\scriptsize
\includegraphics[width=0.8\linewidth]{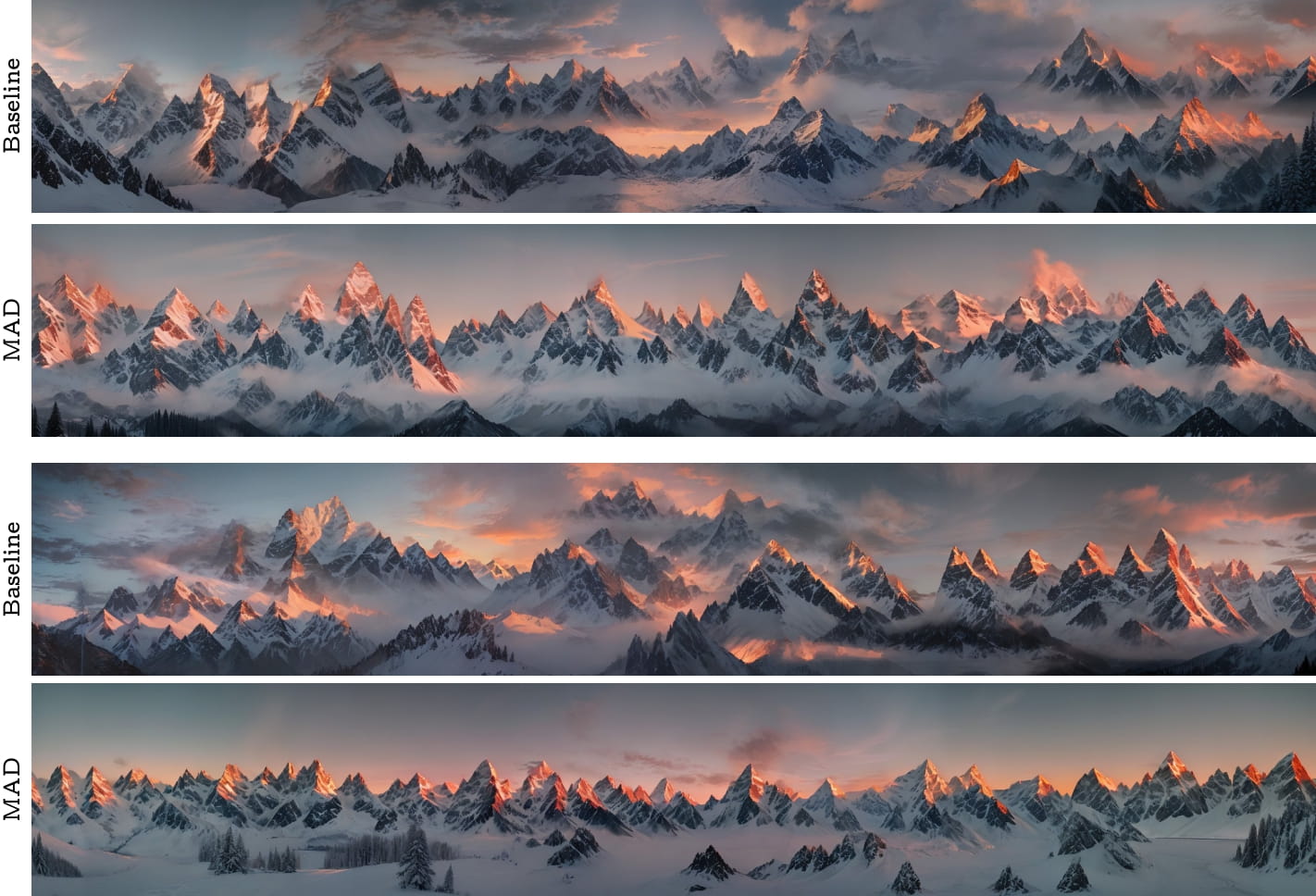}
\caption{Qualitative comparisons with respect to a baseline for the prompt \textit{A photo of a mountain range at twilight}, using the LCM model.}
\label{fig:qualitative_supp_mountain_lcm}
\end{figure*}

\begin{figure*}[ht]
\centering
\scriptsize
\includegraphics[width=0.8\linewidth]{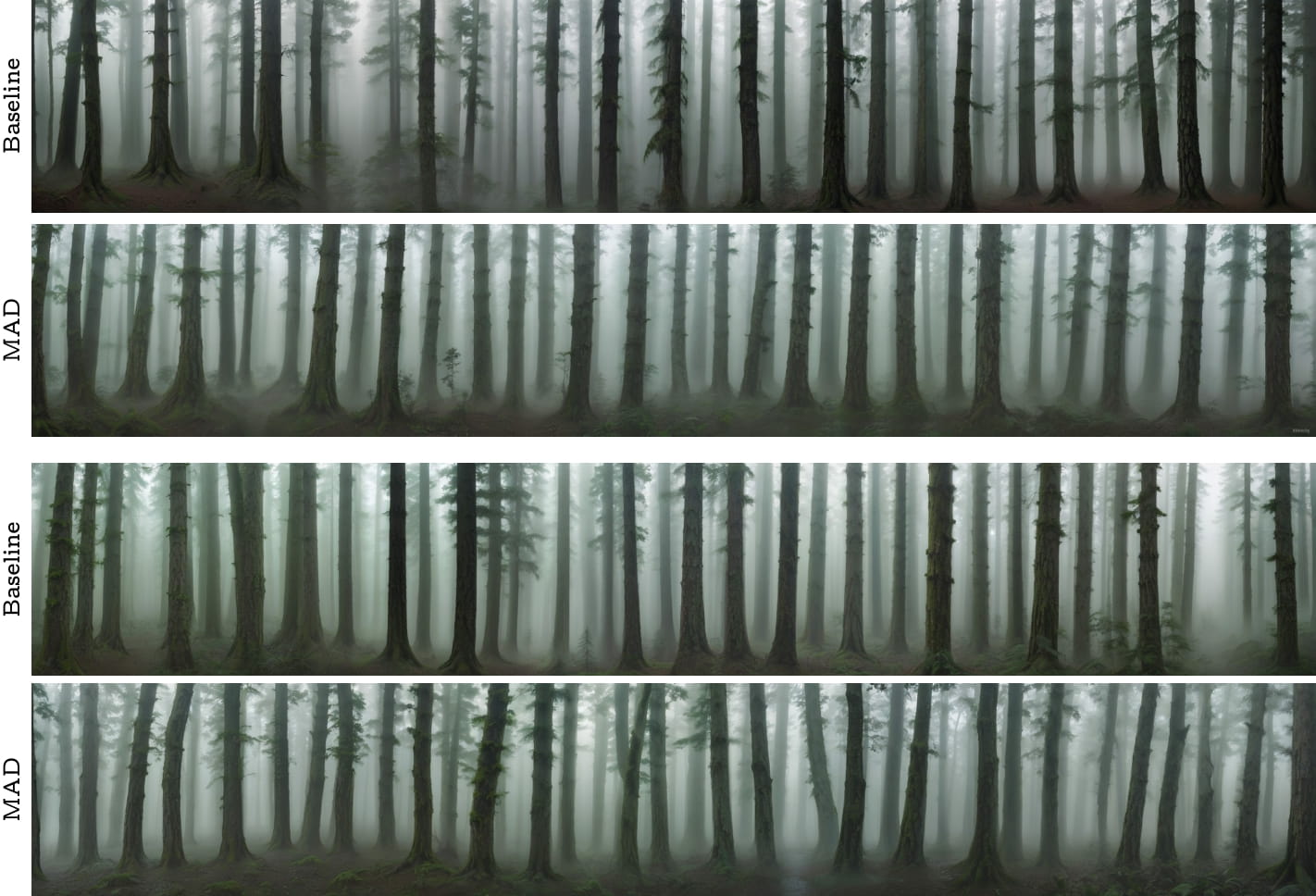}
\caption{Qualitative comparisons with respect to a baseline for the prompt \textit{A photo of a forest with a misty fog}, using the LCM model.}
\label{fig:qualitative_supp_forest_lcm}
\end{figure*}

\begin{figure*}[ht]
\centering
\scriptsize
\includegraphics[width=0.8\linewidth]{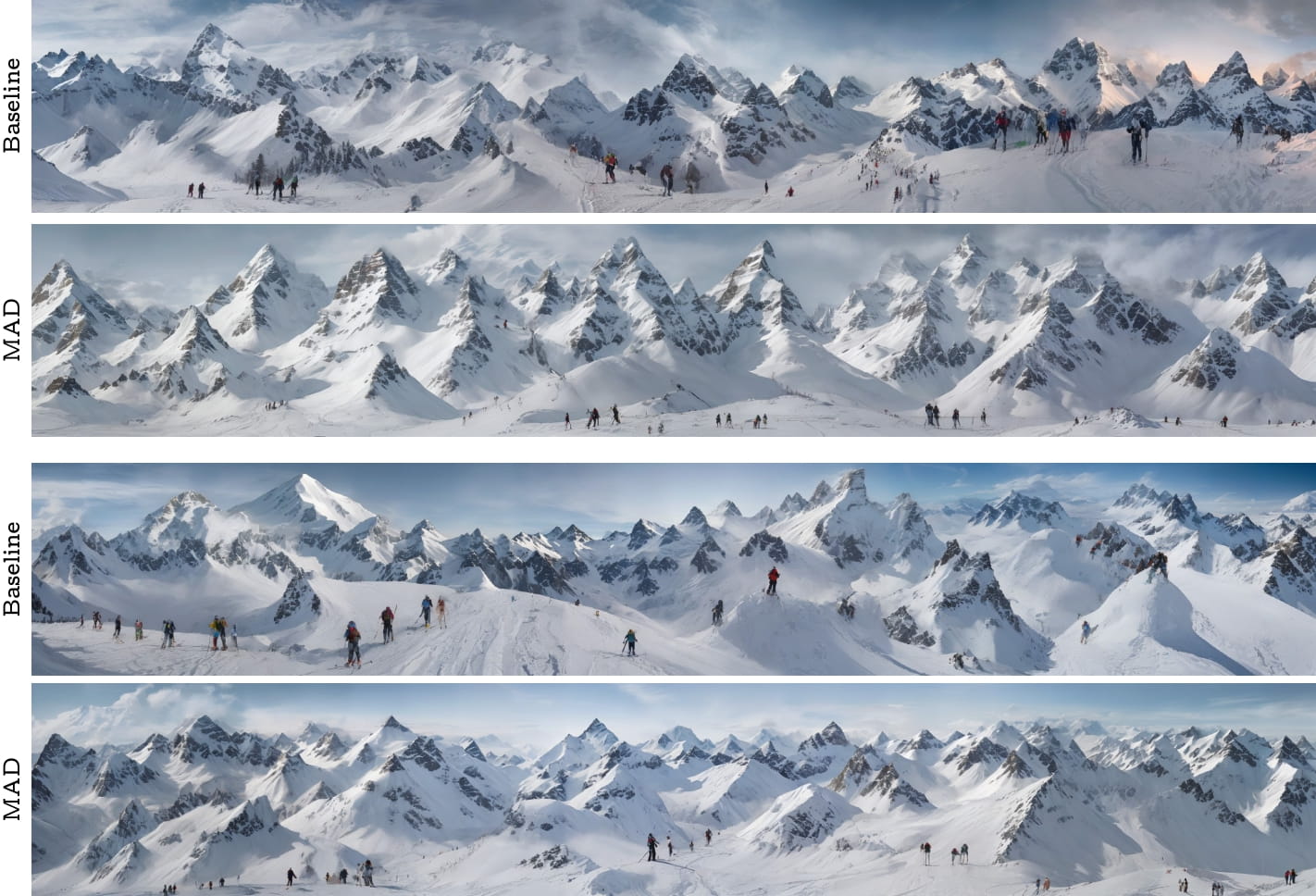}
\caption{Qualitative comparisons with respect to a baseline for the prompt \textit{A photo of a snowy mountain peak with skiers}, using the LCM model.}
\label{fig:qualitative_supp_skiers_lcm}
\end{figure*}

\begin{figure*}[ht]
\centering
\scriptsize
\includegraphics[width=0.8\linewidth]{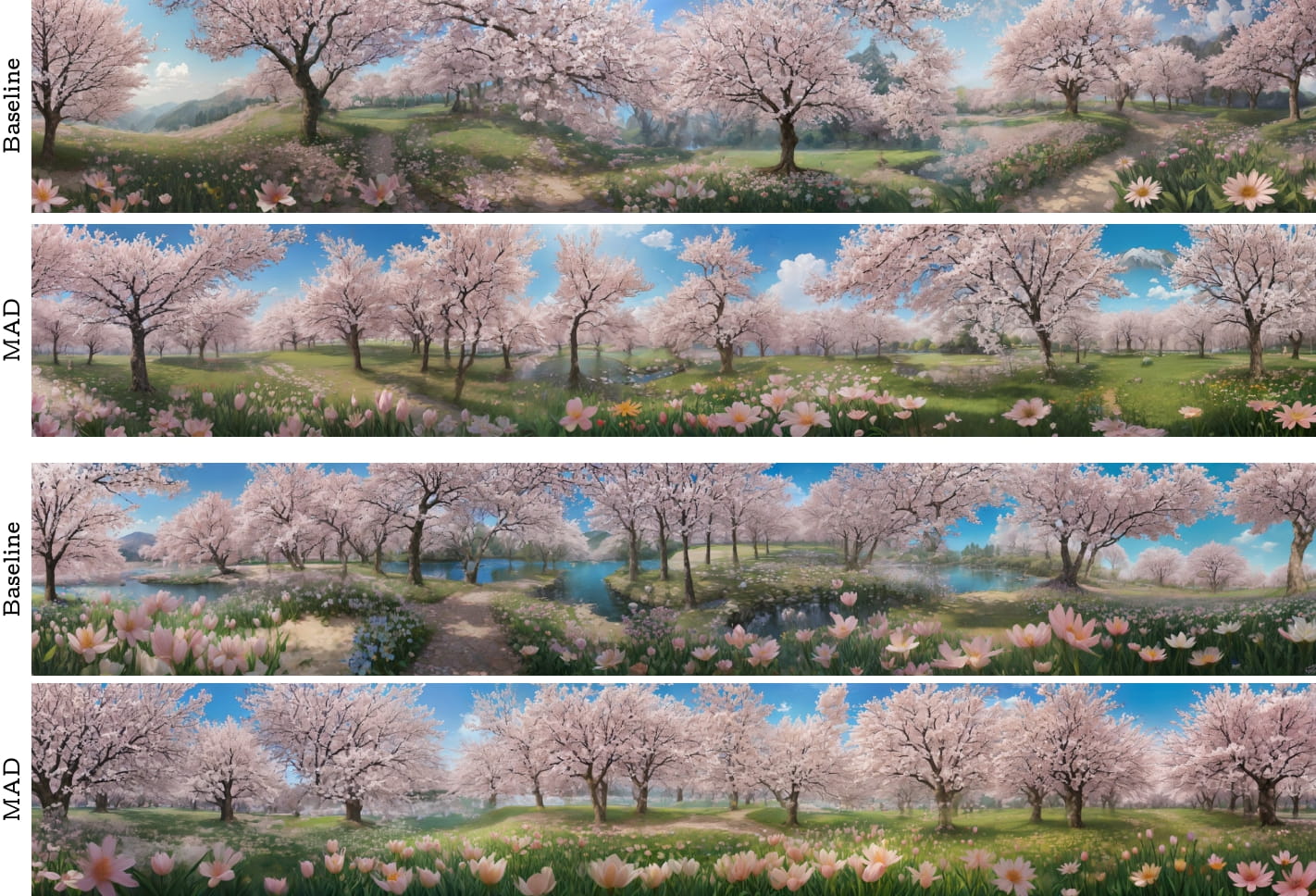}
\caption{Qualitative comparisons with respect to a baseline for the prompt \textit{A cartoon panorama of spring summer beautiful nature}, using the LCM model.}
\label{fig:qualitative_supp_spring_lcm}
\end{figure*}

\begin{figure*}[ht]
\centering
\scriptsize
\includegraphics[width=0.8\linewidth]{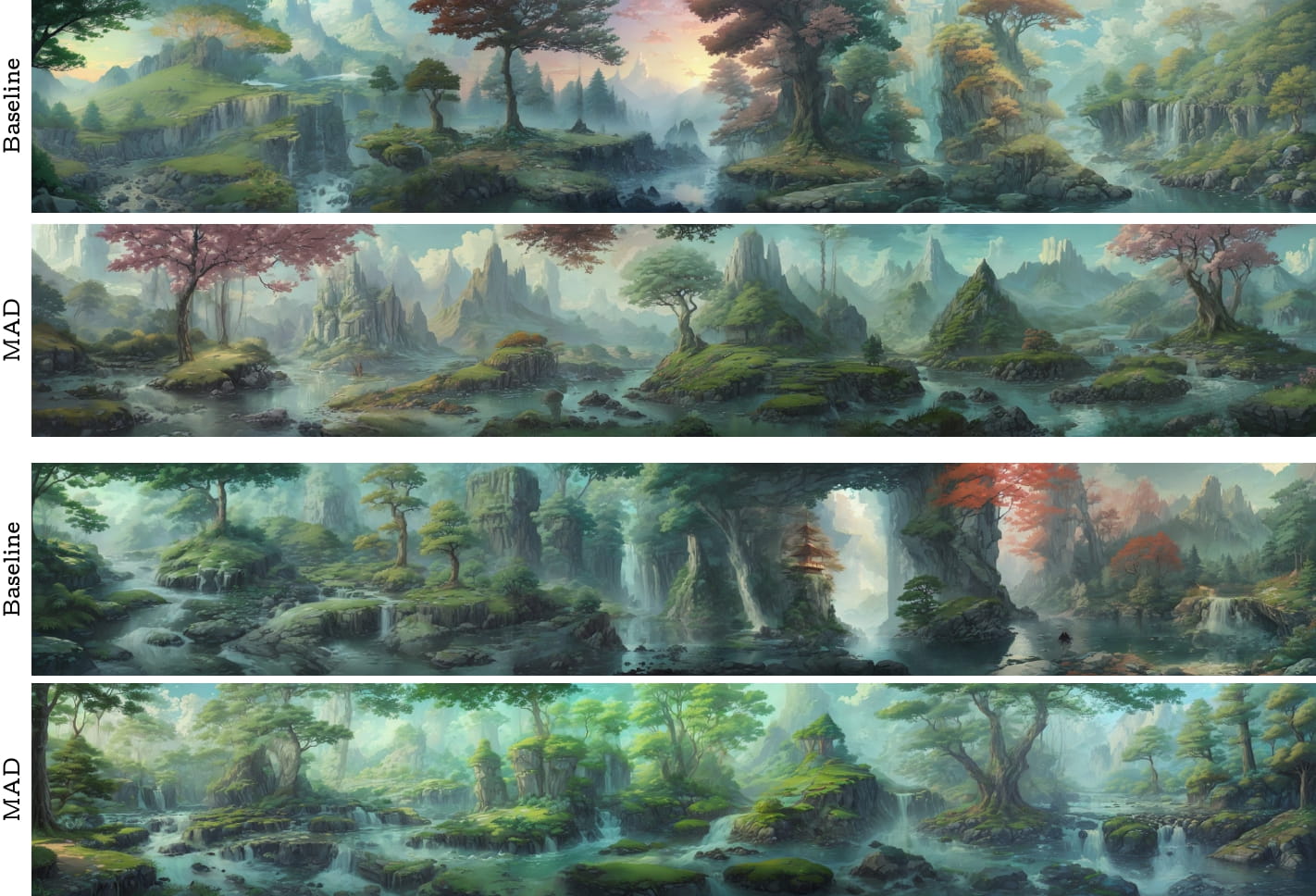}
\caption{Qualitative comparisons with respect to a baseline for the prompt \textit{A natural landscape in anime style illustration}, using the LCM model.}
\label{fig:qualitative_supp_anime_lcm}
\end{figure*}

We report some randomly picked qualitative results on the GPT1k prompts and the six prompts used for the quantitative analysis of the LDM in~\cref{fig:qualitative_supp_gpt1k_1,fig:qualitative_supp_gpt1k_2,fig:qualitative_supp_city,fig:qualitative_supp_mount,fig:qualitative_supp_forest,fig:qualitative_supp_ski,fig:qualitative_supp_cartoon,fig:qualitative_supp_anime} and of the LCM in~\cref{fig:qualitative_supp_gpt1k_lcm_1,fig:qualitative_supp_gpt1k_lcm_2,fig:qualitative_supp_city_lcm,fig:qualitative_supp_mountain_lcm,fig:qualitative_supp_forest_lcm,fig:qualitative_supp_skiers_lcm,fig:qualitative_supp_spring_lcm,fig:qualitative_supp_anime_lcm}. 
Moreover, in~\cref{fig:supp_qualitatives_H_ldm_additional,fig:supp_qualitatives_H_lcm_additional,fig:supp_qualitatives_V_ldm_additional,fig:supp_qualitatives_V_lcm_additional}, we report more qualitative results, both horizontal and vertical, on different GPT1k prompts for the LDM and LCM, respectively.

\section{Plug\&Play Applications}
Note that our MAD operator can be used as it is in settings such as tight and rough region-based generation, and conditional image generation (with Canny edge map guidance). since it is applied to the attention layers of the noise prediction UNet. The visual quality of the resulting images, shown in~\Cref{fig:applications}, is on par with that of the generated panoramas (which is our main focus), even for guidance spanning multiple views of the large image.
    
\section{Further Discussion of the Limitations}
\begin{figure*}[t]
\centering
\scriptsize
\includegraphics[width=0.7\linewidth]{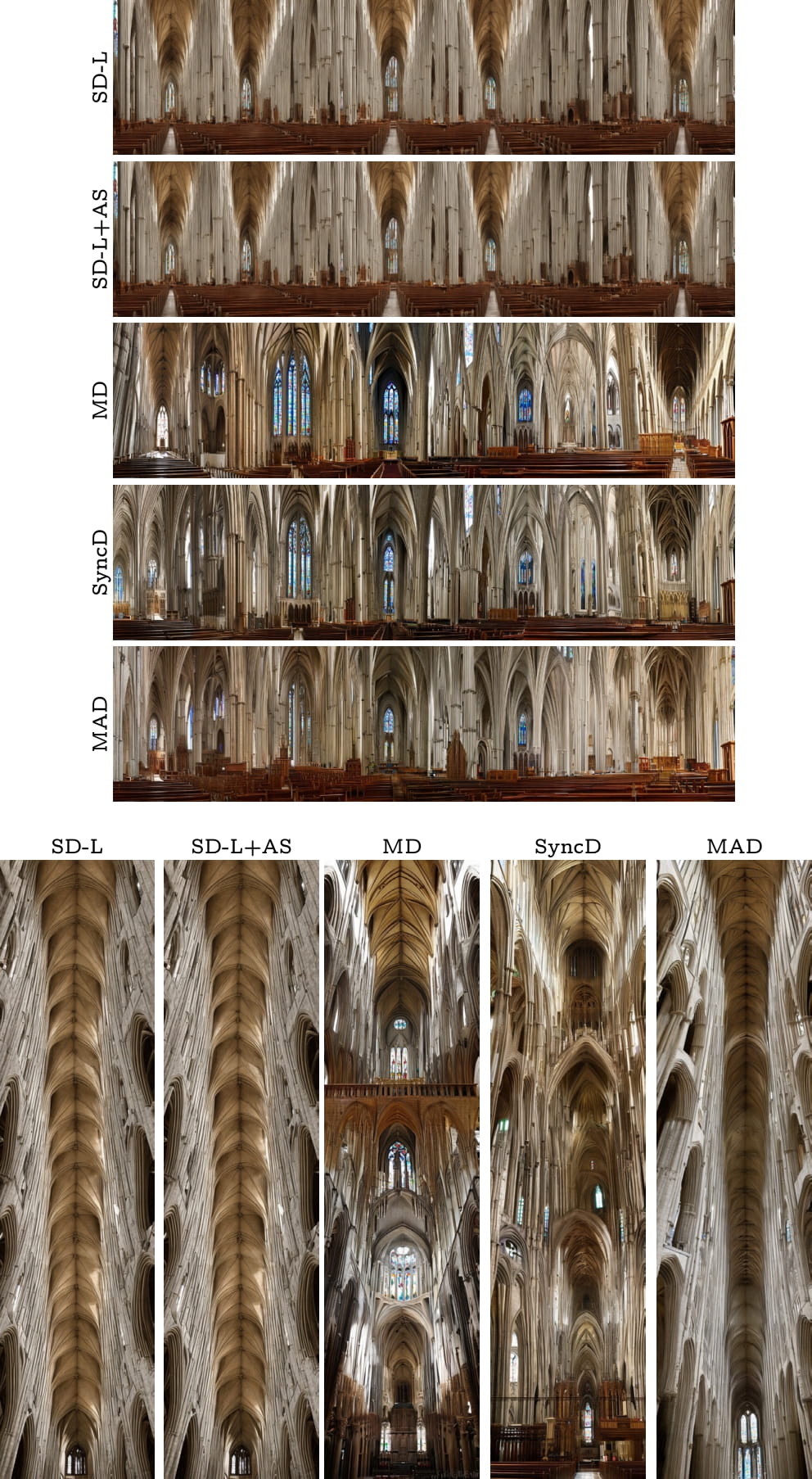}
\caption{Horizontal and vertical images generated with the same seed and for the prompt \textit{A gothic cathedral nave} using different methods applied on the LDM model: direct inference (SD-L), Attention Scaling (SD-L+AS), MD, SyncD, and MAD.}
\label{fig:limitations_v}
\end{figure*}

Approaches for panorama image generation that adapt diffusion models pretrained on squared images, including ours, can be limited by the performance and the image distribution learned by the base model. This results in generated images whose quality depends on the backbone, as showcased in~\cref{fig:backbones}. 
Moreover, inference-time panorama generation approaches struggle with scenes or objects that do not fit well with the specified aspect ratio and prompts where the base model itself does not produce good-quality results. We provide examples in~\cref{fig:limitations_h,fig:limitations_v}. As we can see, the prompt \textit{A gothic cathedral nave} (\cref{fig:limitations_v}) does not fit well with the horizontal aspect ratio. However, when asked to generate a vertical image, the results improve. 
Moreover, the prompt \textit{A fancy living room}, which entails a usually small indoor space, is rendered at an inferior quality when the desired output width is too large. 

\clearpage

\begin{table*}[h]
        \centering
        \footnotesize
        \setlength{\tabcolsep}{.28em}
        \caption{Quantitative results on the generation of images with MAD applied at different stages of the LDM noise prediction model to generate 512${\times}$3072 images.}
        \label{tab:quantitatives_ldm_blocks}\vspace{-.8em}
        \resizebox{\linewidth}{!}{
        \begin{tabular}{lc c c c c c c}
        \toprule
        && \textbf{mCLIP${\uparrow}$} & \textbf{I-LPIPS${\downarrow}$} & \textbf{I-StyleL${\downarrow}$} & \textbf{FID${\downarrow}$} & \textbf{KID${\downarrow}$} & \textbf{mGIQA${\uparrow}$} \\
        && & & \textbf{($\times 10^{-3}$)} & & \textbf{($\times 10^{-3}$)} & \textbf{($\times 10^{-3}$)}\\
        \midrule
        \textbf{No MAD (Baseline)}       && 31.65${\pm}$2.17 & 0.64${\pm}$0.10 & 2.65${\pm}$2.33 & 34.51${\pm}$13.92 & 9.19${\pm}$4.51 & 27.59${\pm}$6.83 \\
        \textbf{MAD in Mid Block}      && 31.66${\pm}$2.16 & 0.64${\pm}$0.10 & 2.19${\pm}$1.17 & 34.80${\pm}$13.80 & 9.35${\pm}$3.91 & 28.04${\pm}$7.23 \\
        \textbf{MAD in Down Blocks}    && 31.83${\pm}$2.21 & 0.62${\pm}$0.10 & 1.98${\pm}$1.03 & 40.46${\pm}$16.58 & 16.40${\pm}$8.83 & 28.11${\pm}$7.42 \\
        \textbf{MAD in Up Blocks}      && 32.09${\pm}$2.20 & 0.53${\pm}$0.10 & 1.19${\pm}$0.86 & 60.84${\pm}$24.65 & 43.56${\pm}$19.82 & 28.60${\pm}$7.51 \\
        \textbf{MAD in All Blocks}     && 32.15${\pm}$2.25 & 0.53${\pm}$0.10 & 1.43${\pm}$1.04 & 61.76${\pm}$23.73 & 43.31${\pm}$17.41 & 28.10${\pm}$7.84 \\
        \bottomrule
        \end{tabular} 
        }\vspace{-2.3em}
\end{table*} 

\begin{table*}[h]
\footnotesize
    \centering
    \setlength{\tabcolsep}{.28em}
    \caption{Quantitative results on the generation of images with MAD applied at different stages of the LCM noise prediction model to generate 512${\times}$3072 images.}
    \label{tab:quantitatives_lcm_blocks}\vspace{-.8em}
    \resizebox{\linewidth}{!}{
    \begin{tabular}{lc c c c c c c}
    \toprule
    && \textbf{mCLIP${\uparrow}$} & \textbf{I-LPIPS${\downarrow}$} & \textbf{I-StyleL${\downarrow}$} & \textbf{FID${\downarrow}$} & \textbf{KID${\downarrow}$}   & \textbf{mGIQA${\uparrow}$}   \\
    &&                           &                               & \textbf{($\times 10^{-3}$)}    &                           & \textbf{($\times 10^{-3}$)} & \textbf{($\times 10^{-3}$)} \\
    \midrule
    \textbf{No MAD (Baseline)}     && 31.36${\pm}$1.83 & 0.55${\pm}$0.05 & 0.90${\pm}$0.35 & 31.35${\pm}$15.42 & 13.27${\pm}$10.05 & 35.44${\pm}$12.38 \\
    \textbf{MAD in Mid Block}      && 31.44${\pm}$1.79 & 0.55${\pm}$0.05 & 0.87${\pm}$0.33 & 29.35${\pm}$15.62 & 13.85${\pm}$11.27 & 35.49${\pm}$12.22 \\
    \textbf{MAD in Down Blocks}    && 31.44${\pm}$1.78 & 0.55${\pm}$0.05 & 0.84${\pm}$0.32 & 29.01${\pm}$14.83 & 13.35${\pm}$11.02 & 35.38${\pm}$12.46 \\
    \textbf{MAD in Up Blocks}      && 31.46${\pm}$1.84 & 0.51${\pm}$0.05 & 0.71${\pm}$0.26 & 40.22${\pm}$20.99 & 28.98${\pm}$17.43 & 35.45${\pm}$13.08 \\
    \textbf{MAD in All Blocks}     && 31.48${\pm}$1.87 & 0.52${\pm}$0.05 & 0.71${\pm}$0.26 & 38.77${\pm}$19.85 & 27.41${\pm}$16.19 & 35.23${\pm}$13.15 \\
    \bottomrule
    \end{tabular}
    }
\end{table*} 

\begin{table*}[h]
    \footnotesize
        \centering
        \setlength{\tabcolsep}{.38em}
        \caption{Results on the generation of 512$\times$3072 images with MAD applied for different numbers of timesteps from the beginning of the reverse diffusion process in the LDM.}
        \label{tab:ablation_threshold_quantitative}\vspace{-.8em}
        \resizebox{.85\linewidth}{!}{
        \begin{tabular}{lc c c c c c c}
        \toprule
        && \textbf{mCLIP${\uparrow}$} & \textbf{I-LPIPS${\downarrow}$} & \textbf{I-StyleL${\downarrow}$} & \textbf{FID${\downarrow}$} & \textbf{KID${\downarrow}$} & \textbf{mGIQA${\uparrow}$} \\
        && & & \textbf{($\times 10^{-3}$)} & & \textbf{($\times 10^{-3}$)} & \textbf{($\times 10^{-3}$)}\\
        \midrule
        \textbf{$\tau{=}$0}  && 31.65${\pm}$2.17 & 0.64${\pm}$0.10 & 2.65${\pm}$2.33 & 34.51${\pm}$13.92 & 9.19${\pm}$4.51 & 27.59${\pm}$6.83 \\
        \textbf{$\tau{=}$5}      && 31.86${\pm}$2.22 & 0.59${\pm}$0.10 & 2.07${\pm}$1.31 & 38.10${\pm}$13.71 & 13.75${\pm}$4.00 & 28.32${\pm}$7.64 \\
        \textbf{$\tau{=}$10}     && 31.95${\pm}$2.24 & 0.57${\pm}$0.10 & 2.00${\pm}$1.34 & 42.60${\pm}$14.94 & 19.38${\pm}$5.56 & 28.36${\pm}$7.83 \\
        \textbf{$\tau{=}$15}     && 32.03${\pm}$2.29 & 0.56${\pm}$0.10 & 1.90${\pm}$1.32 & 48.52${\pm}$17.14 & 27.15${\pm}$9.10 & 28.32${\pm}$7.76 \\
        \textbf{$\tau{=}$20}     && 32.09${\pm}$2.29 & 0.54${\pm}$0.10 & 1.68${\pm}$1.19 & 54.31${\pm}$19.92 & 34.43${\pm}$12.54 & 28.23${\pm}$7.90 \\
        \textbf{$\tau{=}$25}     && 32.15${\pm}$2.25 & 0.53${\pm}$0.10 & 1.43${\pm}$1.04 & 61.76${\pm}$23.73 & 43.31${\pm}$17.41 & 28.10${\pm}$7.84 \\
        \textbf{$\tau{=}$30}     && 32.16${\pm}$2.17 & 0.52${\pm}$0.10 & 1.22${\pm}$0.90 & 70.29${\pm}$28.24 & 54.40${\pm}$23.19 & 27.94${\pm}$7.73 \\
        \textbf{$\tau{=}$35}     && 32.18${\pm}$2.08 & 0.51${\pm}$0.10 & 1.02${\pm}$0.73 & 78.50${\pm}$32.94 & 65.06${\pm}$29.94 & 27.82${\pm}$7.65 \\
        \textbf{$\tau{=}$40}     && 32.16${\pm}$1.95 & 0.50${\pm}$0.10 & 0.88${\pm}$0.61 & 86.20${\pm}$37.23 & 76.15${\pm}$37.73 & 27.59${\pm}$7.61 \\
        \textbf{$\tau{=}$45}     && 32.14${\pm}$1.83 & 0.50${\pm}$0.10 & 0.78${\pm}$0.53 & 92.69${\pm}$40.76 & 84.13${\pm}$44.13 & 27.35${\pm}$7.50 \\
        \textbf{$\tau{=}$50}     && 32.14${\pm}$1.72 & 0.49${\pm}$0.10 & 0.71${\pm}$0.47 & 98.01${\pm}$43.64 & 91.51${\pm}$49.95 & 27.05${\pm}$7.28 \\
        \bottomrule
        \end{tabular}
        }\vspace{-2.3em}
    \end{table*} 

\begin{figure}[t]
\begin{minipage}[b]{0.485\linewidth}
\centering
        \scriptsize
        \setlength{\tabcolsep}{0.15em}
        \begin{tabular}{m{0.6em} c}
             \rotatebox[origin=l]{90}{\textbf{$\tau{=}$0}}  & \makecell{\includegraphics[width=5.5cm]{images/suppl_thresh_horses/horses0.jpg}}\\
             \rotatebox[origin=l]{90}{\textbf{$\tau{=}$5}}  & \makecell{\includegraphics[width=5.5cm]{images/suppl_thresh_horses/horses5.jpg}}\\
             \rotatebox[origin=l]{90}{\textbf{$\tau{=}$10}} & \makecell{\includegraphics[width=5.5cm]{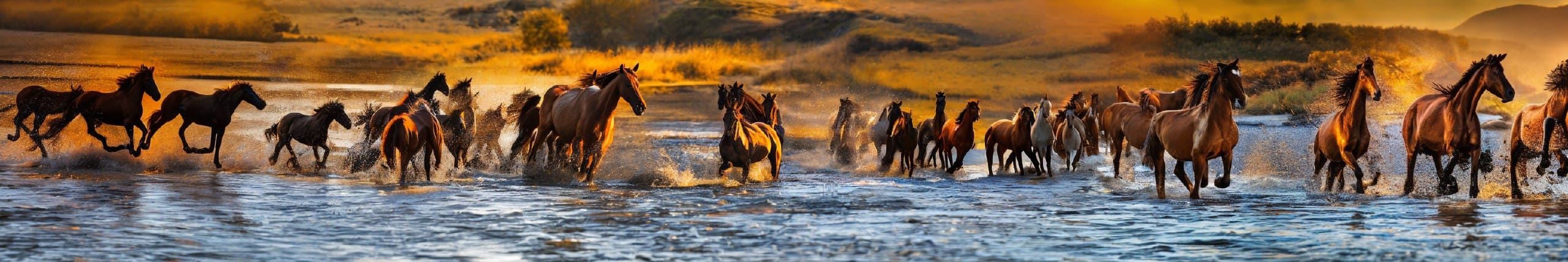}}\\
             \rotatebox[origin=l]{90}{\textbf{$\tau{=}$15}} & \makecell{\includegraphics[width=5.5cm]{images/suppl_thresh_horses/horses15.jpg}}\\
             \rotatebox[origin=l]{90}{\textbf{$\tau{=}$20}} & \makecell{\includegraphics[width=5.5cm]{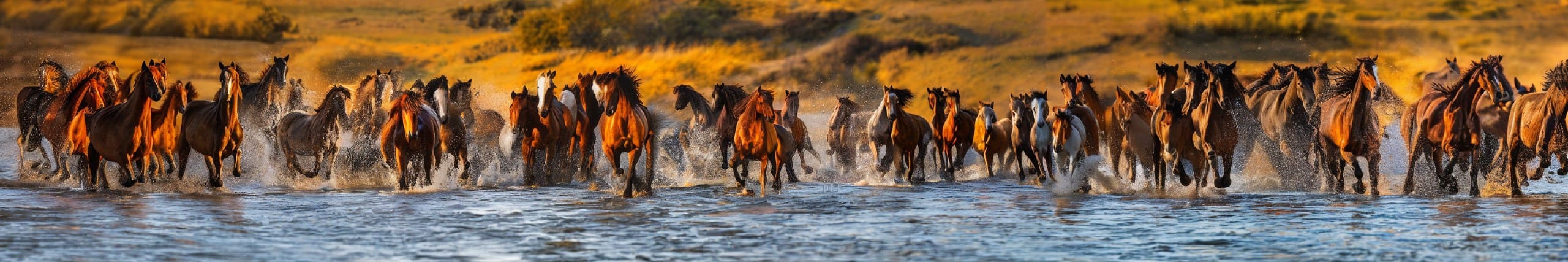}}\\
             \rotatebox[origin=l]{90}{\textbf{$\tau{=}$25}} & \makecell{\includegraphics[width=5.5cm]{images/suppl_thresh_horses/horses25.jpg}}\\
             \rotatebox[origin=l]{90}{\textbf{$\tau{=}$30}} & \makecell{\includegraphics[width=5.5cm]{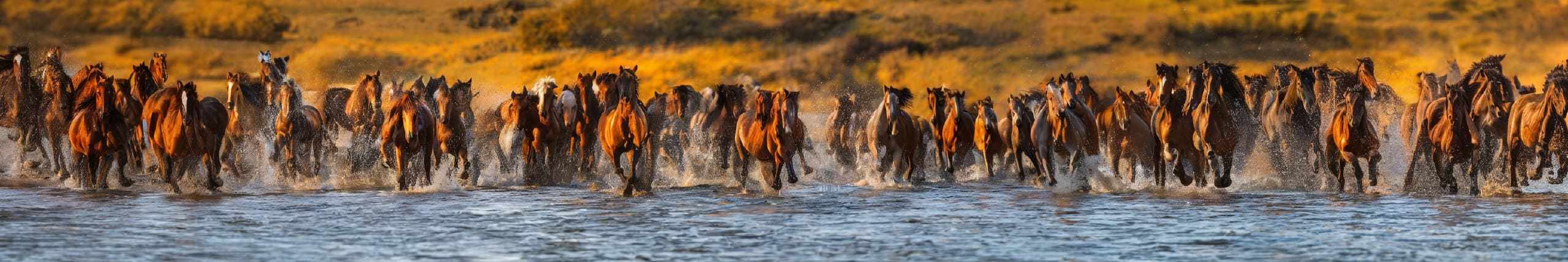}}\\
             \rotatebox[origin=l]{90}{\textbf{$\tau{=}$35}} & \makecell{\includegraphics[width=5.5cm]{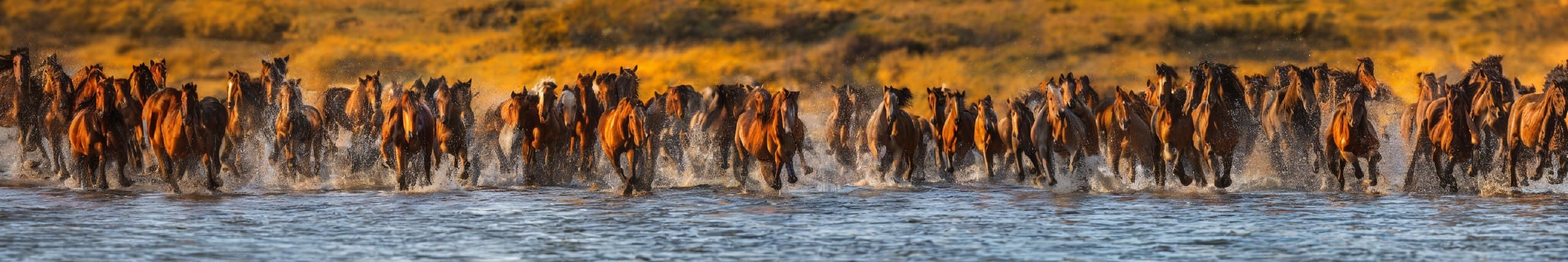}}\\
             \rotatebox[origin=l]{90}{\textbf{$\tau{=}$40}} & \makecell{\includegraphics[width=5.5cm]{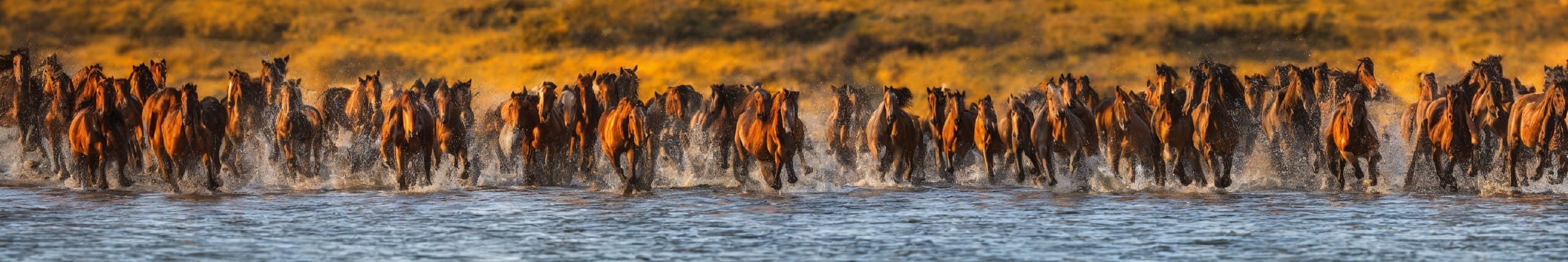}}\\
             \rotatebox[origin=l]{90}{\textbf{$\tau{=}$45}} & \makecell{\includegraphics[width=5.5cm]{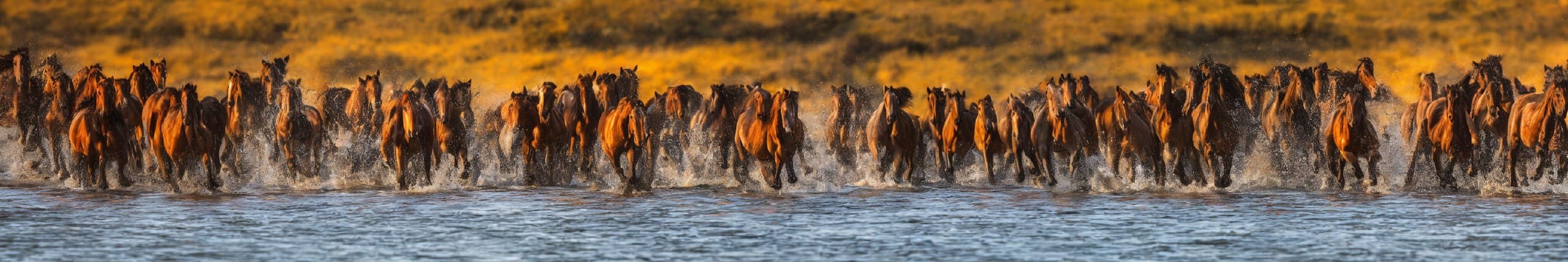}}\\
             \rotatebox[origin=l]{90}{\textbf{$\tau{=}$50}} & \makecell{\includegraphics[width=5.5cm]{images/suppl_thresh_horses/horses50.jpg}}\\
        \end{tabular}
    \caption{Long images generated by the considered LDM with MAD applied up to different numbers of inference steps for the prompt \textit{A herd of Mustang horses crossing a river at sunset}. When $\tau$ is too low, the view interactions are not enough to produce a globally coherent image. As $\tau$ increases, the image becomes more and more coherent, with maximal uniformity when MAD is applied at every timestep.}
    \label{fig:ablation_threshold_qualitative}
\end{minipage}
\hfill
\begin{minipage}[b]{0.485\linewidth}
        \centering
        \scriptsize
        \setlength{\tabcolsep}{0.15em}
        \begin{tabular}{m{0.6em} c}
             \rotatebox[origin=l]{90}{\textbf{None}} & \makecell{\includegraphics[width=5.5cm]{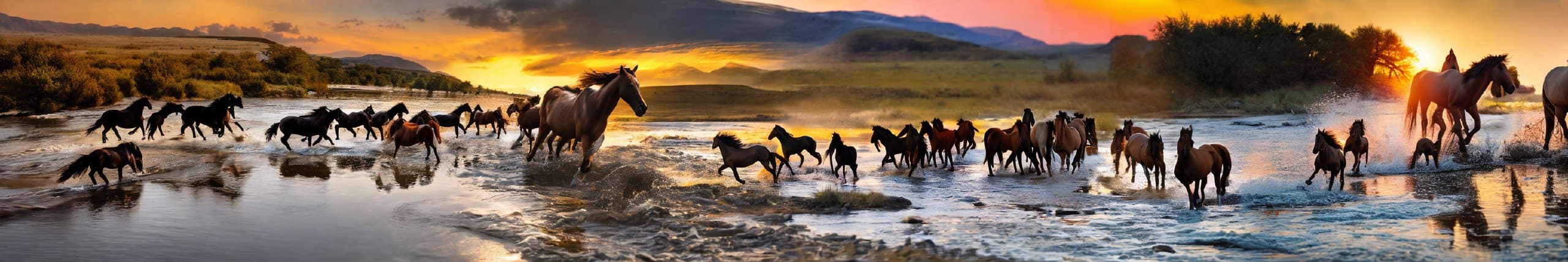}}\\
             \rotatebox[origin=l]{90}{\textbf{Mid}}  & \makecell{\includegraphics[width=5.5cm]{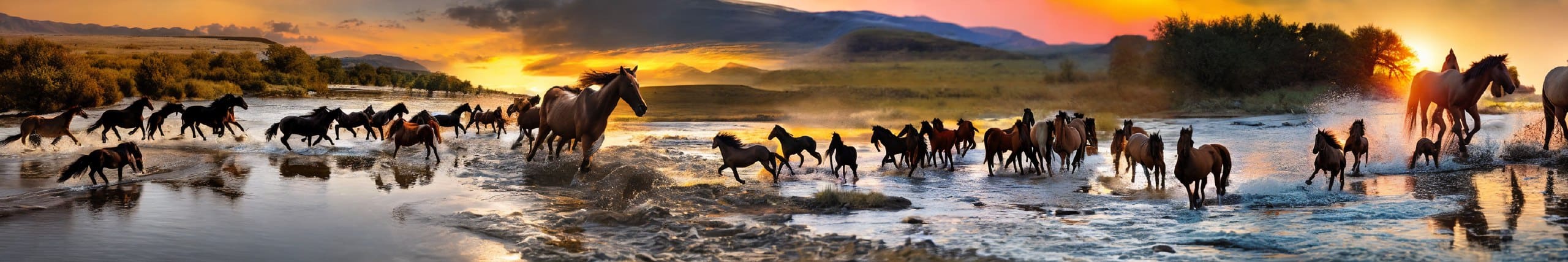}} \\
             \rotatebox[origin=l]{90}{\textbf{Down}} & \makecell{\includegraphics[width=5.5cm]{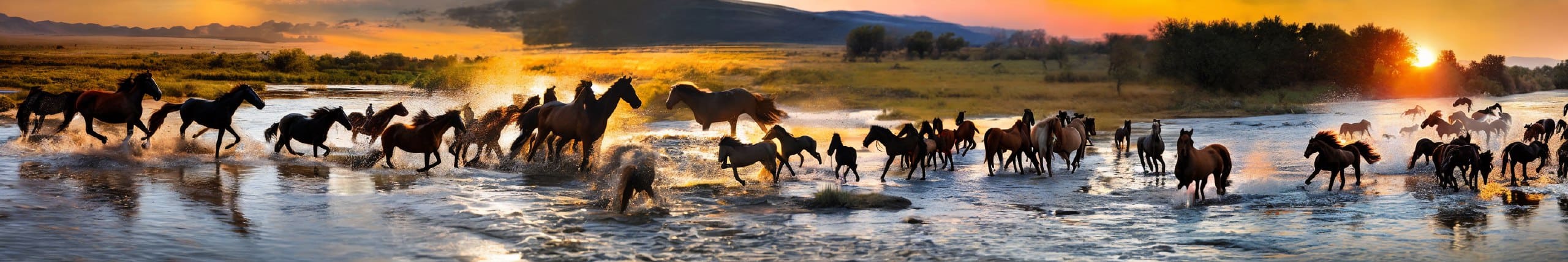}}\\
             \rotatebox[origin=l]{90}{\textbf{Up}}   & \makecell{\includegraphics[width=5.5cm]{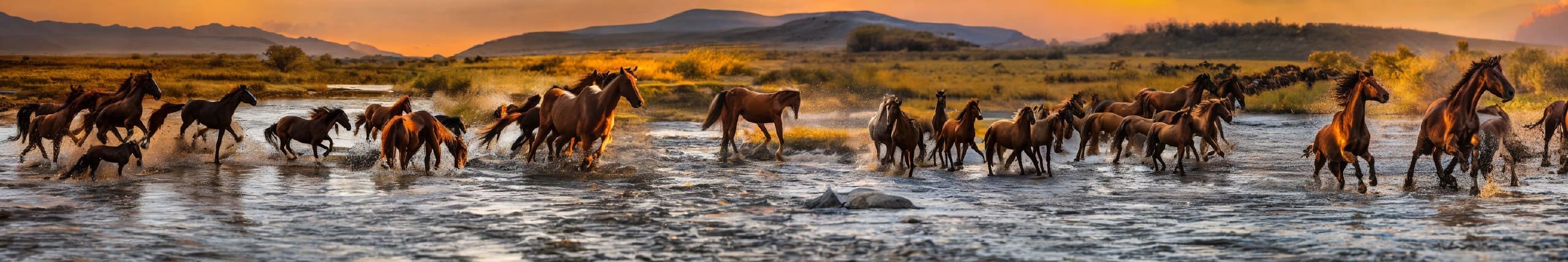}}  \\
             \rotatebox[origin=l]{90}{\textbf{All}}  & \makecell{\includegraphics[width=5.5cm]{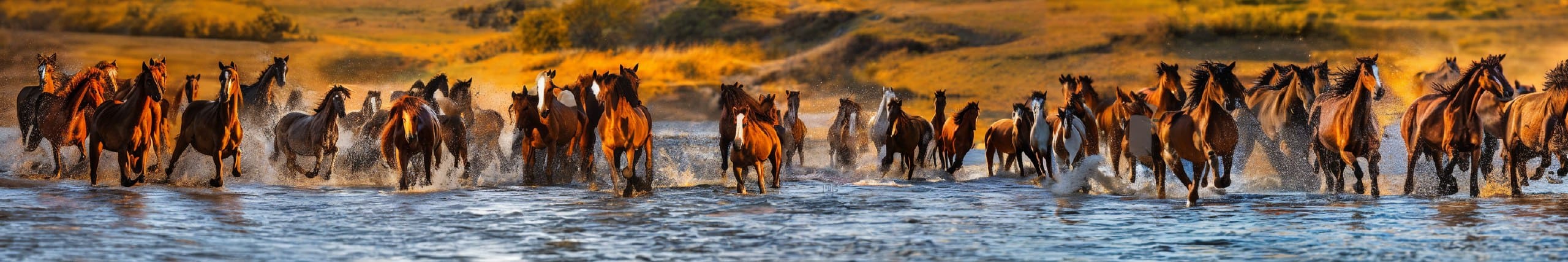}} 
        \end{tabular}\vspace{-.5em}
    \caption{Long images generated by the considered LDM with MAD applied in different blocks of the noise prediction model, with $\tau{=}$15 for the prompt \textit{A herd of Mustang horses crossing a river at sunset}. 
    }
    \label{fig:ablation_blocks_sup}\vspace{1.5em}
\vfill
    \centering
    \includegraphics[width=.83\linewidth]{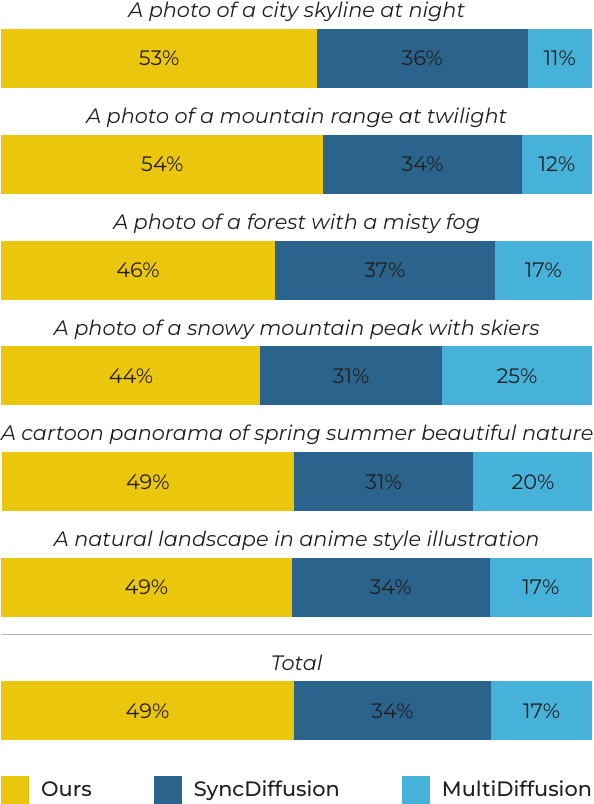}
    \vspace{-.5em}
    \caption{\strut User study per-prompt results of MAD, SyncDiffusion, and MultiDiffusion.}
    \label{fig:user_study}
\end{minipage}
\end{figure}
\clearpage
\begin{table}[h!]
\renewcommand{\arraystretch}{.95}
    \footnotesize
        \centering
        \setlength{\tabcolsep}{.28em}
        \caption{Quantitative comparison on 512$\times$3072 panorama generation using the LDM. I-StyleL, KID, and mGIQA values are scaled by 10$^{\text{3}}$. For MAD, $\tau{=}$15.}\vspace{-.8em}
        \label{tab:ldm_attns}
        \resizebox{\linewidth}{!}{
        \begin{tabular}{lc c c c c c c}
        \toprule
        && \textbf{mCLIP $\uparrow$} & \textbf{I-LPIPS $\downarrow$} & \textbf{I-StyleL $\downarrow$} & \textbf{FID $\downarrow$} & \textbf{KID $\downarrow$} & \textbf{mGIQA $\uparrow$}\\
	\midrule
    	\multicolumn{8}{c}{\textbf{Standard Prompts}}\\
    	\midrule
        \textbf{SD-L}              && 32.01${\pm}$1.67 & 0.50${\pm}$0.11 & 0.58${\pm}$0.40 & 87.64${\pm}$30.25 & 76.83${\pm}$30.52 & 27.72${\pm}$7.83  \\
        \textbf{SD-L+Attn-S}       && 32.02${\pm}$1.66 & 0.52${\pm}$0.11 & 0.73${\pm}$0.47 & 80.16${\pm}$27.11 & 67.54${\pm}$26.00 & 27.89${\pm}$7.67 \\
        \midrule
        \textbf{MAD}               && 32.03${\pm}$2.29 & 0.56${\pm}$0.10 & 1.90${\pm}$1.32 & 48.52${\pm}$17.14 & 27.15${\pm}$9.10~ & 28.32${\pm}$7.76 \\
        \textbf{MAD+Attn-S}        && 32.17${\pm}$2.20 & 0.53${\pm}$0.11 & 1.19${\pm}$0.75 & 64.08${\pm}$28.19 & 47.33${\pm}$23.77 & 28.30${\pm}$0.75 \\ 
	\midrule
    	\multicolumn{8}{c}{\textbf{GPT1k Prompts}}\\
    	\midrule
	\textbf{SD-L}              && 31.89 & 0.52 & ~0.73 & 68.03 & ~6.61 & 12.58 \\
    	\textbf{SD-L+Attn-S}   && 32.03 & 0.53 & ~0.87 & 65.08 & ~5.47 & 12.68 \\
    	\midrule
        \textbf{MAD}               && 32.47 & 0.58 & ~3.89 & 54.44 & ~1.28 & 13.03 \\
    	\textbf{MAD+Attn-S}    && 32.40 & 0.57 & ~2.94 & 55.58 & ~1.82 & 13.09 \\
        \bottomrule
        \end{tabular}
        }
        \vspace{-3.7em}
\end{table}
\begin{table}[h!]
\renewcommand{\arraystretch}{.95}
\footnotesize
    \centering
    \setlength{\tabcolsep}{.28em}
    \caption{Quantitative comparison on 512$\times$3072 panorama generation with the LCM for different numbers of inference steps. I-StyleL, KID, and mGIQA are scaled by 10$^{\text{3}}$. For MAD,  $\tau$=1/1/2 for 1/2/4 inference steps, respectively.}\vspace{-.8em}
    \label{tab:lcm_attns}
    \resizebox{.9\linewidth}{!}{
    \begin{tabular}{lc c c c c c c }
        \toprule
        && \textbf{mCLIP $\uparrow$}& \textbf{I-LPIPS $\downarrow$} & \textbf{I-StyleL $\downarrow$} & \textbf{FID $\downarrow$} & \textbf{KID $\downarrow$} & \textbf{mGIQA $\uparrow$} \\
    \midrule
    \multicolumn{8}{c}{\textbf{1 Inference Step}}\\
    \midrule
    \textbf{LCD-L}            && 29.50${\pm}$1.48 & 0.40${\pm}$0.08 & 0.21${\pm}$0.18 & 67.54${\pm}$15.53 & 66.86${\pm}$21.22 & 32.21${\pm}$4.50~ \\  
    \textbf{LCD-L+Attn-S}     && 29.77${\pm}$1.43 & 0.41${\pm}$0.08 & 0.33${\pm}$0.33 & 77.50${\pm}$14.72 & 79.76${\pm}$20.96 & 31.13${\pm}$4.15~ \\
    \midrule
    \textbf{MAD} && 29.01${\pm}$1.66 & 0.40${\pm}$0.07 & 0.37${\pm}$0.28 & 72.43${\pm}$23.75 & 75.47${\pm}$33.24 & 32.23${\pm}$4.95~ \\
    \textbf{MAD+Attn-S} && 29.36${\pm}$1.46 & 0.41${\pm}$0.07 & 0.51${\pm}$0.42 & 80.53${\pm}$24.19 & 86.79${\pm}$36.77 & 31.28${\pm}$4.75~\\
    \midrule
    \multicolumn{8}{c}{\textbf{2 Inference Steps}}\\
    \midrule
    \textbf{LCD-L}            && 30.77${\pm}$2.09 & 0.47${\pm}$0.06 & 0.56${\pm}$0.29 & 49.98${\pm}$25.20 & 44.29${\pm}$26.88 & 35.45${\pm}$11.94 \\  
    \textbf{LCD-L+Attn-S}     && 31.11${\pm}$1.95 & 0.49${\pm}$0.06 & 0.66${\pm}$0.32 & 56.37${\pm}$28.61 & 54.55${\pm}$33.00 & 34.94${\pm}$12.09 \\
    \midrule
    \textbf{MAD} && 30.97${\pm}$2.15 & 0.50${\pm}$0.06 & 0.85${\pm}$0.34 & 35.69${\pm}$17.88 & 23.74${\pm}$15.82 & 35.32${\pm}$11.49 \\
    \textbf{MAD+Attn-S} && 31.16${\pm}$2.07 & 0.50${\pm}$0.06 & 0.81${\pm}$0.31 & 36.84${\pm}$18.15 & 25.83${\pm}$16.26 & 35.68${\pm}$11.60 \\
    \midrule
    \multicolumn{8}{c}{\textbf{4 Inference Steps}}\\
    \midrule
    \textbf{LCD-L}            && 31.30${\pm}$1.63 & 0.50${\pm}$0.06 & 0.58${\pm}$0.24 & 55.52${\pm}$32.82 & 51.71${\pm}$37.79 & 35.44${\pm}$12.94 \\  
    \textbf{LCD-L+Attn-S}     && 31.60${\pm}$1.57 & 0.51${\pm}$0.06 & 0.72${\pm}$0.27 & 59.25${\pm}$32.32 & 57.57${\pm}$37.91 & 34.92${\pm}$13.27 \\
    \midrule
    \textbf{MAD} && 31.48${\pm}$1.87 & 0.52${\pm}$0.05 & 0.71${\pm}$0.26 & 38.77${\pm}$19.85 & 27.41${\pm}$16.19 & 35.23${\pm}$13.15 \\
    \textbf{MAD+Attn-S} && 31.64${\pm}$1.87 & 0.51${\pm}$0.06 & 0.68${\pm}$0.23 & 38.68${\pm}$19.10 & 26.49${\pm}$14.77 & 35.45${\pm}$13.27\\
    \bottomrule
    \end{tabular}
    }
    \vspace{-3.2em}
\end{table} 
\begin{figure}[h!]
\begin{minipage}[b]{0.49\linewidth}
\centering
 \arrayrulecolor{white}
    \scriptsize
    \setlength{\tabcolsep}{0.em}
    \renewcommand{\arraystretch}{0}
    \begin{tabular}{m{0.95em} c}
         \rotatebox[origin=l]{90}{\textbf{SD-L}} &
         \makecell{\includegraphics[width=5.8cm, height=1cm]{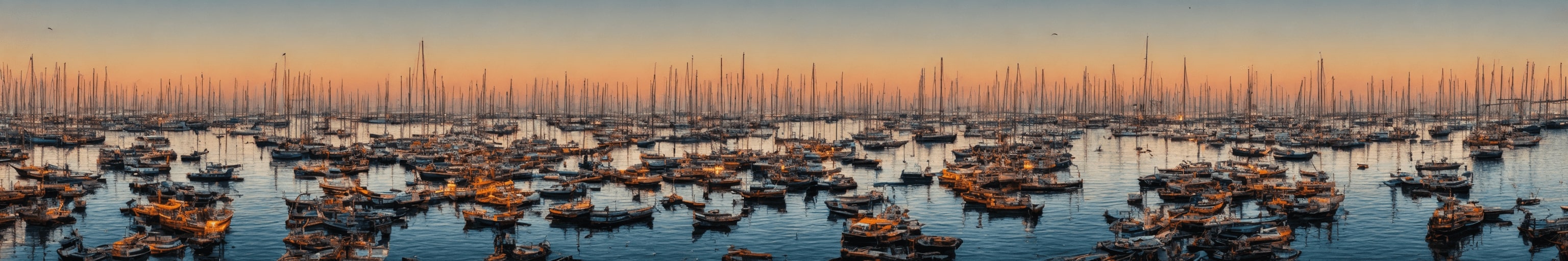}}\\
         \rotatebox[origin=l]{90}{\textbf{+AS}} &
         \makecell{\includegraphics[width=5.8cm, height=1cm]{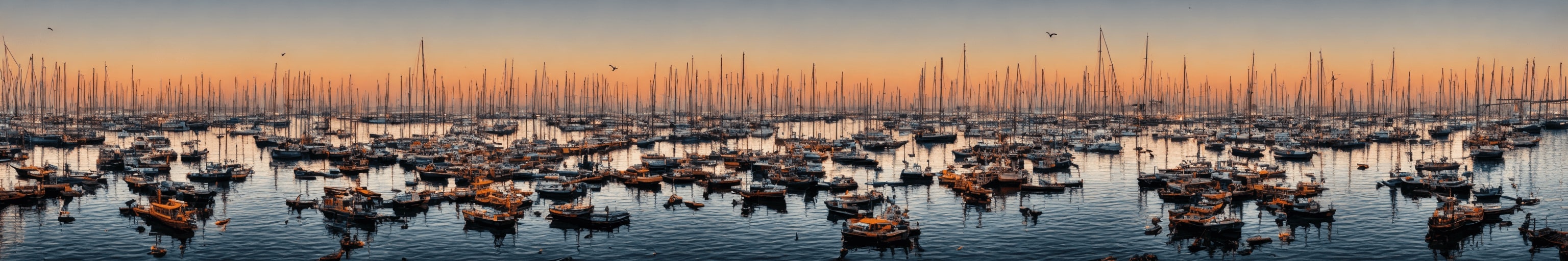}}\\
         \hline \hline
         \rotatebox[origin=l]{90}{\textbf{MAD}} & 
         \makecell{\includegraphics[width=5.8cm, height=1cm]{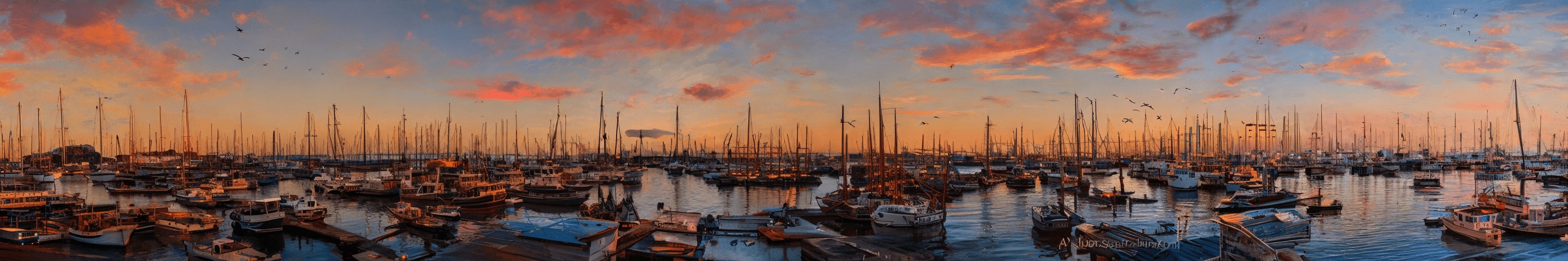}}\\
         \rotatebox[origin=l]{90}{\textbf{+AS}} &
         \makecell{\includegraphics[width=5.8cm, height=1cm]{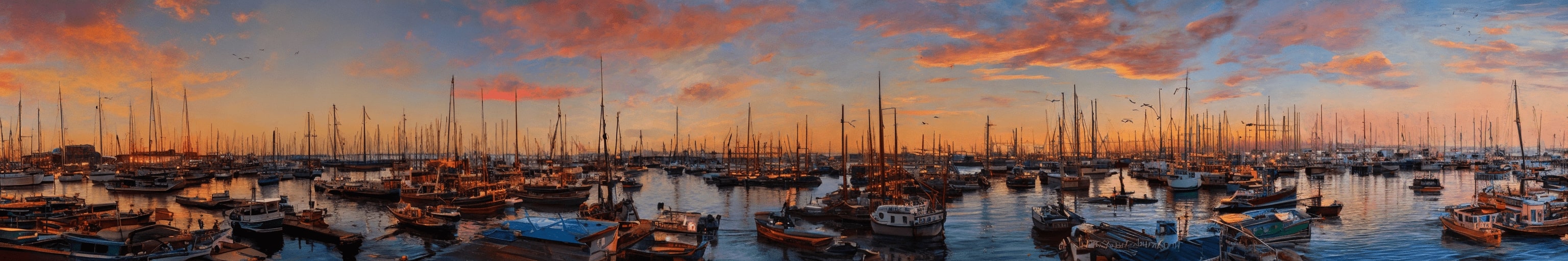}}\\
    \end{tabular}\vspace{-.8em}
\caption{\strut Effect of attention scaling (AS) on LDMs used directly and with MAD.}
\label{fig:ldm_attns}\vspace{-2.3em}
\end{minipage}
\hfill
\begin{minipage}[b]{0.49\linewidth}
    \scriptsize
    \renewcommand{\arraystretch}{0}
    \setlength{\tabcolsep}{0.em}
        \begin{tabular}{m{0.95em} c}
         \rotatebox[origin=l]{90}{\textbf{LCD-L}} &
         \makecell{\includegraphics[width=5.8cm, height=1cm]{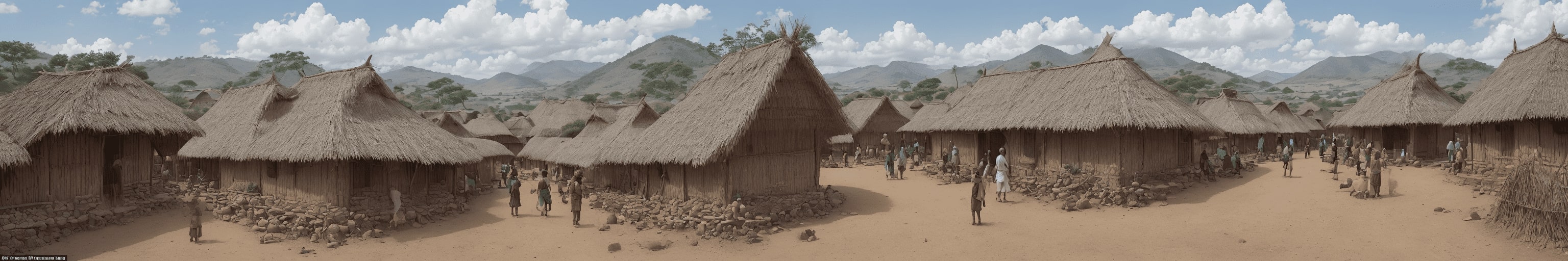}}\\
         \rotatebox[origin=l]{90}{\textbf{+AS}} &
         \makecell{\includegraphics[width=5.8cm, height=1cm]{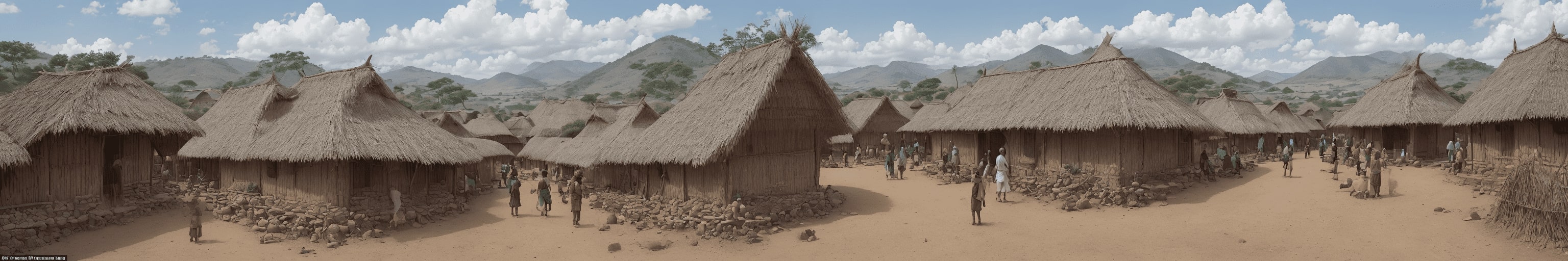}}\\  
         \hline \hline
         \rotatebox[origin=l]{90}{\textbf{MAD}} & 
         \makecell{\includegraphics[width=5.8cm, height=1cm]{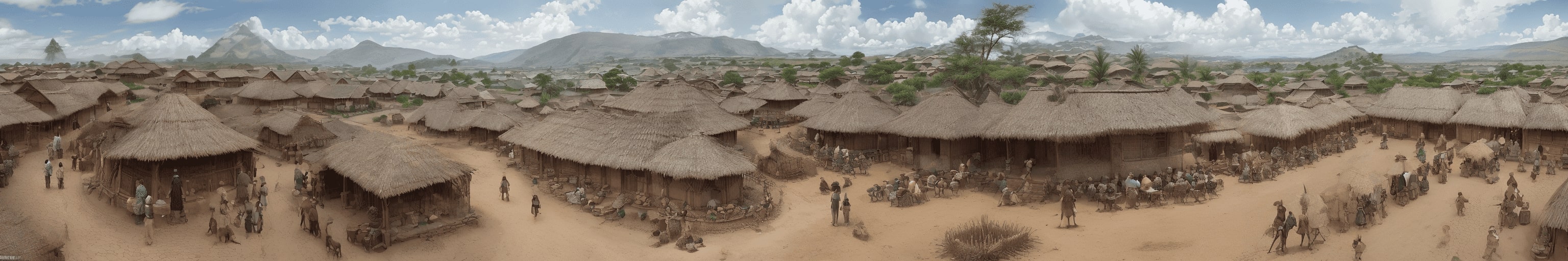}}\\
         \rotatebox[origin=l]{90}{\textbf{+AS}} &
         \makecell{\includegraphics[width=5.8cm, height=1cm]{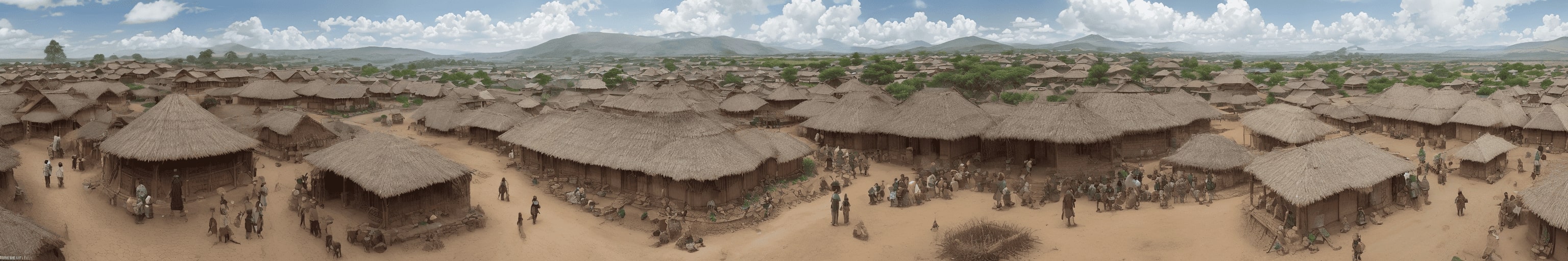}}
    \end{tabular}\vspace{-.8em}
\caption{\strut Effect of attention scaling (AS) on LCMs used directly and with MAD.}
\label{fig:lcm_attns}\vspace{-2.3em}
\end{minipage}
\end{figure}

\clearpage

\begin{figure*}[t]
    \centering
    \tiny
    \renewcommand{\arraystretch}{1.5}
    \begin{tabular}{c}
         \textit{Colorful houses facades during night with Van Gogh style} \\ 
         \makecell{\includegraphics[width=0.9\linewidth]{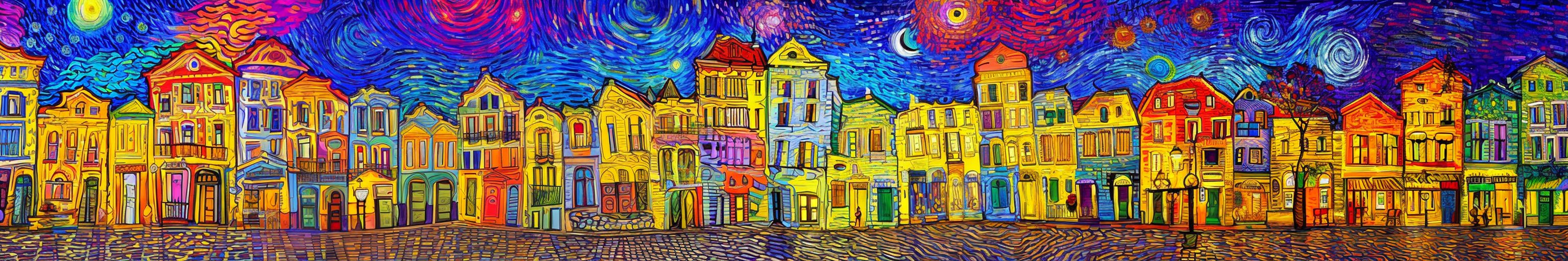}} \\
         \textit{Northern Lights at the North Pole} \\ 
         \makecell{\includegraphics[width=0.9\linewidth]{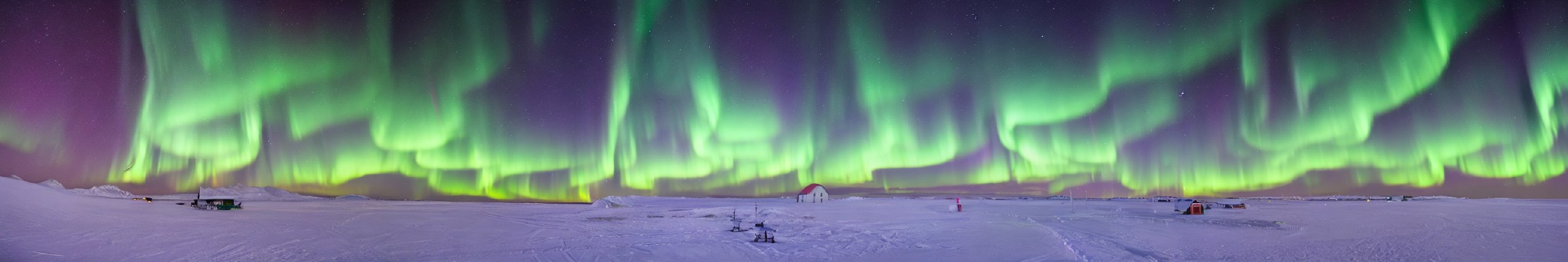}} \\
         \textit{Top view of a magma river} \\ 
         \makecell{\includegraphics[width=0.9\linewidth]{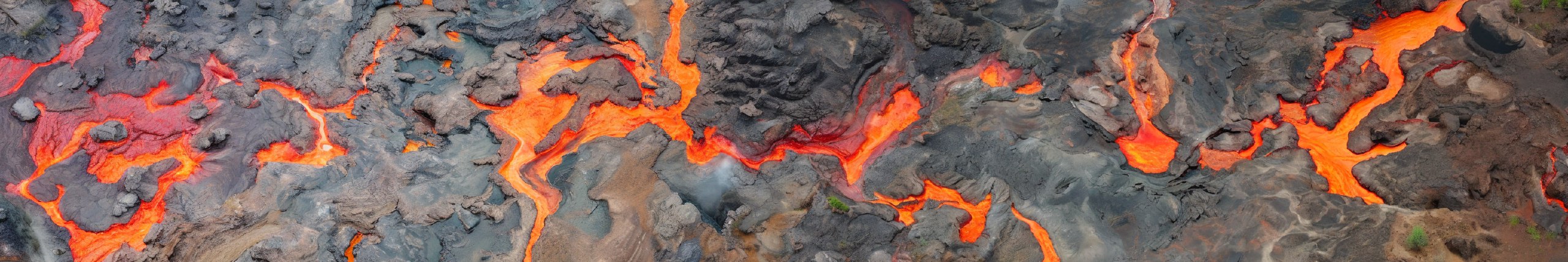}}\\
         \textit{Blossoming fruit trees in an orchard} \\ 
         \makecell{\includegraphics[width=0.9\linewidth]{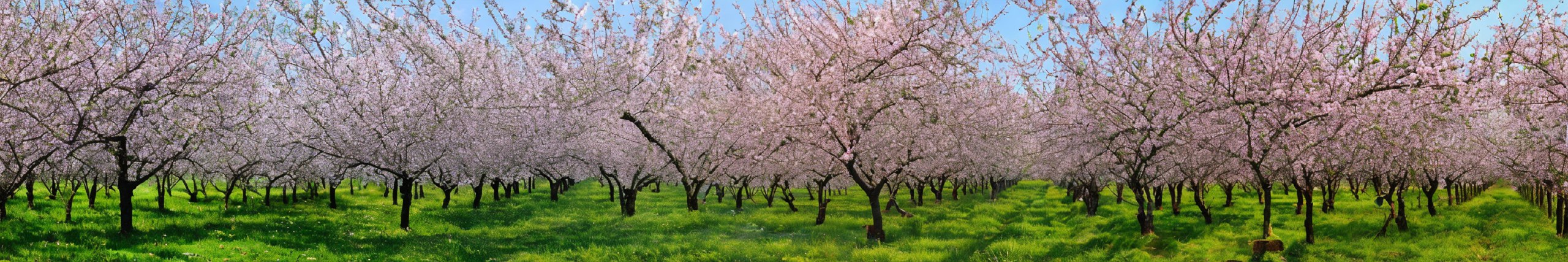}} \\
         \textit{A rose garden at dawn} \\ 
         \makecell{\includegraphics[width=0.9\linewidth]{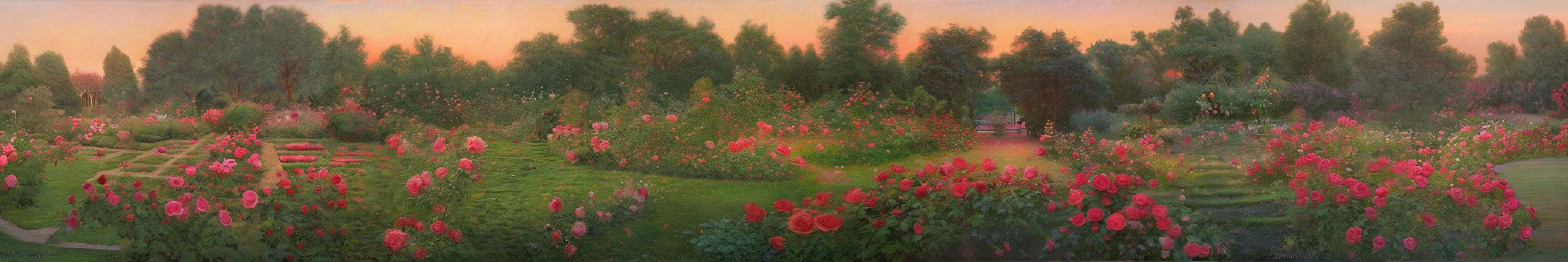}} \\
         \textit{A table with fruit } \\ 
         \makecell{\includegraphics[width=0.9\linewidth]{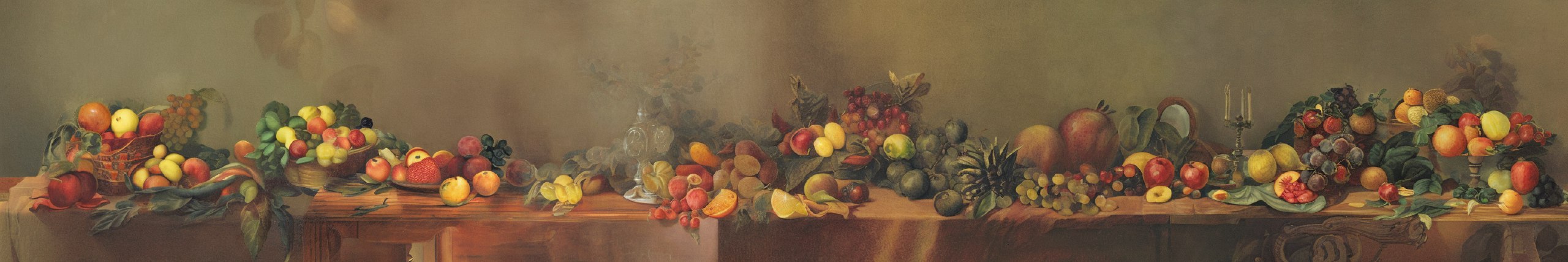}} \\
         \textit{A medieval tapestry of a battle} \\ 
         \makecell{\includegraphics[width=0.9\linewidth]{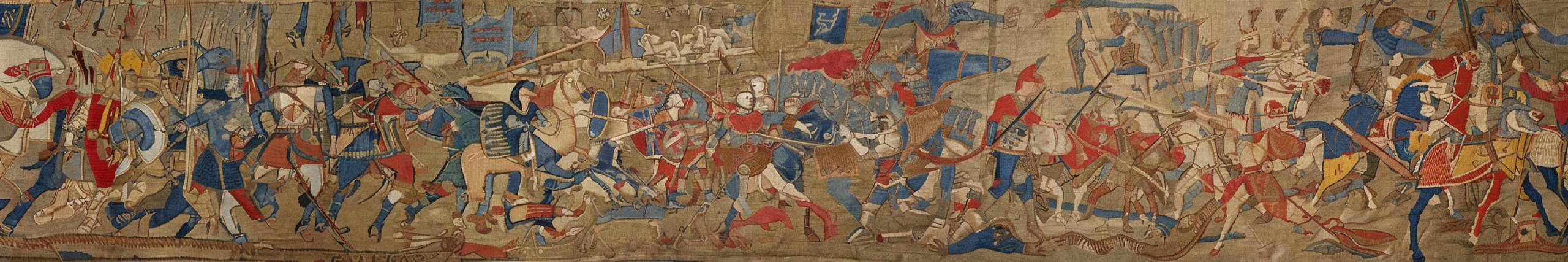}} \\
         \textit{Hieroglyphics} \\ 
         \makecell{\includegraphics[width=0.9\linewidth]{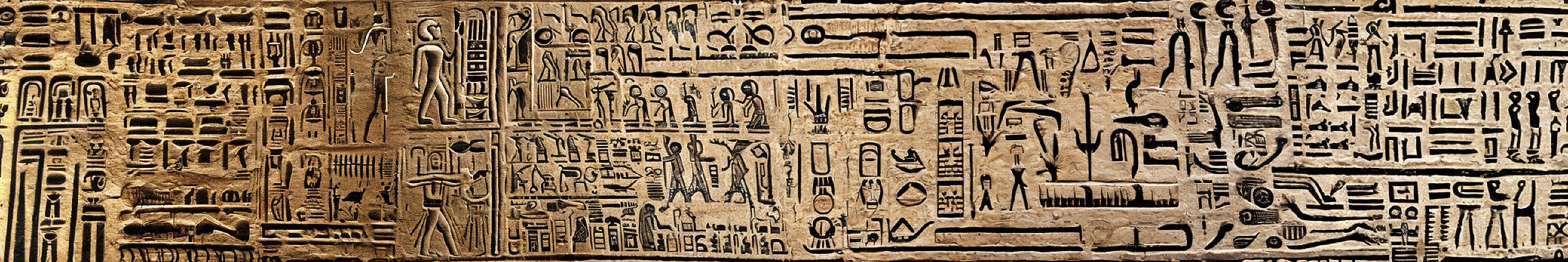}} \\
    \end{tabular}\vspace{-.75em}
\caption{Horizontal panorama images with various prompts and the LDM model.}
\label{fig:supp_qualitatives_H_ldm_additional}
\end{figure*} 

\begin{figure*}[t]
    \centering
    \tiny
    \renewcommand{\arraystretch}{1.5}
    \begin{tabular}{c}
         \textit{A blooming lavender field, with the sun setting on the horizon, creating a peaceful and colorful landscape} \\ 
         \makecell{\includegraphics[width=0.9\linewidth, height=0.15\linewidth]{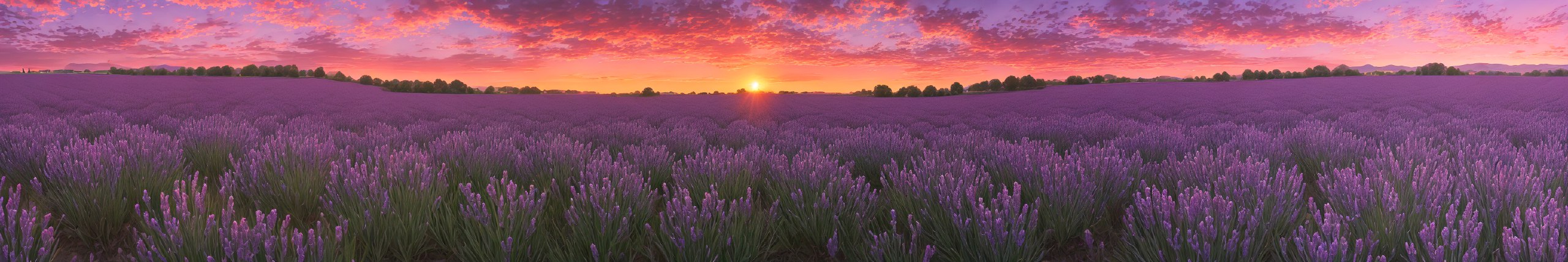}} \\
         \textit{A pink panorama at sunset with mountains and a river} \\ 
         \makecell{\includegraphics[width=0.9\linewidth, height=0.15\linewidth]{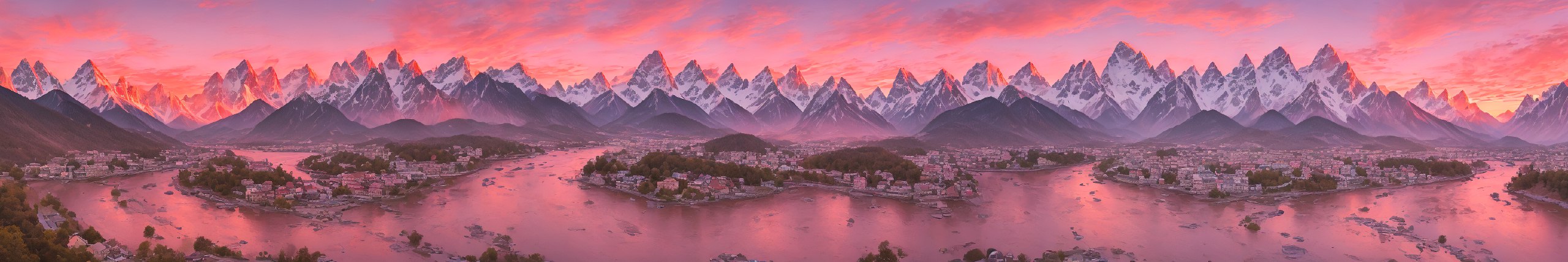}} \\
         \textit{A cove with a sandy beach and clear, calm sea under a sunset sky} \\ 
         \makecell{\includegraphics[width=0.9\linewidth, height=0.15\linewidth]{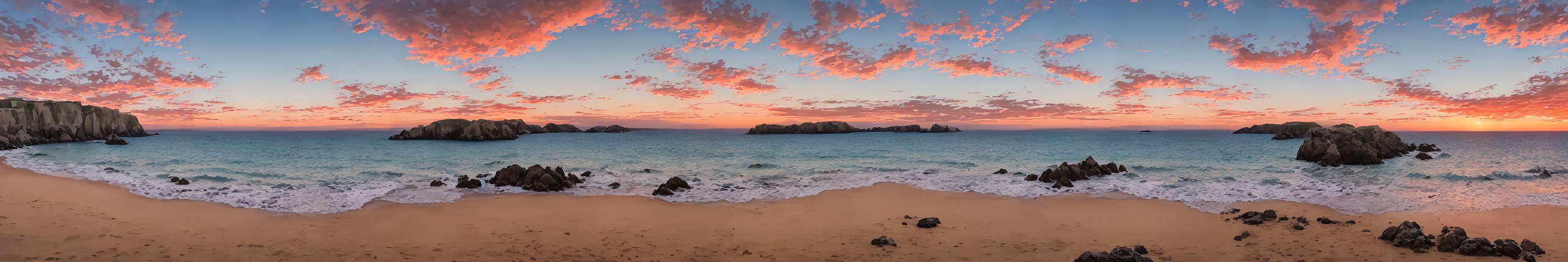}} \\
         \textit{A  serene sunrise over a misty lake, with soft colors reflecting on the water's surface} \\ 
         \makecell{\includegraphics[width=0.9\linewidth, height=0.15\linewidth]{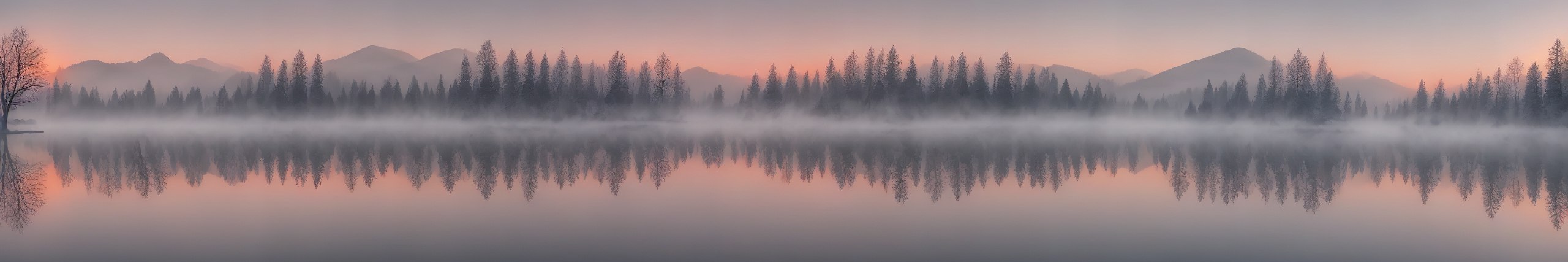}} \\
         \textit{Raging waves crashing against a rocky coast, under gray and stormy skies} \\ 
         \makecell{\includegraphics[width=0.9\linewidth, height=0.15\linewidth]{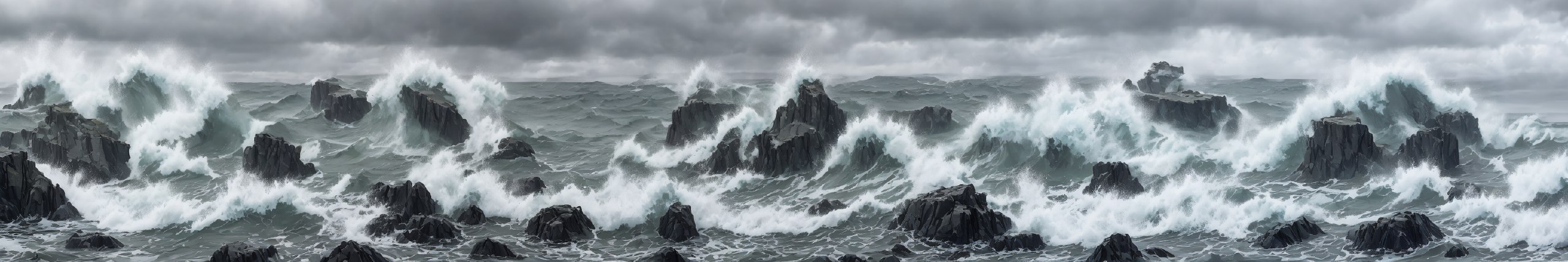}} \\
         \textit{Majestic icebergs floating in the Arctic Ocean, under a bright polar sun} \\ 
         \makecell{\includegraphics[width=0.9\linewidth, height=0.15\linewidth]{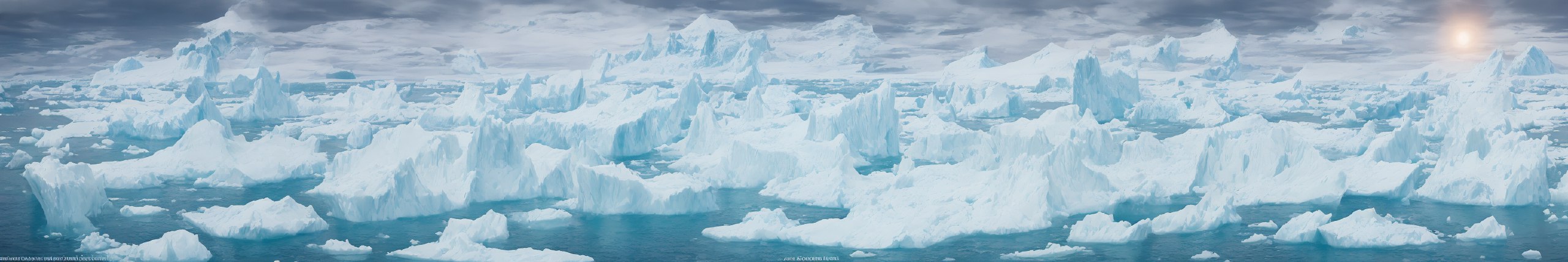}} \\
         \textit{Gentle rolling hills covered in vibrant wildflowers, under a clear blue sky} \\ 
         \makecell{\includegraphics[width=0.9\linewidth, height=0.15\linewidth]{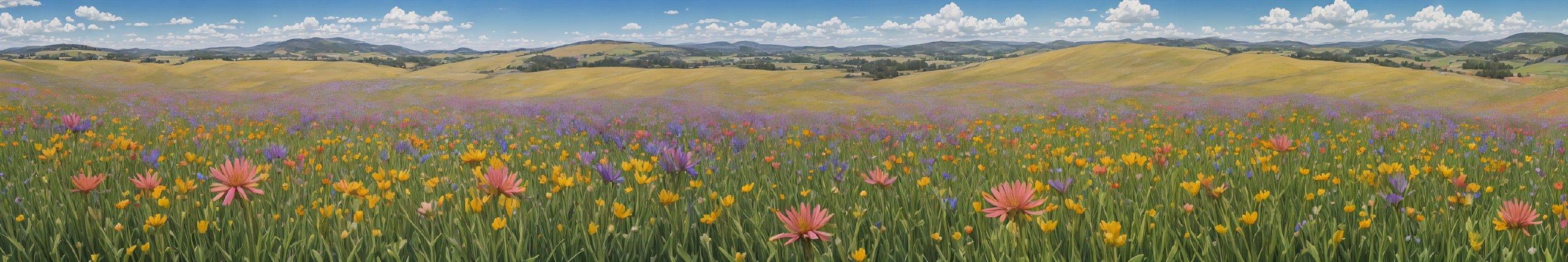}} \\
         \textit{The park of a royal palace} \\ 
         \makecell{\includegraphics[width=0.9\linewidth, height=0.15\linewidth]{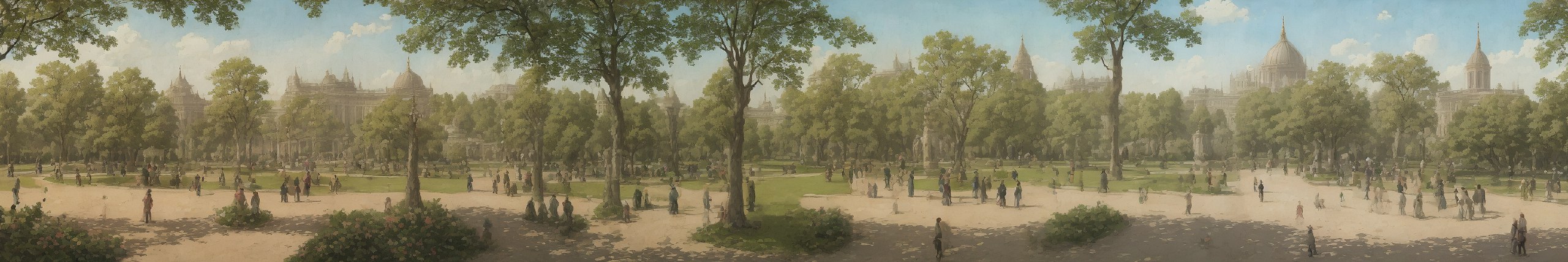}} \\
    \end{tabular}\vspace{-.75em}
\caption{Horizontal panorama images with various prompts and the LCM model.}
\label{fig:supp_qualitatives_H_lcm_additional}
\end{figure*} 

\begin{figure*}[t]
    \centering
    \tiny
    \setlength{\tabcolsep}{0.1cm}
    \begin{tabular}{m{0.05em}c m{0.05em}c m{0.05em}c m{0.05em}c m{0.05em}c m{0.05em}c}
         \rotatebox[origin=l]{90}{\textit{A cactus}} &
         \makecell{\includegraphics[width=0.125\linewidth]{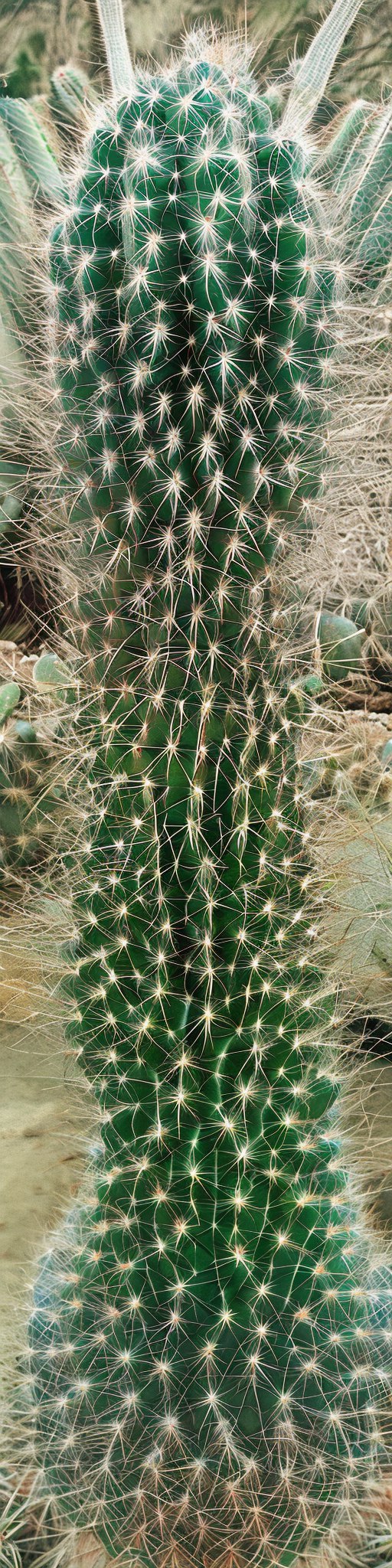}} &
         \rotatebox[origin=l]{90}{\textit{A monastery on a cliffside, overlooking a mountainous valley}} &
         \makecell{\includegraphics[width=0.125\linewidth]{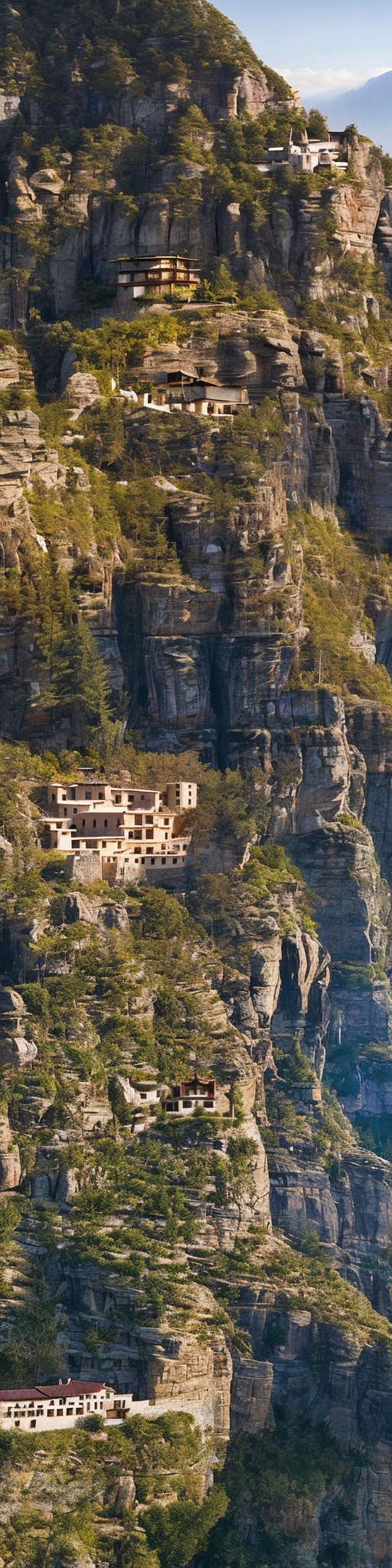}}&
         \rotatebox[origin=l]{90}{\textit{Great Wall of China}} &
         \makecell{\includegraphics[width=0.125\linewidth]{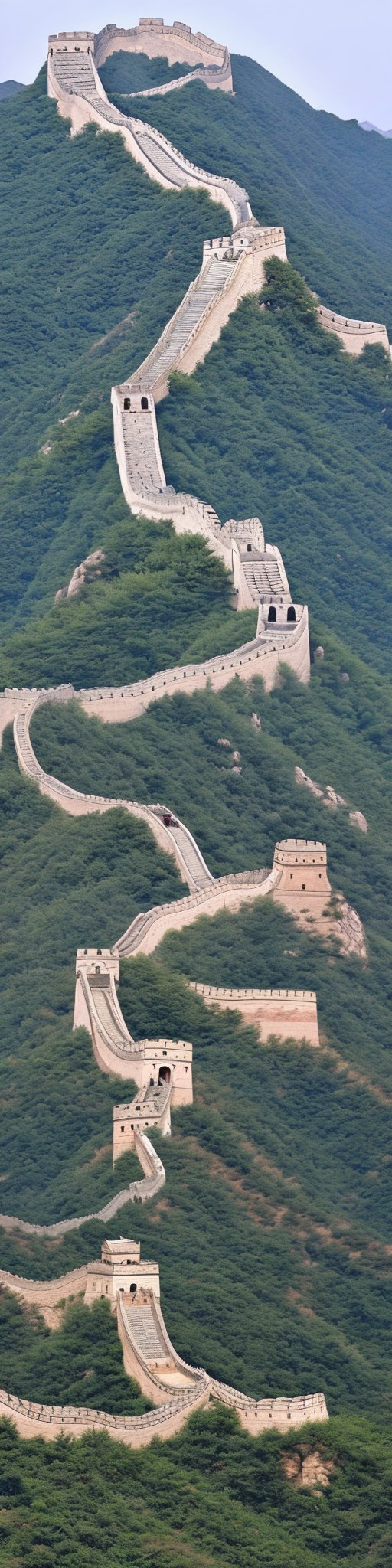}} &
         \rotatebox[origin=l]{90}{\textit{Mountain stairs}} &
         \makecell{\includegraphics[width=0.125\linewidth]{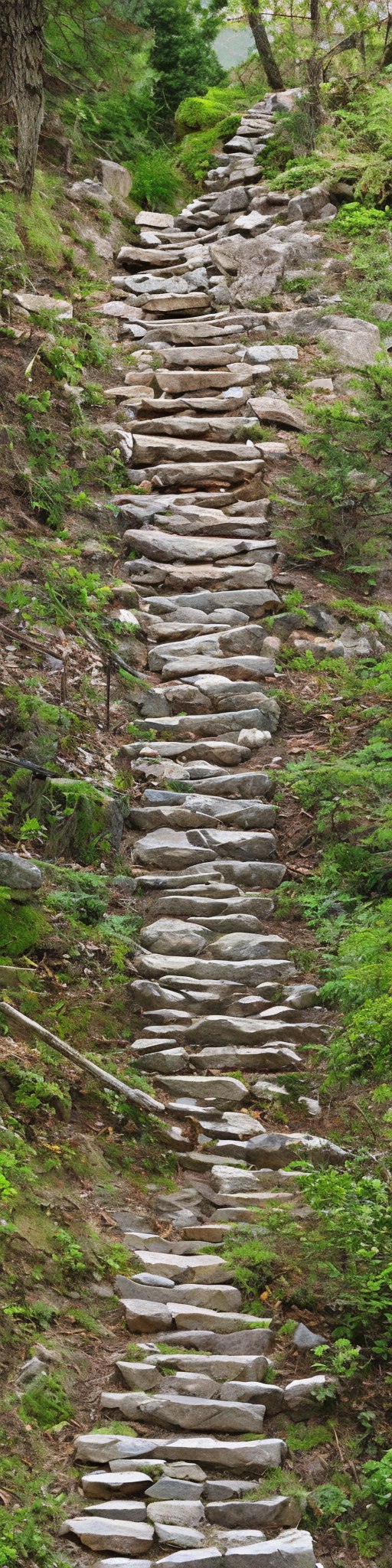}} &
         \rotatebox[origin=l]{90}{\textit{An old paper full of handwritten text}} &
         \makecell{\includegraphics[width=0.125\linewidth]{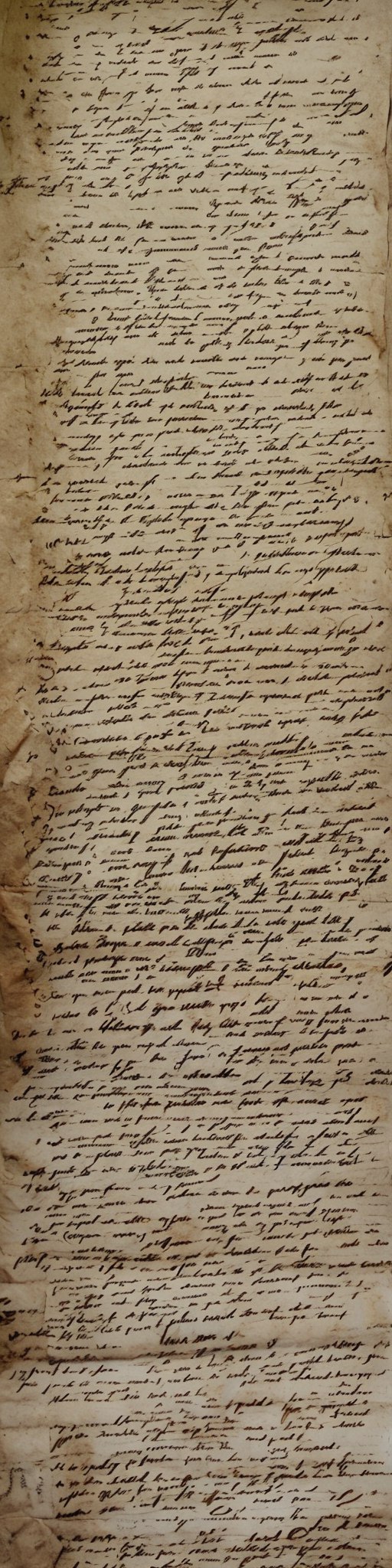}} &
         \rotatebox[origin=l]{90}{\textit{Top view of a straight road in the desert}} &
         \makecell{\includegraphics[width=0.125\linewidth]{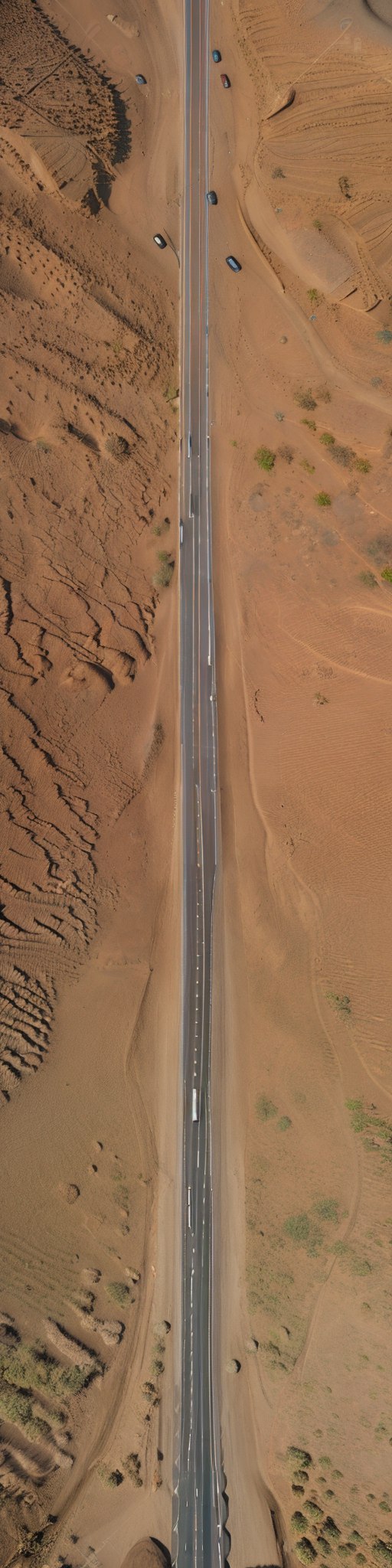}} \\
    \end{tabular}
\caption{Vertical panorama images with various prompts and the LDM model.}
\label{fig:supp_qualitatives_V_ldm_additional}
\end{figure*} 

\begin{figure*}[t]
    \centering
    \tiny
    \setlength{\tabcolsep}{0.1cm}
    \begin{tabular}{m{0.05em}c m{0.05em}c m{0.05em}c m{0.05em}c m{0.05em}c m{0.05em}c}
         \rotatebox[origin=l]{90}{\textit{A view of climbers scaling a high, steep mountain peak.}} &
         \makecell{\includegraphics[width=0.125\linewidth]{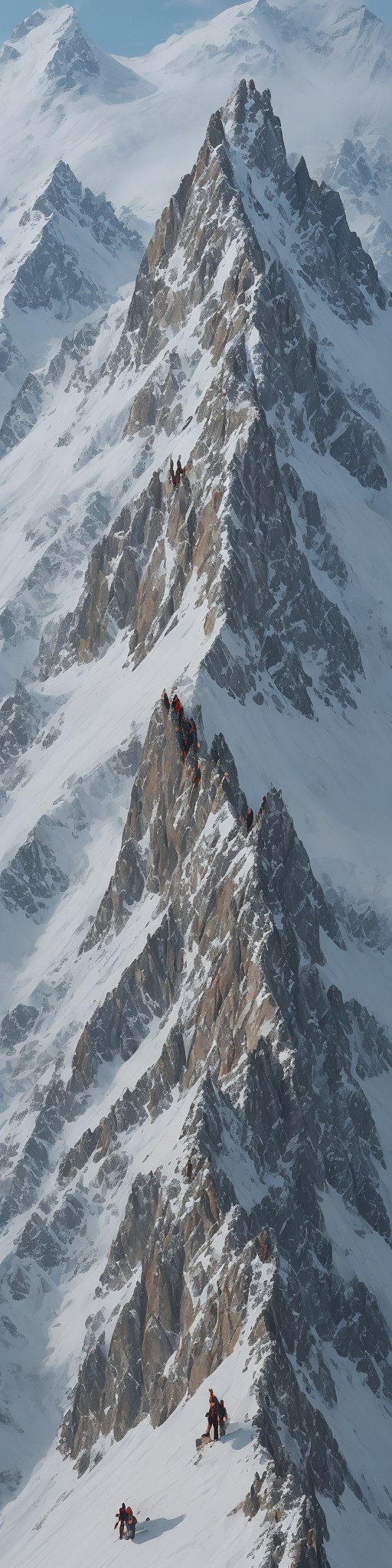}} &
         \rotatebox[origin=l]{90}{\textit{Steep vertical cliffs with birds soaring, set against a clear blue sky}} &
         \makecell{\includegraphics[width=0.125\linewidth]{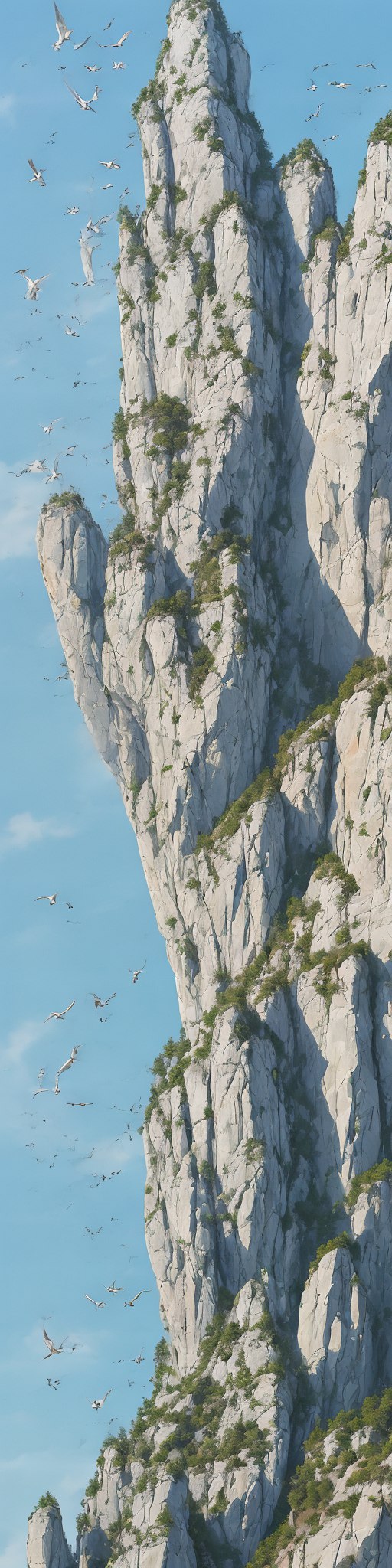}} &
         \rotatebox[origin=l]{90}{\textit{Giant redwood trees towering high, with the forest floor below}} &
         \makecell{\includegraphics[width=0.125\linewidth]{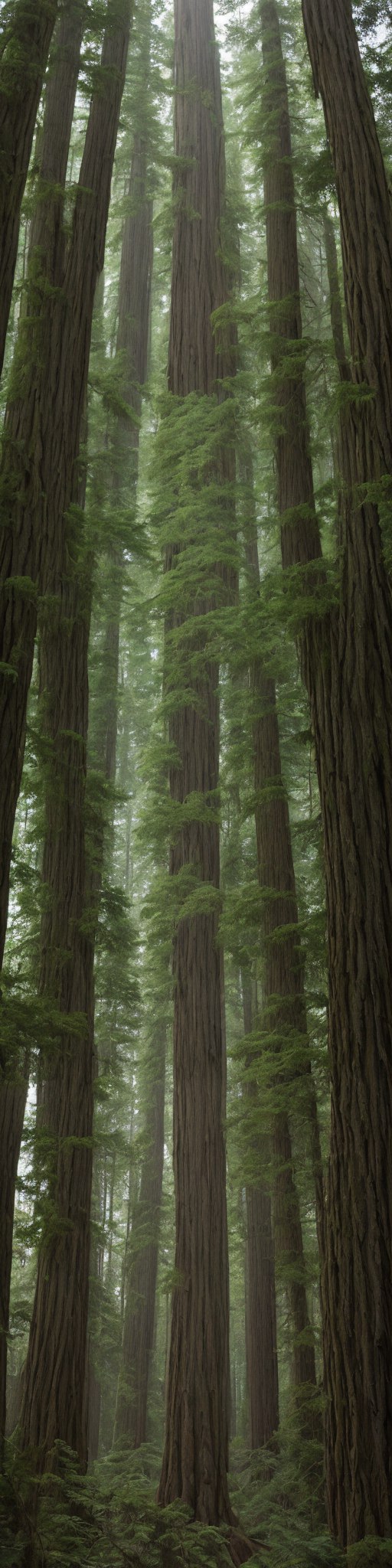}} &
         \rotatebox[origin=l]{90}{\textit{A modern building with lush vertical garden and flowers.}} &
         \makecell{\includegraphics[width=0.125\linewidth]{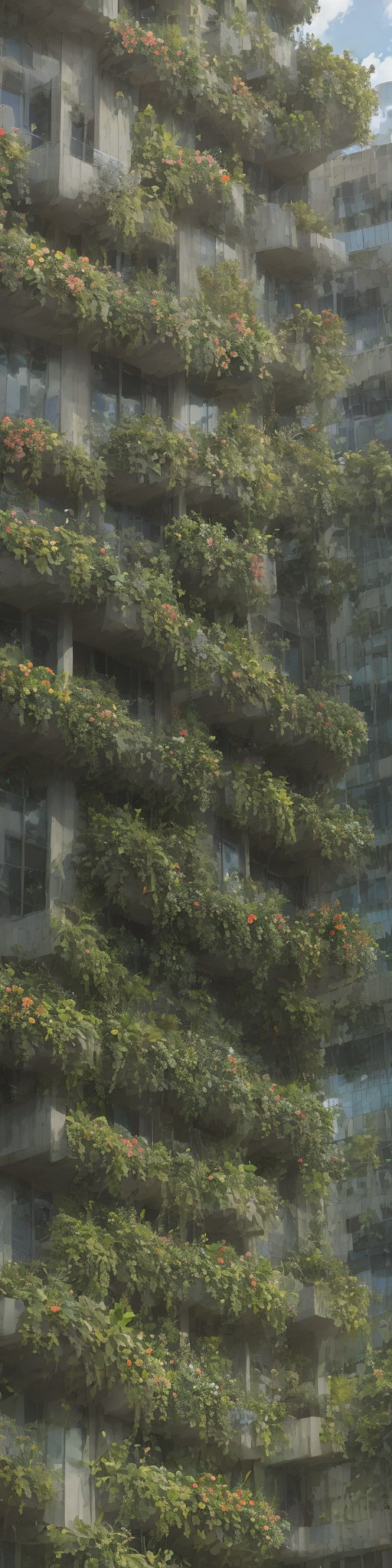}} &
         \rotatebox[origin=l]{90}{\textit{A rural landscape, with fields and a clear sky in the background}} &
         \makecell{\includegraphics[width=0.125\linewidth]{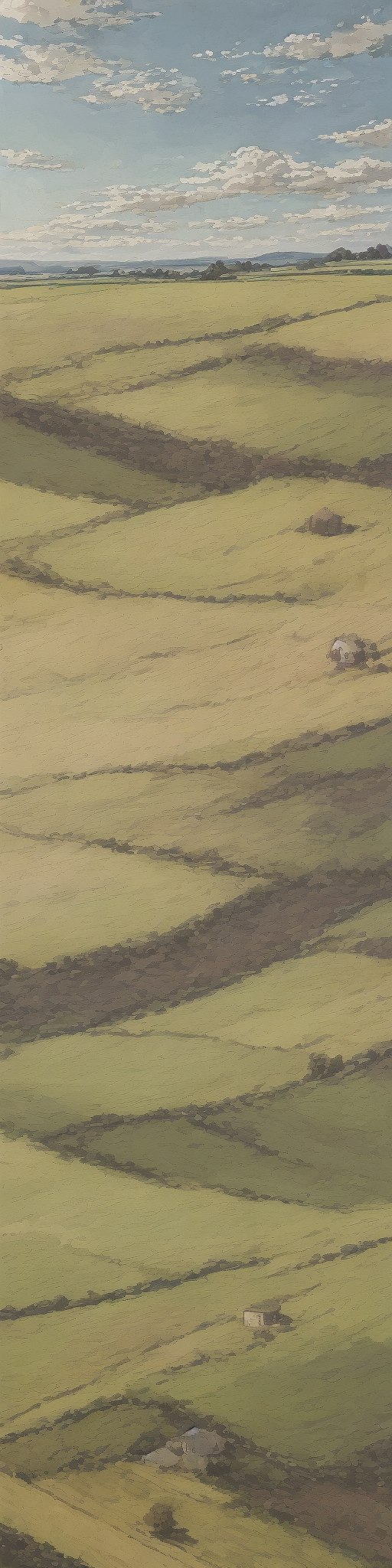}} &
         \rotatebox[origin=l]{90}{\textit{Cobbled alley in old European town with historic buildings}} &
         \makecell{\includegraphics[width=0.125\linewidth]{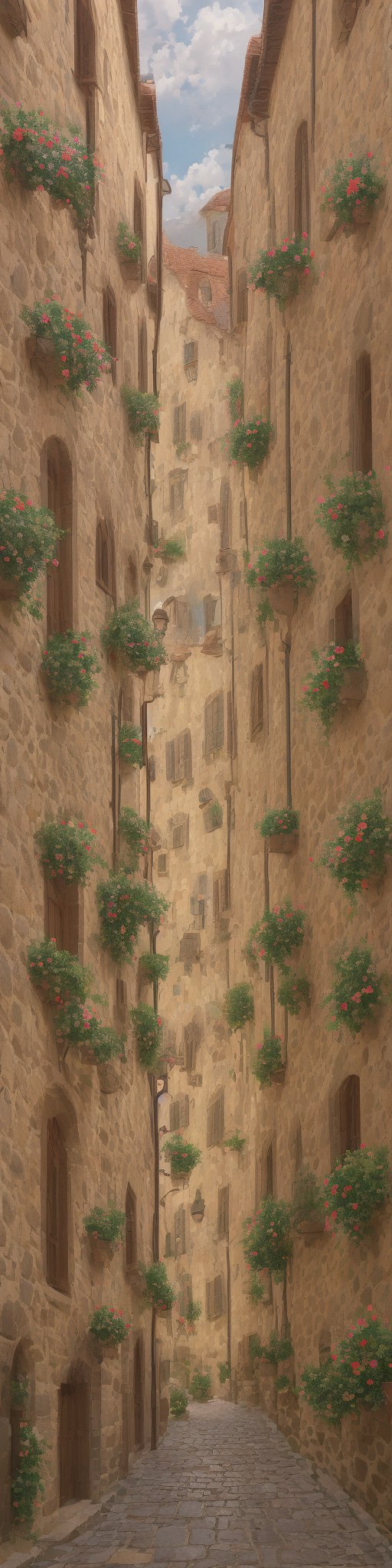}} \\
    \end{tabular}
\caption{Vertical panorama images with various prompts and the LCM model.}
\label{fig:supp_qualitatives_V_lcm_additional}
\end{figure*} 
\begin{figure}[]
\centering
    \setlength{\tabcolsep}{.0em}
    \renewcommand{\arraystretch}{.15}
    \begin{tabular}{c}
        \tiny{\textbf{Layout guided: }{\textit{A sandy flat beach, \textcolor{NavyBlue}{Rocks on the beach} \textcolor{OurOrange}{A bonfire with few logs}}}}\\
        \makecell{
        \includegraphics[width=0.5\linewidth]{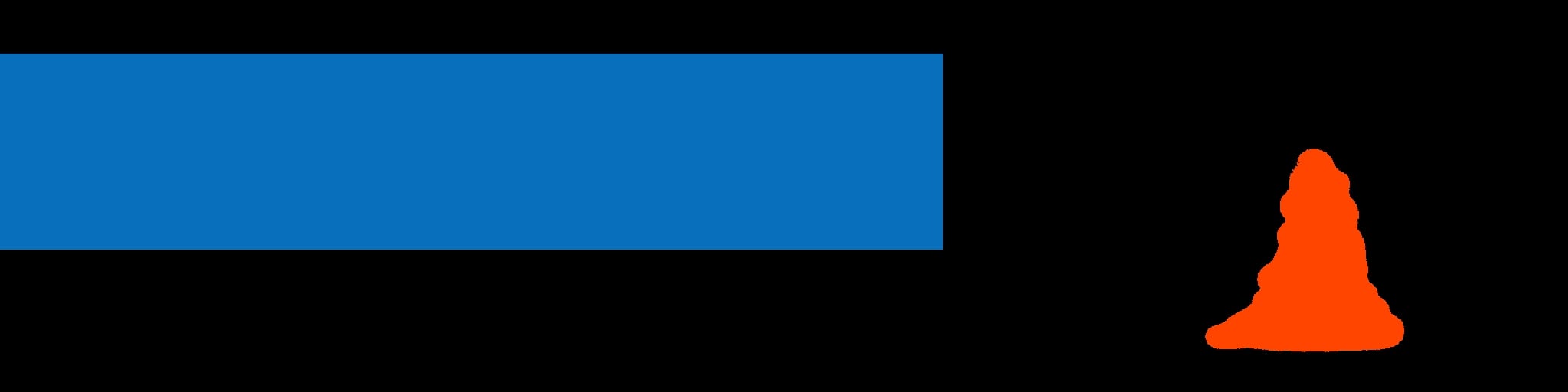}
        \includegraphics[width=0.5\linewidth]{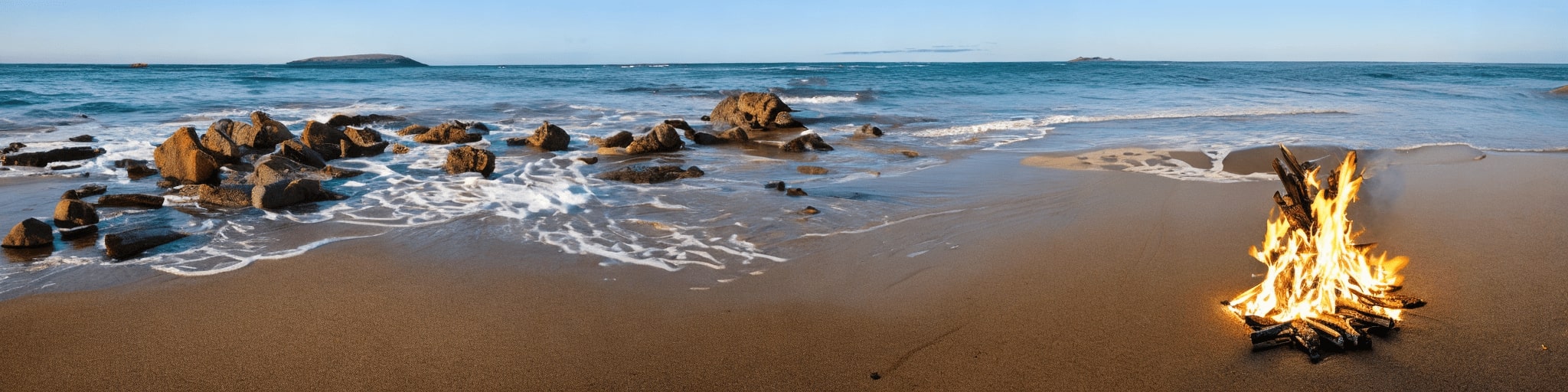}
        }\\
        \tiny{\textbf{Conditional: }{{\textit{An ancient building in the style of an oil painting.}}}}\\
        \makecell{
        \includegraphics[width=0.5\linewidth,height=0.125\linewidth]{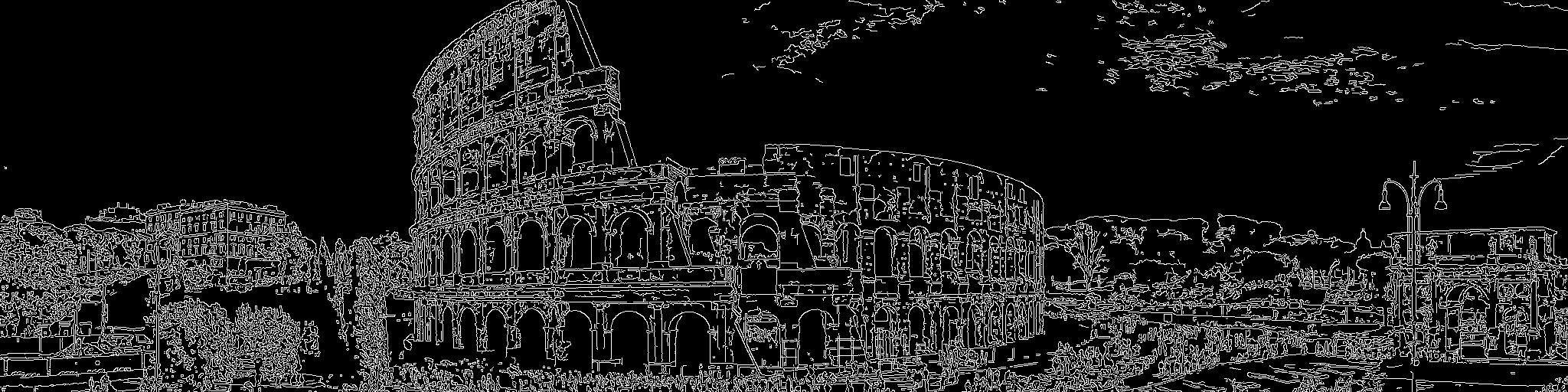}
        \includegraphics[width=0.5\linewidth,height=0.125\linewidth]{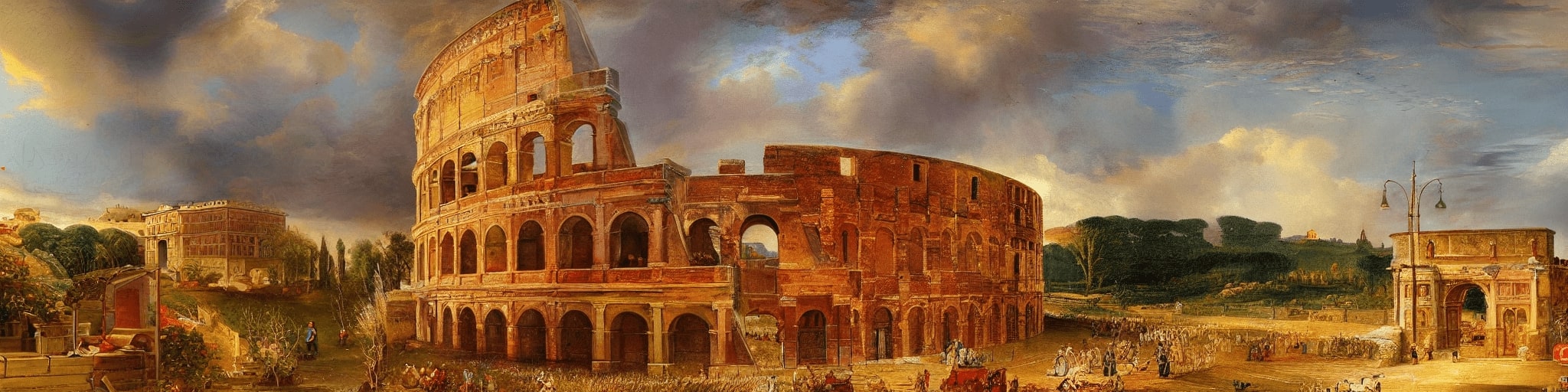}
        }    
    \end{tabular} \vspace{-.5em}
\caption{Plug\&Play applications of MAD.\vspace{-.75em}}
\label{fig:applications}
\end{figure}
\begin{figure*}[t]
\centering
\scriptsize
    \setlength{\tabcolsep}{.2em}
    \begin{tabular}{m{.5em}c}
        & \tiny{\textit{A row of colorful houses on an Amsterdam canal}}\\
        \rotatebox[origin=l]{90}{\textbf{SD1.5}} & \makecell{\includegraphics[height=1.6cm, width=9.6cm]{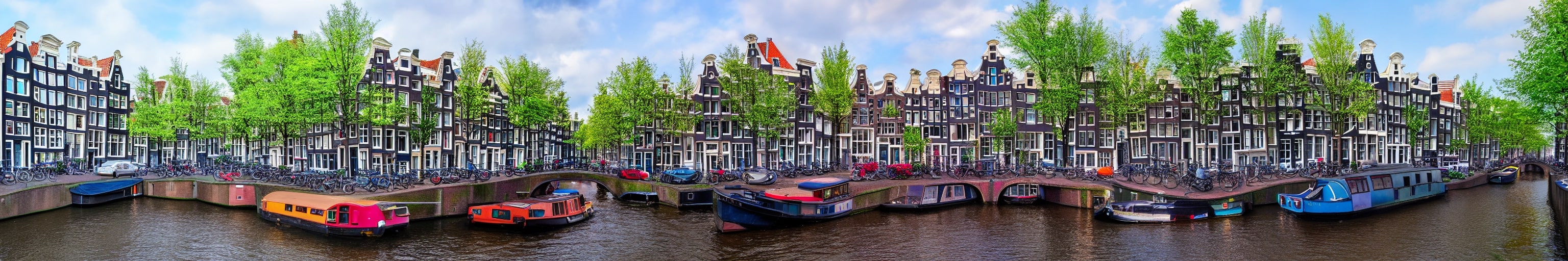}}\\
        \rotatebox[origin=l]{90}{\textbf{SD2.0}} & \makecell{\includegraphics[height=1.6cm, width=9.6cm]{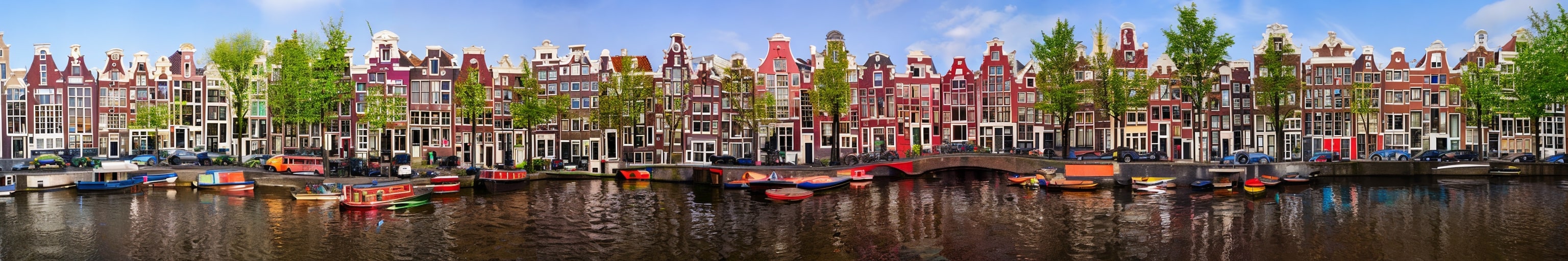}}\\
        \rotatebox[origin=l]{90}{\textbf{SD2.1}} & \makecell{\includegraphics[height=1.6cm, width=9.6cm]{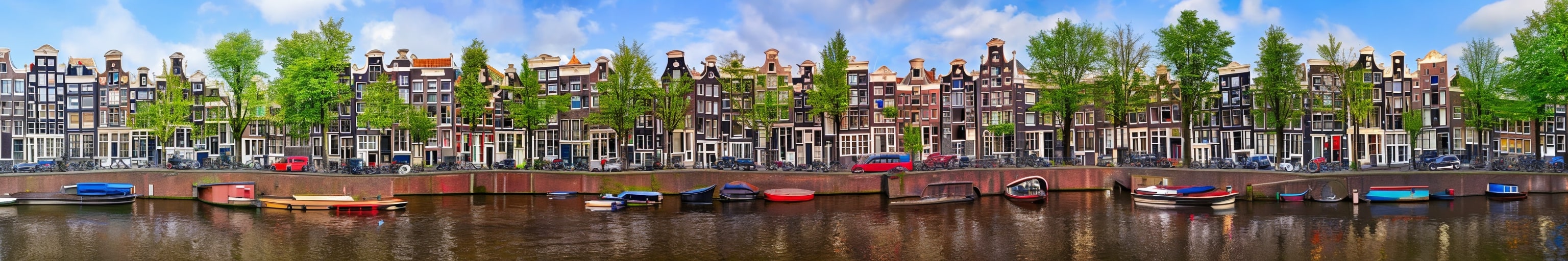}}\\
         \rotatebox[origin=l]{90}{\textbf{SDXL1.0}} & \makecell{\includegraphics[height=1.6cm, width=9.6cm]{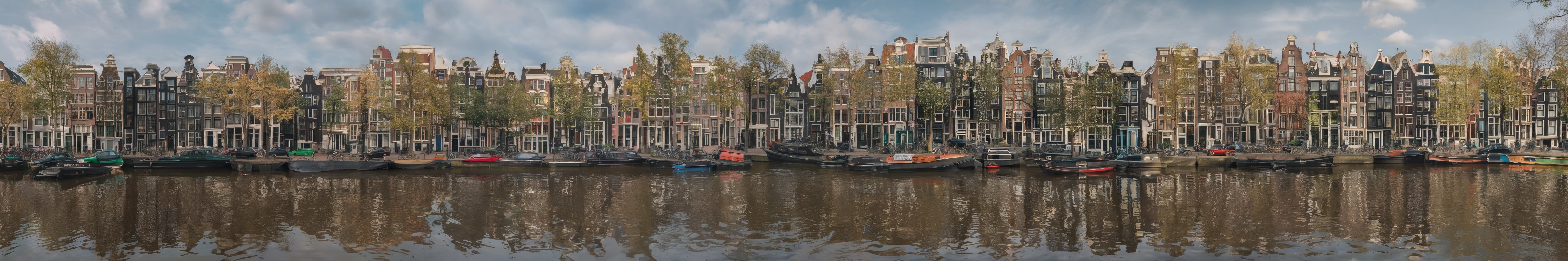}}\\
         & \tiny{\textit{A fancy bathroom}}\\
        \rotatebox[origin=l]{90}{\textbf{SD1.5}} & \makecell{\includegraphics[height=1.6cm, width=9.6cm]{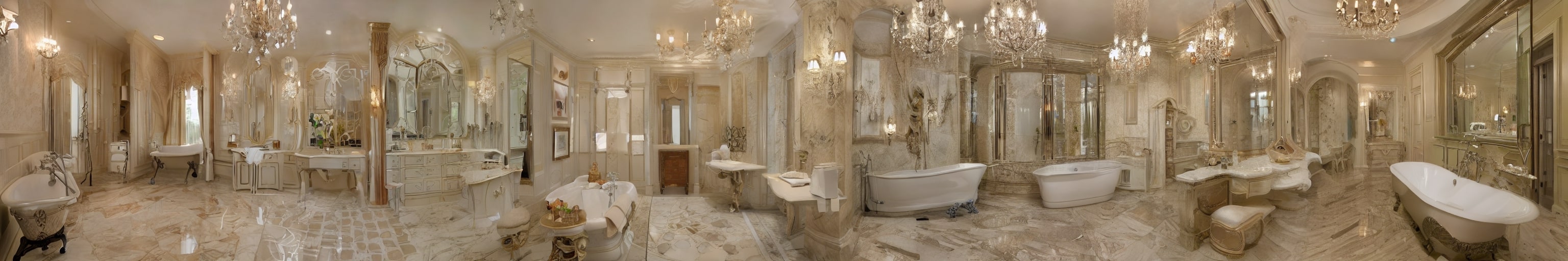}}\\
        \rotatebox[origin=l]{90}{\textbf{SD2.0}} & \makecell{\includegraphics[height=1.6cm, width=9.6cm]{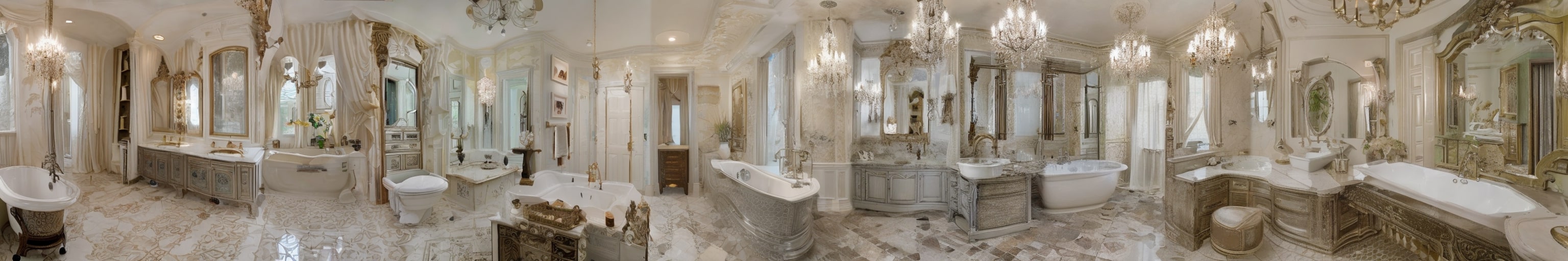}}\\
        \rotatebox[origin=l]{90}{\textbf{SD2.1}} & \makecell{\includegraphics[height=1.6cm, width=9.6cm]{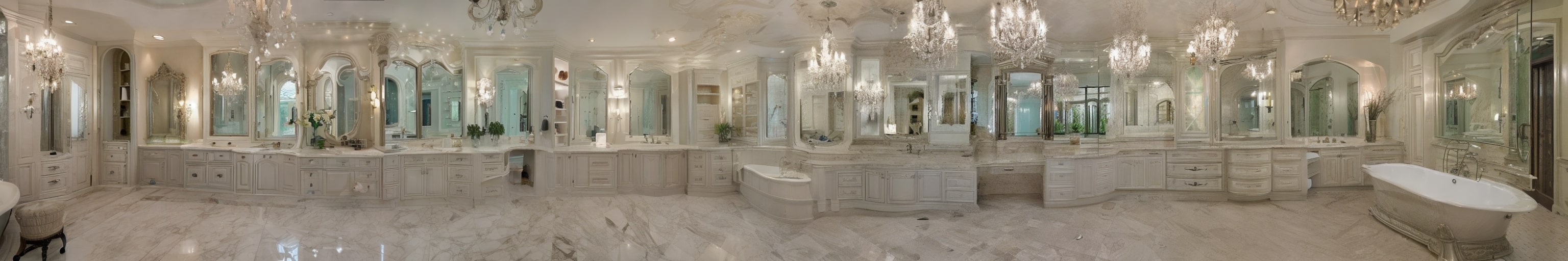}}\\
         \rotatebox[origin=l]{90}{\textbf{SDXL1.0}} & \makecell{\includegraphics[height=1.6cm, width=9.6cm]{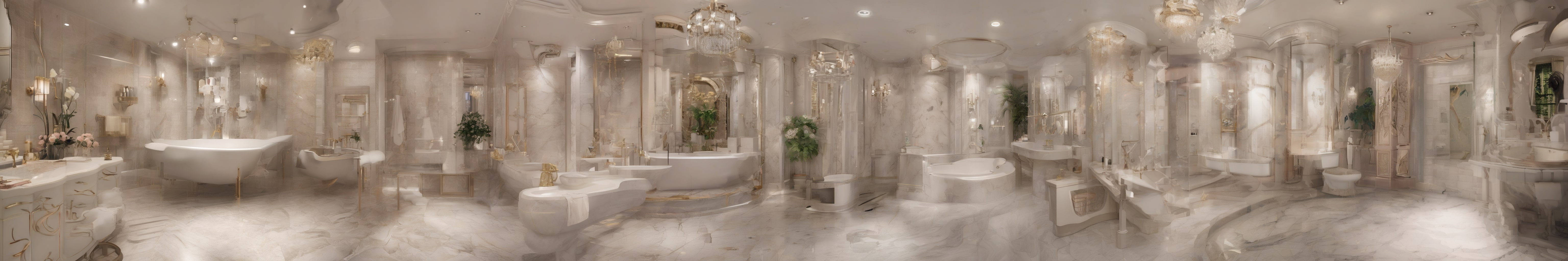}}\\
    \end{tabular}\vspace{-.5em}
\caption{Images generated with MAD applied to different LDM backbones and resolution 512$\times$3072. For SDXL1.0~\cite{podell2023sdxl}, the resolution is 1024$\times$6144.}
\label{fig:backbones}
\vspace{-2.2em}
\end{figure*}

\begin{figure*}[t]
\centering
\scriptsize
    \arrayrulecolor{white}  
    \setlength{\tabcolsep}{0.1cm}
    \begin{tabular}{m{0.25em} r}
    \rotatebox[origin=l]{90}{\textbf{SD-L}} &
     \makecell{
     \includegraphics[height=1.6cm, width=9.6cm ]{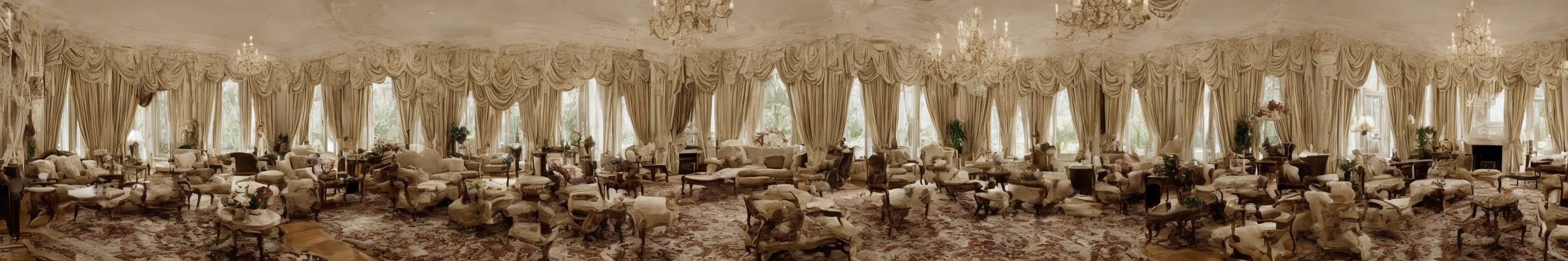}\\
     \includegraphics[height=1.6cm, width=6.35cm]{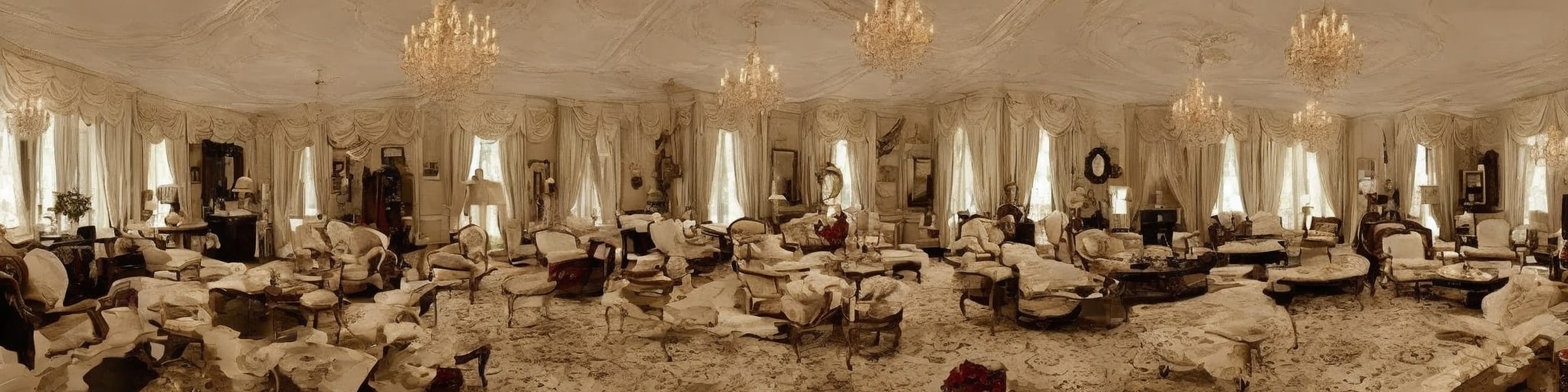}
     \includegraphics[height=1.6cm, width=3.15cm]{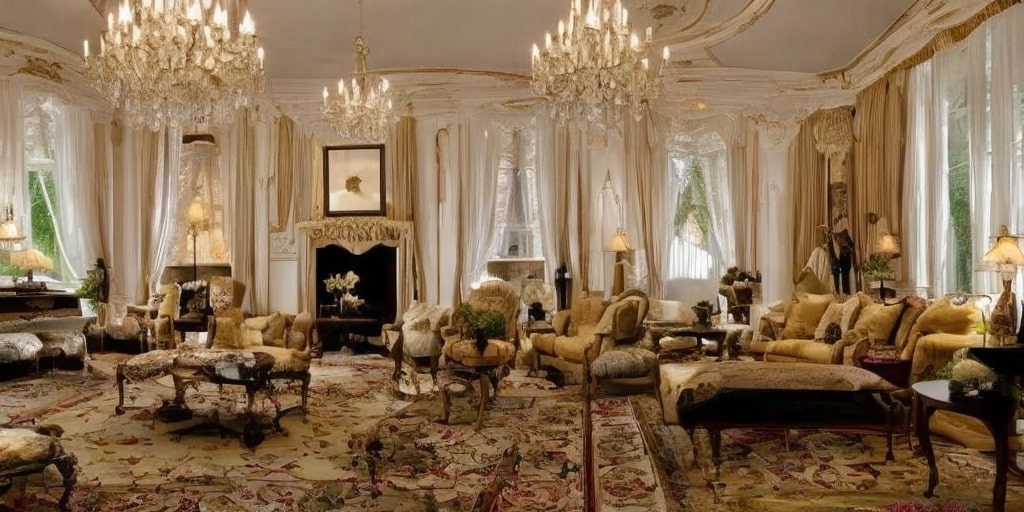}} \\
    \hline
    \hline  
    \rotatebox[origin=l]{90}{\textbf{SD-L+AttnS}} &
    \makecell{
     \includegraphics[height=1.6cm, width=9.6cm ]{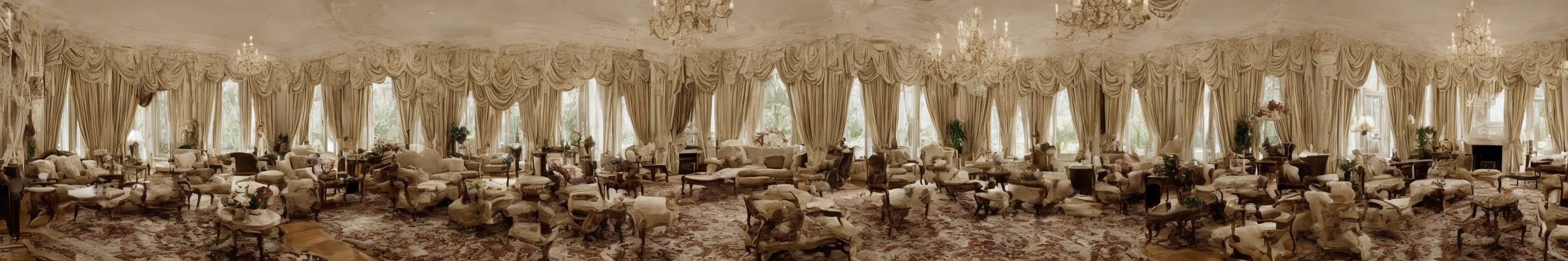}\\
     \includegraphics[height=1.6cm, width=6.35cm]{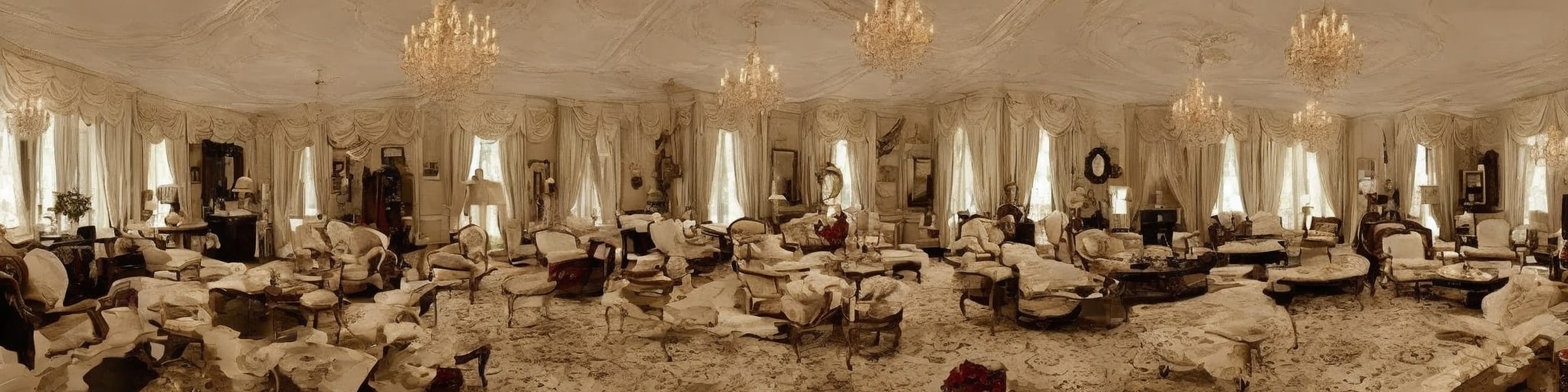}
     \includegraphics[height=1.6cm, width=3.15cm]{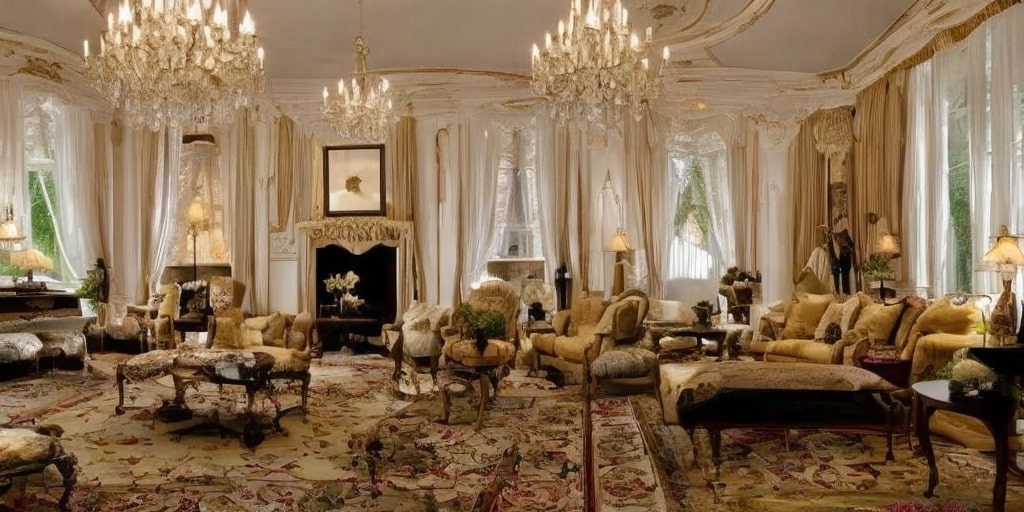}} \\
    \hline
    \hline  
    \rotatebox[origin=l]{90}{\textbf{MD}} &
    \makecell{
     \includegraphics[height=1.6cm, width=9.6cm ]{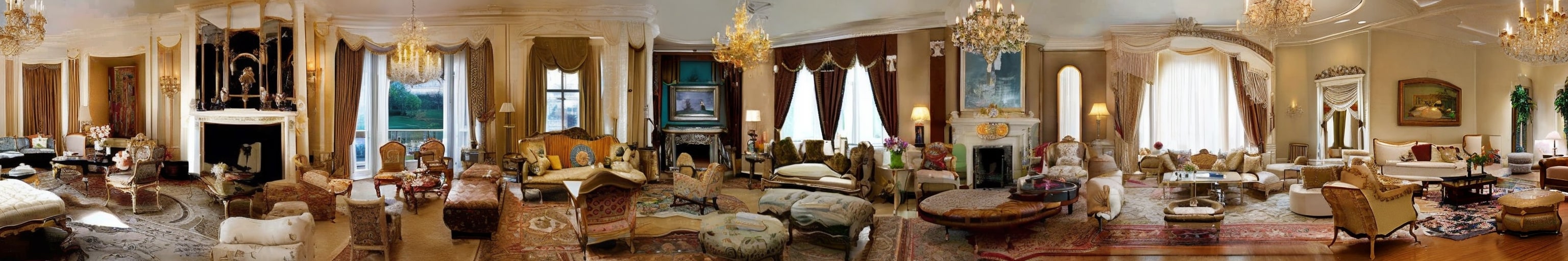}\\
     \includegraphics[height=1.6cm, width=6.35cm]{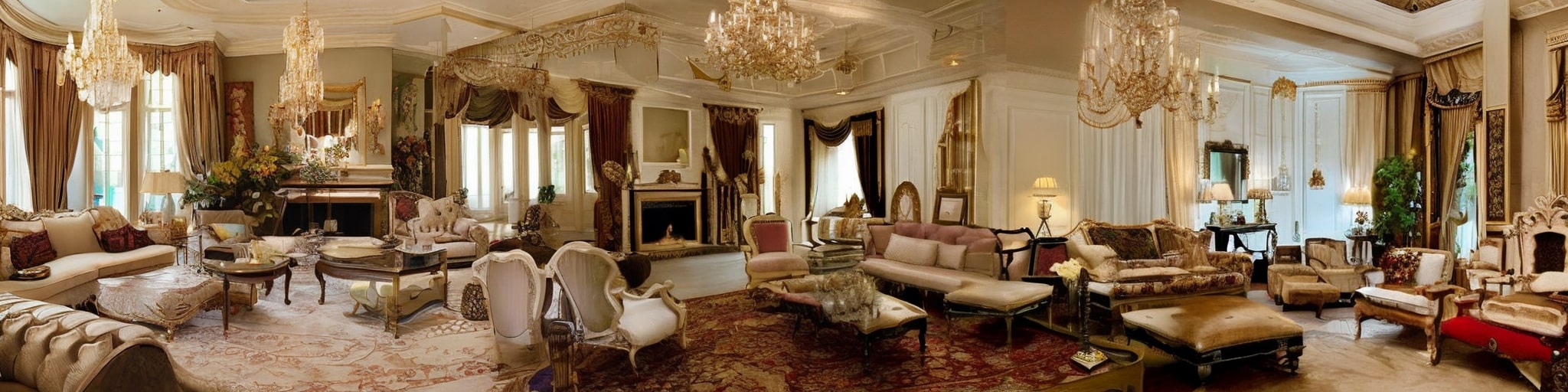}
     \includegraphics[height=1.6cm, width=3.15cm]{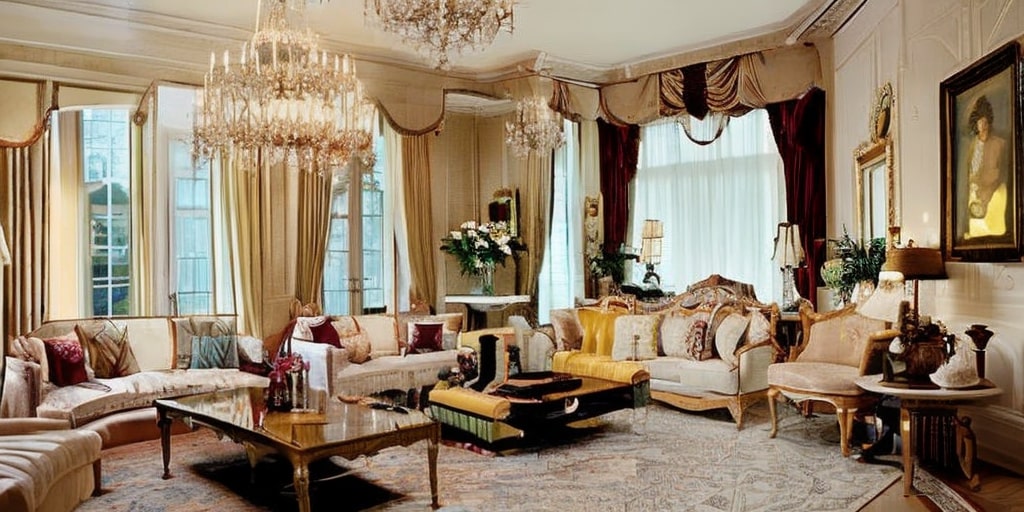}} \\
    \hline
    \hline
    \rotatebox[origin=l]{90}{\textbf{SyncD}} &
    \makecell{
     \includegraphics[height=1.6cm, width=9.6cm ]{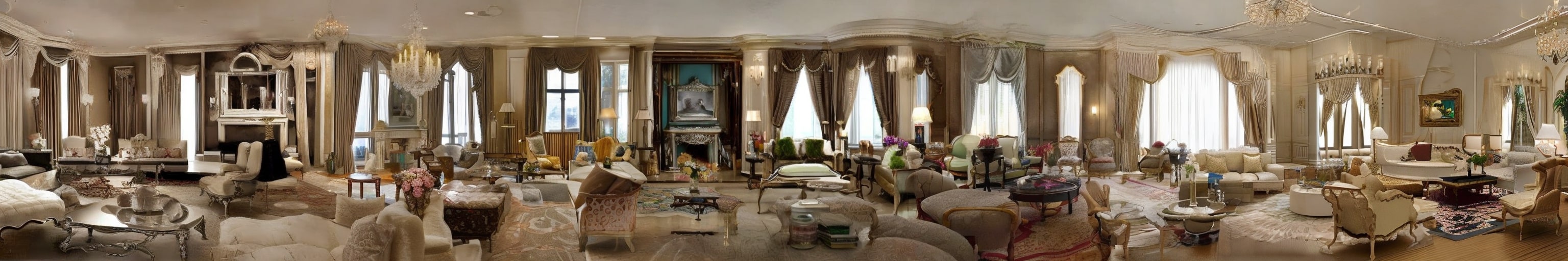}\\
     \includegraphics[height=1.6cm, width=6.35cm]{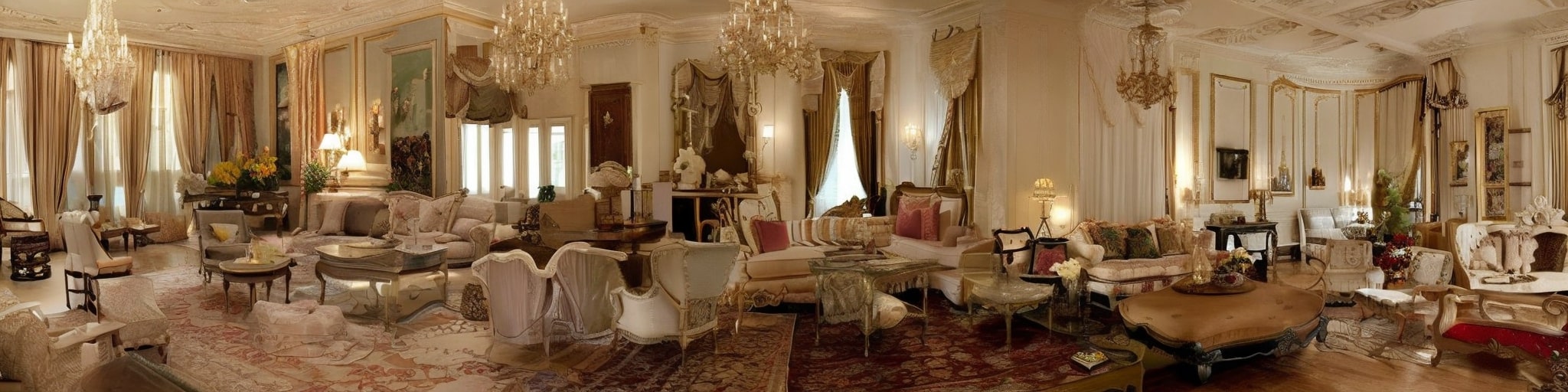}
     \includegraphics[height=1.6cm, width=3.15cm]{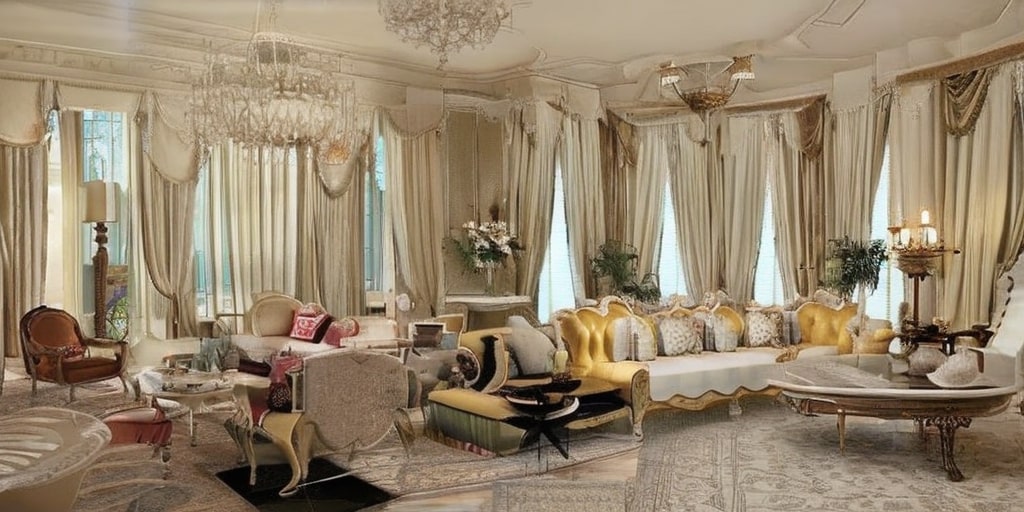}} \\
    \hline
    \hline  
     \rotatebox[origin=l]{90}{\textbf{MAD}} & 
     \makecell{
     \includegraphics[height=1.6cm, width=9.6cm ]{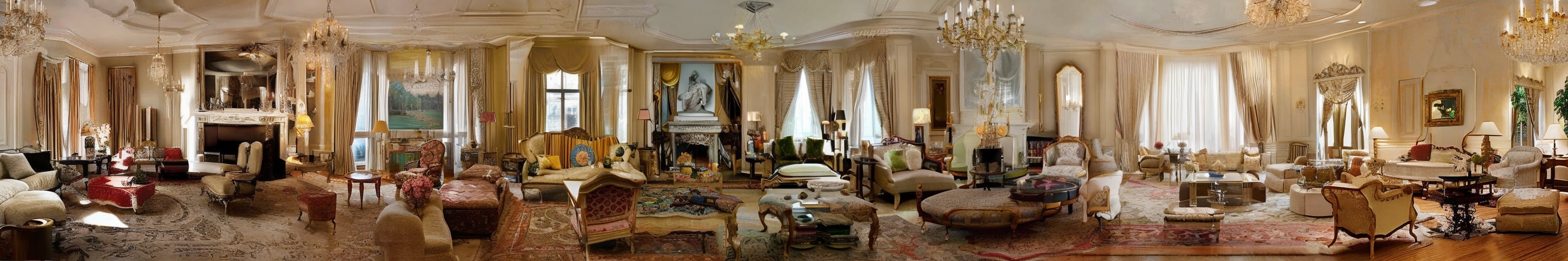}\\
     \includegraphics[height=1.6cm, width=6.35cm]{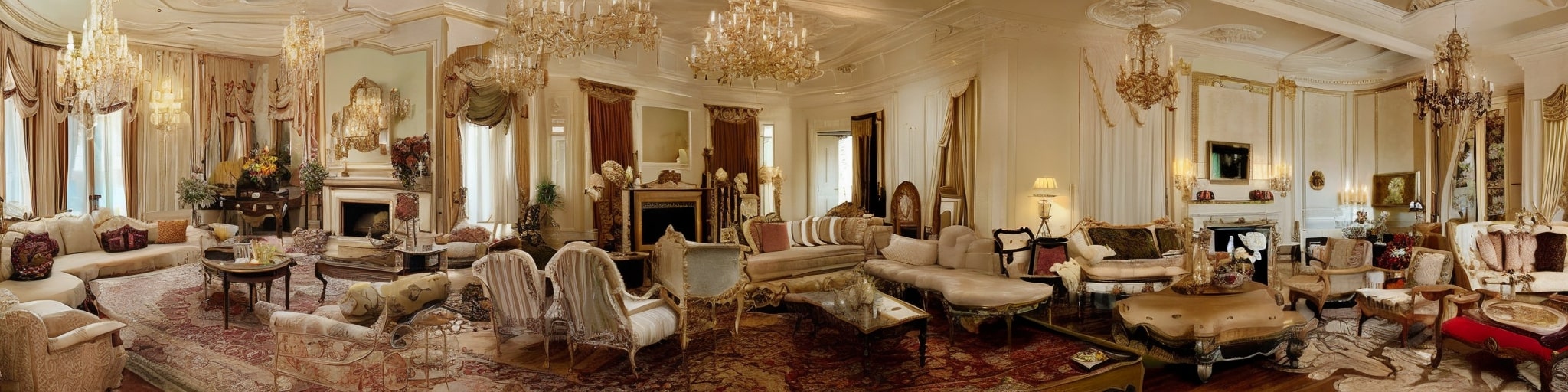}
     \includegraphics[height=1.6cm, width=3.15cm]{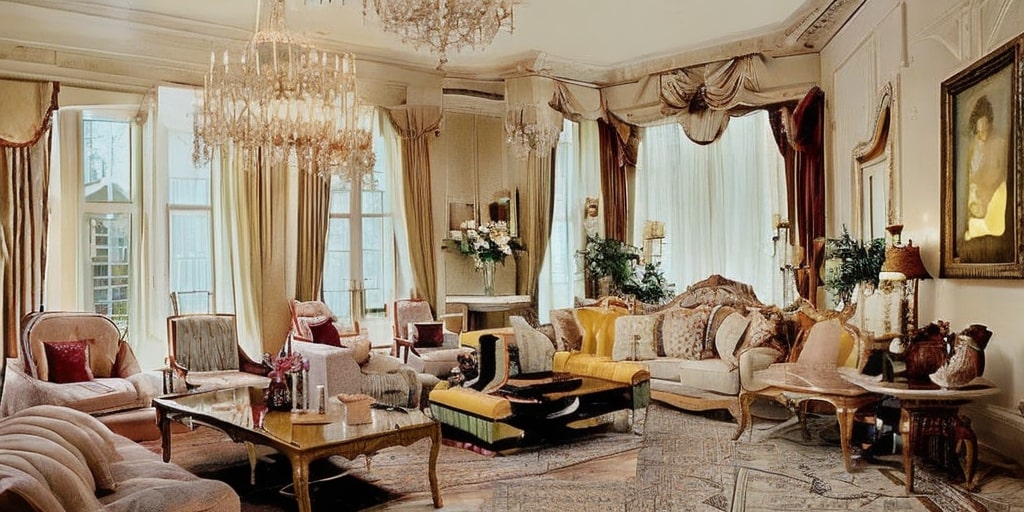}} \\
    \end{tabular}   
    \arrayrulecolor{black}
\vspace{-1.em}
\caption{Qualitative comparison between SD-L, SD-L+AttnS, MD, SyncD, and MAD on the generation of images with a prompt that does not fit well into all the specified aspect ratios (top: 512$\times$3072, bottom-left: 512$\times$2048, bottom-right: 512$\times$1024). We use the LDM as backbone, the same seed, and the prompt \textit{A fancy living room}.}
\label{fig:limitations_h}
\end{figure*}

\end{document}